\documentclass[a4paper, 10pt, conference]{ieeeconf}      % Use this line for a4
\usepackage{FG2026}

\FGfinalcopy % *** Uncomment this line for the final submission

\IEEEoverridecommandlockouts                              % This command is only
\usepackage{graphicx} % for pdf, bitmapped graphics files
\usepackage{amsmath} % assumes amsmath package installed
\usepackage{float}
\usepackage{booktabs}
\usepackage{amsmath}
\usepackage{amssymb}
\usepackage{multirow}
\usepackage{subcaption}

\usepackage{xcolor}
\usepackage[normalem]{ulem}
\providecommand{\coloneqq}{\mathrel{\mathop:}=}

\newcommand{\joshua}[1]{\textcolor{black}{#1}}
\newcommand{\joshuav}[1]{\textcolor{black}{#1}}

\def\FGPaperID{18} % *** Enter the FG2026 Paper ID here

\title{\LARGE \textbf{EMA: Effort Metric Attention for Anatomical Effort-Guided Human Motion Diffusion}}

\author{
  Joshua Siy$^{1}$, Huakun Liu$^{1}$, Yutaro Hirao$^{1}$,\\
  Monica Perusquia-Hernandez$^{1}$, Hideaki Uchiyama$^{1}$, Kiyoshi Kiyokawa$^{1}$\\
  $^{1}$Nara Institute of Science and Technology, Ikoma, Nara, Japan\\
  \texttt{\small siy.joshua\_samuel.sl7@g.ext.naist.jp,
  liu.huakun.li0@is.naist.jp,
  yutaro.hirao@is.naist.jp,}\\
  \texttt{\small perusquia@ieee.org,
  hideaki.uchiyama@is.naist.jp,
  kiyo@is.naist.jp}
}

\newcommand{\trendrow}[5]{
\begin{minipage}{\textwidth}\centering
\begin{subfigure}{.24\textwidth}\includegraphics[width=\linewidth]{#2}\subcaption{SALAD (Peak)}\end{subfigure}
\begin{subfigure}{.24\textwidth}\includegraphics[width=\linewidth]{#4}\subcaption{EMA (Peak)}\end{subfigure}
\begin{subfigure}{.24\textwidth}\includegraphics[width=\linewidth]{#3}\subcaption{SALAD (Collective)}\end{subfigure}
\begin{subfigure}{.24\textwidth}\includegraphics[width=\linewidth]{#5}\subcaption{EMA (Collective)}\end{subfigure}
\caption*{#1}
\end{minipage}\vspace{3mm}
}

\begin{document}

\ifFGfinal
\thispagestyle{empty}
\pagestyle{empty}
\else
\author{Anonymous FG2026 submission\\ Paper ID \FGPaperID \\}
\pagestyle{plain}
\fi
\maketitle

% \thispagestyle{fancy}
% \renewcommand{\headrulewidth}{0pt}
% \fancyhf{}
% \fancyhead[C]{2026 International Conference on Automatic Face and Gesture Recognition (FG)}
% !!!!!!!!!!! IMPORTANT: SELECT THE APPROPRIATE COPYRIGHT NOTICE BELOW !!!!!!!!!
% \fancyfoot[L]{U.S. Government work not protected by U.S. copyright}
% Crown government authors only:
% \fancyfoot[L]{978-X-XXXX-XXXX-X/26/\$31.00 \copyright 2026 Crown}
% European Union authors only:
% \fancyfoot[L]{978-X-XXXX-XXXX-X/26/\$31.00 \copyright 2026 European Union}
% All other papers:
% \fancyfoot[L]{979-8-3315-7231-0/26/\$31.00 \copyright 2026 IEEE}
%%% END BLOCK HEADER AND COPYRIGHT NOTICE %%%

\begin{abstract}
\joshua{
Human motion diffusion models can synthesize action sequences from text, but controlling motion intensity remains challenging.
Existing approaches rely on effort-related adverbs, which are ambiguous and fail to capture quantitative aspects such as pacing, often resulting in flat and monotonous dynamics.
We propose an intensity-control framework based on Effort Metric Attention (EMA), a cross-attention module that conditions diffusion on numerical effort signals.
Inspired by Laban Movement Analysis (LMA), the framework focuses on the \textit{Time} and \textit{Weight} effort factors.
We approximate these factors using two kinematic metrics: peak joint positional change for pacing and collective joint positional change for motion amount.
EMA enables fine-grained, region-wise control without costly post-hoc optimization.
We introduce two evaluation tasks, metric-to-motion consistency and body-part-level effort modulation, to assess numerical fidelity and localized control.
Experiments and a user study show near-monotonic alignment between specified effort levels, generated motion dynamics, and established LMA descriptors.
These results indicate effective and interpretable control of effort dynamics in practice.
}
\end{abstract}

%---------------------------------------------------------------------------------
\section{INTRODUCTION}
%---------------------------------------------------------------------------------

Human motion diffusion models present a promising alternative by generating animations from text~\cite{tevet2022human,chen2023executing}.
Such models reduce manual effort and enable motion creation on demand~\cite{tevet2022human, chen2023executing, hong2025salad}.
Nevertheless, current models face major challenges: they yield monotonous pacing and weak force due to the limited diversity of motion training data, require costly post-hoc editing, and have insufficient control for fine-grained aspects such as pacing and intensity.
Since body language accounts for 55\% of communication meaning~\cite{mehrabian1967Decoding}, achieving expressive motion control is essential.
However, models trained on existing motion-language datasets inherit inherent limitations. 
Sparse annotations and a limited range of dynamic or expressive motions restrict their ability to capture extreme-effort actions, leading to generic motions with limited variety.

One potential approach to addressing this limitation is to employ movement-analysis frameworks that explicitly model expressive dynamics.
Laban Movement Analysis (LMA) provides an interpretable representation of motion quality~\cite{aristidou2015folk,zacharatos2013emotion,turab2025dancestylerecognitionusing}.
LMA defines four effort factors—\textit{Time}, \textit{Weight}, \textit{Space}, and \textit{Flow}—that characterize qualitative aspects of motion.
%Such control is critical for applications in virtual reality and robotics, where agents must generate adaptive motions. 
In this work, we focus on \textit{Time}, relating to pacing, and \textit{Weight}, relating to the sense of force, which we approximate through motion amount. 
This is because both can be quantified through kinematic quantities.
These two factors do not fully represent LMA.
Nevertheless, they provide a practical basis for interpretable numerical control.

Motivated by the LMA’s framework, we propose an intensity-based motion generation framework that enables quantitative control of pacing and movement extent. %amount. 
Unlike prior work that relies on linguistic effort descriptions, our approach provides direct numerical control using LMA-inspired kinematic measures, based on the research linking kinematic features to LMA Effort~\cite{zacharatos2013emotion,aristidou2015folk}.
In particular, we defined peak joint positional change for pacing and collective joint positional change for motion amount. 
We designed Effort Metric Attention (EMA), which integrates these metrics via cross-attention to align user-specified values with motion synthesis, inspired by a skeleton-aware latent diffusion model (SALAD)~\cite{hong2025salad}.
EMA supports body-part–specific modulation while preserving coherence and eliminates the need for costly post-hoc editing. 
Finally, we augmented HumanML3D~\cite{guo2022generating} with speed-modified sequences to enhance training diversity.
This provides controllable variation in motion dynamics.
We evaluated EMA through two tasks: metric-to-motion consistency and body-part effort modulation.
\joshua{
While the proposed metrics serve as kinematic approximations of LMA effort factors, 
Both quantitative evaluations and a user study show that manipulating these controls leads to consistent and monotonic changes in perceived effort as well as established LMA descriptors.
These results indicate effective control of LMA effort dynamics in practice.
}

\joshua{
Our contributions are as follows:  
\begin{itemize}
    \item \textbf{Intensity-Based Motion Control}: We propose an intensity-control framework for motion generation that enables quantitative and flexible modulation of pacing and motion amount across diverse actions.  
    \item \textbf{LMA-Inspired Kinematic Metrics}: Inspired by LMA’s \textit{Time} and \textit{Weight} factors, we define interpretable kinematic metrics for numerical control, namely peak joint positional change for pacing and collective joint positional change for motion amount. 
    \item \textbf{Effort Metric Attention Module}: We design EMA, which incorporates these metrics into a skeleton-aware diffusion model via cross-attention and enables expressive and coherent motion synthesis without costly post-hoc optimization.
    \item \textbf{Effort-Oriented Evaluation}: We introduce two effort-specific evaluation tasks, \textit{Metric-to-Motion Consistency} and \textit{Body-Part Effort Modulation}, and conduct a user study to assess perceived effort controllability and alignment with human perception.
\end{itemize}
}

%---------------------------------------------------------------------------------
\section{Related Work}
\subsection{Text-to-Motion Generation}
%---------------------------------------------------------------------------------
Text-to-motion generation aligns textual prompts with motion sequences. 
Early works relied on predefined action labels~\cite{guo2020action2motion,petrovich2022temos}. 
Later studies adopted diffusion probabilistic frameworks and generated coherent sequences by denoising conditioned noise~\cite{ho2020denoising,rombach2022high}. 
Tevet et al. introduced diffusion to text-to-motion synthesis and produced high-fidelity motions~\cite{tevet2022human}.

Recent approaches can be grouped into two directions: generation on raw motion features and on latent spaces.  
On raw features, MDM~\cite{tevet2022human} directly denoises motion trajectories. However, it suffers from high dimensionality.  
On latent spaces, models such as T2M-GPT and SALAD~\cite{zhang2023t2mgpt,hong2025salad} compress motions into compact representations before applying diffusion, improving efficiency and alignment with language encoders such as CLIP, BERT, or DistilBERT~\cite{radford2021learning,devlin2019bertpretrainingdeepbidirectional}.  
Latent VAEs~\cite{habibie2017recurrent,hong2025salad} approximate the posterior with a neural encoder and optimize a variational lower bound for scalable inference, enabling structured representation learning through skeleton-aware designs~\cite{Jang_motion_puzzle_2022,park_2021_STGraph}.  
For instance, SALAD’s VAE employs skeleto-temporal convolution and pooling to compress SMPL-based joints into atomic limb-level representations, enabling information exchange across joints and frames while preserving temporal coherence~\cite{hong2025salad}.

Diffusion in latent spaces also facilitates conditioning on sparse cues such as keyframes, trajectories, or action styles~\cite{karunratanakul2023guidedmotiondiffusioncontrollable,wan2024tlcontrol}.  
Qualitative prompts such as fast (pacing) or strong (force) are ambiguous because interpretation depends on dataset variability. 
Quantitative priors enable precise, single-pass, effort-guided synthesis~\cite{yao2023moconvq,chen2023executing}.  

Zero- and few-shot controllability has also been explored, since motion control inevitably requires generalization to unseen conditions.  
Image-editing methods such as SDEdit~\cite{meng2022sdeditguidedimagesynthesis} and Prompt-to-Prompt~\cite{hertz2022prompttopromptimageeditingcross} inspired motion editing via attention modulation~\cite{hong2025salad}.  
In motion generation, SALAD modulates cross-attention for text-driven edits without retraining, while MDM and CoMo~\cite{tevet2022human,huang2024comocontrollablemotiongeneration} rely on masking or optimization strategies.  
FG-MDM~\cite{shi2024fgmdmzeroshothumanmotion} refines vague prompts into body-part descriptions via LLM guidance, and Go to Zero~\cite{fan2025zerozeroshotmotiongeneration} scales to compositional synthesis with large-scale datasets and LLaMA-based models.  
These approaches require iterative adjustments or extensive text priors, contrasting with our pre-computed metric approach.

\joshua{
Nevertheless, dataset scarcity remains a bottleneck: benchmarks such as HumanML3D provide limited paired text--motion samples, which restricts few-shot adaptation and causes extrapolation bias.
These limitations make it difficult to reliably control motion intensity through language alone, especially under unseen or sparsely represented conditions.
This motivates the use of quantitative priors as fixed, pre-computed conditioning signals that can generalize across sparse data regimes; a detailed comparison of MDM variants, their controllability mechanisms, and target tasks is provided in the supplementary material.
}

%---------------------------------------------------------------------------------
\subsection{Quantitative Motion Control}
%---------------------------------------------------------------------------------
Quantitative motion control employs measurable values such as velocity, force, or spatial metrics. 
Laban Movement Analysis (LMA) defined four effort factors: \textit{Space}, \textit{Weight}, \textit{Time}, and \textit{Flow}, later extended with attributes like velocity and force~\cite{aristidou2015folk,zacharatos2013emotion}. 
Trajectory optimization under space-time constraints achieved realism but required heavy computation~\cite{witkin1988spacetime}. 
Recent studies applied LMA to recognition tasks in dance, linking kinematics to effort~\cite{turab2025dancestylerecognitionusing}.

Diffusion-based motion control incorporates numerical values. However, it often introduces overhead because it adds additional cross-attention or embedding computations at each step, further increasing computational cost. 
FlexMotion encoded joint velocities under rigid constraints~\cite{tashakori2025flexmotion}. 
Mojito aligned motion via optical flow with calibration cost~\cite{he2024mojito}. 
Multi-view diffusion conditioned on camera parameters for consistent 3D synthesis~\cite{long2023wonder3d,huang2024mvadapter}.

\joshua{
Our method differs from these approaches in that it directly regulates motion intensity through pre-computed effort metrics such as maximum and cumulative positional changes.
This design conditions motion generation on explicit numerical effort signals, rather than relying on linguistic effort descriptions or iterative optimization procedures.
As a result, motion intensity can be controlled consistently, and interpretable effort modulation can be achieved without additional post-hoc adjustments.
}

%---------------------------------------------------------------------------------
\section{Proposed Method}
\label{sec:method}
\subsection{Overview}
\begin{figure*}[t]
\centering
\includegraphics[width=0.8\textwidth]{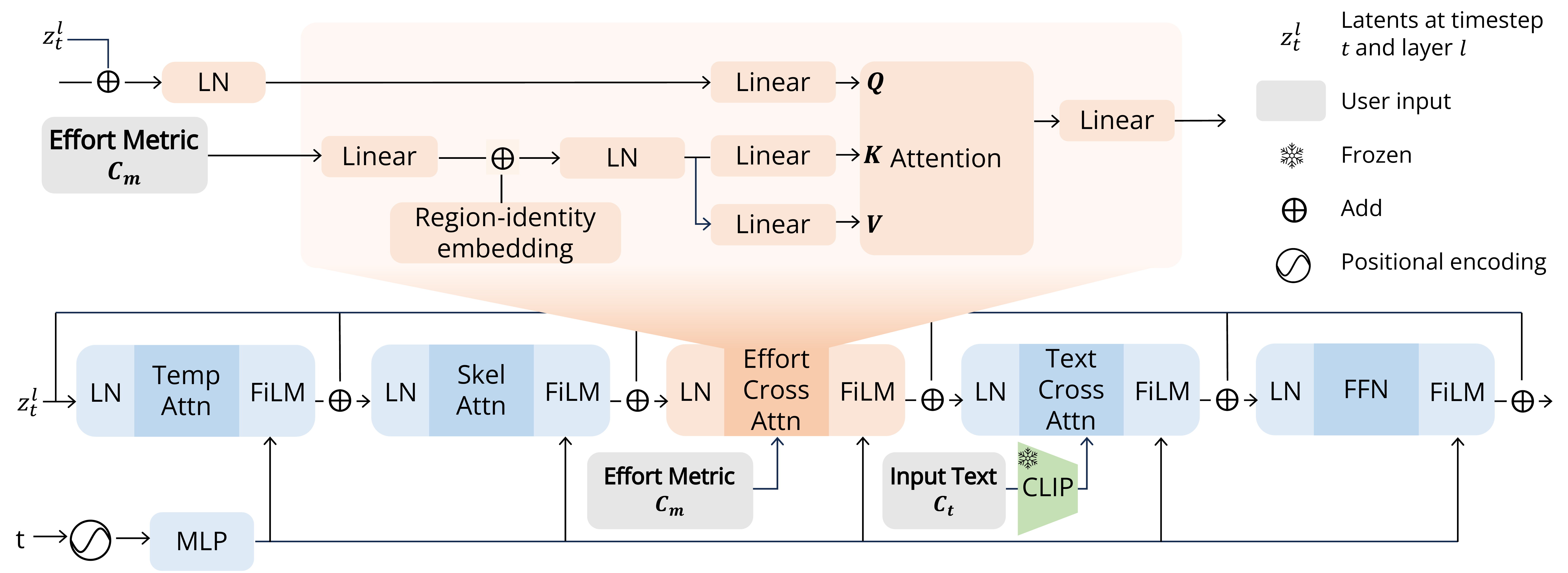} 
\caption{
\joshua{
Architecture of the Effort Metric Attention (EMA) module and its integration within the transformer block for intensity-based motion control.
The upper part illustrates EMA: effort metrics $\mathbf{c}_m$, defined by peak and collective joint positional change, are embedded, combined with region-identity embeddings, and used as keys and values in cross-attention, while motion latents serve as queries.
The lower part shows the transformer block, where temporal attention, skeletal attention, EMA-based cross-attention, and text cross-attention with CLIP-encoded prompts $\mathbf{c}_t$ are applied in sequence, followed by feed-forward layers.
User inputs are shown in bold black, and frozen modules are marked with a snowflake.}
}
\label{fig:updated_arch}
\end{figure*}

%---------------------------------------------------------------------------------
Our objective is to implement intensity-based motion control by enabling quantitative modulation of pacing and motion amount.  
To achieve this, we adopt a skeleton-aware motion latent representation~\cite{hong2025salad} that encodes both temporal and skeletal dynamics of a motion sequence.  
Based on this representation, we train a diffusion model with a latent denoiser that is equipped with Effort Metric Attention (EMA).  
EMA encodes effort through quantitative proxies such as peak and collective joint positional change.  
It injects these values into the diffusion process through cross-attention.  
The overall framework models the interactions among skeletal, temporal, effort, and textual cues~(Fig.~\ref{fig:updated_arch}).  
This design enables expressive and controllable motion generation.
  
%---------------------------------------------------------------------------------
\subsection{Effort Metric}
%---------------------------------------------------------------------------------
We define effort metrics directly from raw 3D joint positions, which can be obtained from any SMPL-based or joint-based motion representation.  
Joint angles are not required to reduce redundancy, because whole-body dynamics can be reconstructed from positions through inverse kinematics~\cite{li2022hybrikhybridanalyticalneuralinverse}.  
We compute effort metrics from these positional changes as quantitative measures of motion kinematics.  
These metrics serve as conditioning signals for transformer-based and body-part–specific motion synthesis. 

First, joints are grouped into \(N_g\) anatomical regions, such as the left and right legs. 
This grouping allows the computation of compact effort metrics, which enable efficient and interpretable motion synthesis. 
The grouping scheme is flexible and not tied to a specific dataset.

Given a motion sequence of shape \([T,D_f]\), where \(T\) is the number of frames and \(D_f\) the feature dimension, we obtain 3D joint positions as  
\(\mathbf{pos}\in\mathbb{R}^{T\times N_j\times 3}\), where \(N_j\) is the number of joints.  
For \(t=0,\dots,T-2\) and joint \(j\), the per-joint positional change is
\begin{align}
\mathrm{diff}_{t,j} &= \bigl\lVert \mathbf{pos}_{t+1,j}-\mathbf{pos}_{t,j}\bigr\rVert_2
\end{align}
which is the Euclidean displacement of joint \(j\) between consecutive frames.  
Group-wise per-frame change is then defined as
\begin{align}
\overline{\mathrm{diff}}_{t,g} &= \operatorname{mean}_{j\in\mathcal{G}_g}\bigl(\mathrm{diff}_{t,j}\bigr),\quad g=1,\dots,N_g
\end{align}
that is, the average positional change of joints within anatomical group \(\mathcal{G}_g\).  
Two metrics per group are derived as
\begin{align}
\text{peak change}_g &= \max_{\,t}\,\overline{\mathrm{diff}}_{t,g}\\
\text{collective change}_g &= \sum_{t=0}^{T-2}\overline{\mathrm{diff}}_{t,g}
\end{align}
Let \(N_p = 2\) denote the number of metrics per group. 
The complete metric vector is then defined as \joshua{$\mathbf{c}_m \in \mathbb{R}^{B \times N_g \times N_p}$}.

Inspired by LMA, peak joint positional change captures maximum displacement, indicating the \emph{Time} factor for pacing.
Collective positional change accumulates displacement, reflecting the \emph{Weight} factor for force, approximated by motion amount. 
These kinematic measures enable interpretable effort dynamics~\cite{kim_comprehensive_2025,voas_what_2023}. 
The control signal $\mathbf{c}_m$ adjusts pacing and force across body parts during generation.

\begin{figure}[t]
\centering
\includegraphics[width=0.8\linewidth]{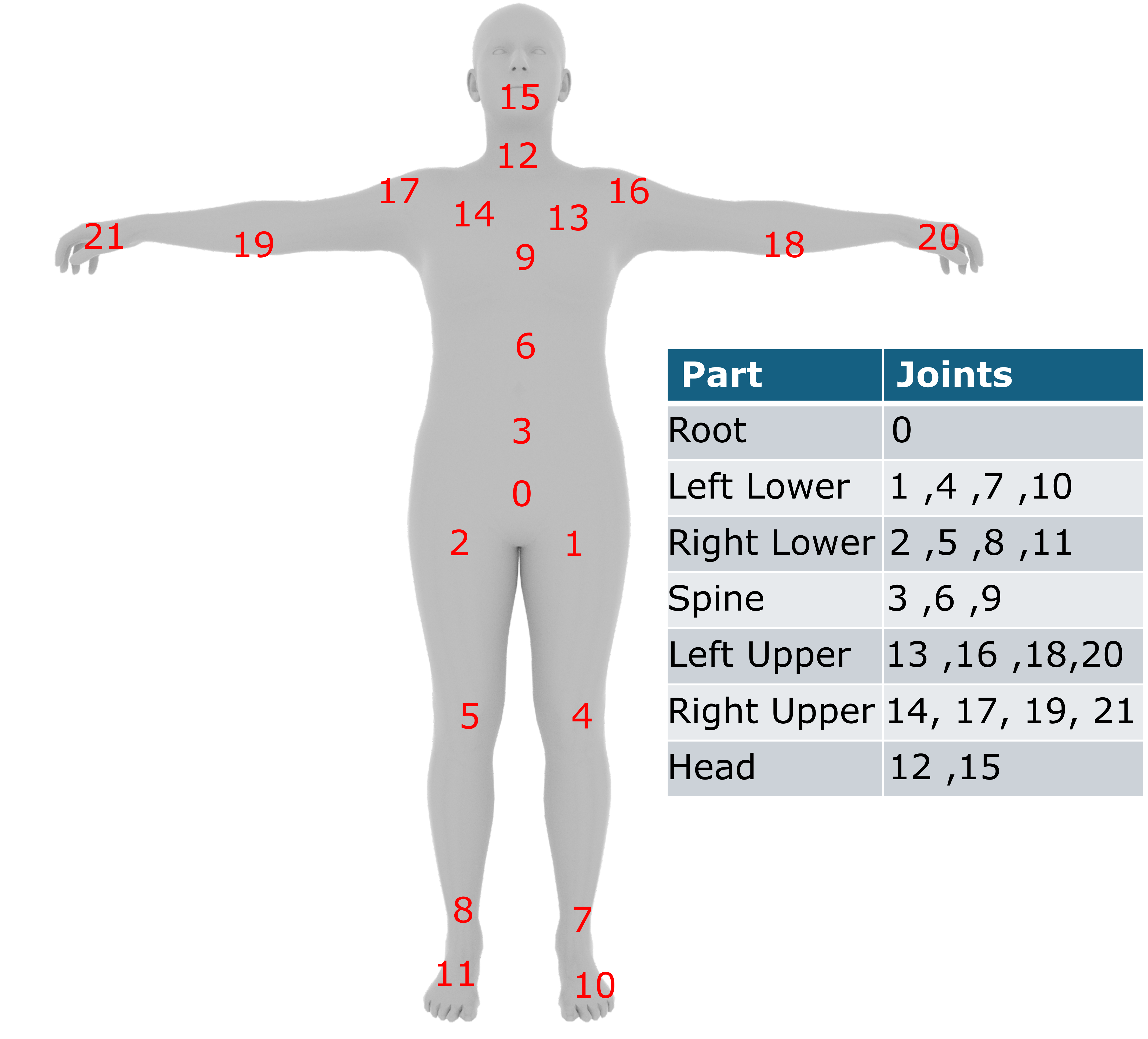}
\caption{SMPL joints are grouped into \(N_g\) anatomical regions such as the left and right legs. 
This grouping allows the computation of compact effort metrics, which capture peak and collective positional changes used in the EMA module.}
\label{fig:SMPL_joint_map}
\end{figure}

In our implementation, we adopt HumanML3D~\cite{guo2022generating} as the source dataset.  
We use the SALAD grouping scheme~\cite{hong2025salad}, where the set of \(N_j\) SMPL joints is organized into seven anatomical regions. 
These regions are root, left lower, right lower, spine, left upper, right upper, and head (Fig.~\ref{fig:SMPL_joint_map}).  

%---------------------------------------------------------------------------------
\subsection{Effort Metric Attention Module}
%---------------------------------------------------------------------------------
To incorporate quantitative effort metrics into motion generation, we propose the Effort Metric Attention (EMA) module within the skeleton-aware latent diffusion model~\cite{hong2025salad} (Fig.~\ref{fig:updated_arch}).
The EMA module conditions the generation process on numerical effort values, which enables fine-grained control of motion intensity and speed across skeletal groups.

\joshua{
Formally, let the effort metrics be $\mathbf{c}_m \in \mathbb{R}^{B \times N_g \times N_p}$ and the motion latents at diffusion timestep $t$ and transformer layer $l$ be $\mathbf{z}_t^l \in \mathbb{R}^{B \times T \times N_g \times D}$, where $B$ is the batch size, $T$ the number of frames, $N_g$ the number of skeletal groups, $N_p$ the metric dimension, and $D$ the latent dimension.
We linearly project $\mathbf{c}_m$ to $\mathbb{R}^{B \times N_g \times D}$ and add a learnable region-identity embedding $\mathbf{P}_{\mathrm{id}} \in \mathbb{R}^{1 \times N_g \times D}$ to preserve group semantics.
After LayerNorm, we obtain keys and values via linear projections, $\mathbf{K}, \mathbf{V} \in \mathbb{R}^{B \times H \times N_g \times D'}$.
We reshape motion latents to $\mathbb{R}^{B \times (T N_g) \times D}$ and linearly project them into queries $\mathbf{Q} \in \mathbb{R}^{B \times H \times (T N_g) \times D'}$, where $H$ is the number of heads and $D' = D/H$.
Cross-attention is defined as
\begin{align}
\operatorname{Attn}(\mathbf{Q}, \mathbf{K}, \mathbf{V})
&= \operatorname{softmax}\!\left(\frac{\mathbf{Q}\mathbf{K}^\top}{\sqrt{D'}}\right)\mathbf{V},
\end{align}
where the softmax is taken over the key index (i.e., over $N_g$) for each query.
}

The attention output of shape \(\mathbb{R}^{B \times H \times (T N_g) \times D'}\) is projected back to \(\mathbb{R}^{B \times (T N_g) \times D}\), reshaped to \(\mathbb{R}^{B \times T \times N_g \times D}\), and combined with the original motion latents through a residual connection.
A dropout rate of \(0.1\) is applied to the attention weights to improve generalization.

\joshua{
As illustrated in Fig.~\ref{fig:updated_arch}, the EMA module is placed after temporal and skeletal attention and before text cross-attention.
The module operates on motion latents that have already incorporated spatio-temporal and skeletal context, and its output is subsequently processed by text-based cross-attention.
}

%---------------------------------------------------------------------------------
\subsection{Effort-Guided Skeleton-Aware Denoiser}
%---------------------------------------------------------------------------------
\joshua{
We build upon SALAD’s VAE~\cite{hong2025salad}, which encodes skeletal structure and motion dynamics into a compact latent space.
On top of this latent representation, we train a transformer-based diffusion denoiser jointly conditioned on text prompts $\mathbf{c}_t$ and effort metrics $\mathbf{c}_m$.
Further implementation details and design rationales for the denoiser architecture, including the justification of the hierarchical attention ordering, are provided in the supplementary material.
}

The denoiser consists of sequential transformer blocks integrating temporal attention (TempAttn), skeletal attention (SkelAttn), effort metric attention (MetricAttn), text-based cross-attention (CrossAttn), and feed-forward network (FFN). 
Each block is followed by residual connections~\cite{he2015deepresiduallearningimage}, layer normalization (LN)~\cite{ba2016layernormalization}, and FiLM modulation~\cite{perez2017filmvisualreasoninggeneral} with respect to the diffusion timestep. 
For motion latents \(\mathbf{z}_t^l \in \mathbb{R}^{B \times T \times N_g \times D}\) at timestep \(t\) and layer \(l\), the update is defined as:
\begin{align}
\mathbf{z}_t^l &\leftarrow \mathbf{z}_t^l 
  + \operatorname{FiLM}\!\bigl(\operatorname{TempAttn}(\operatorname{LN}(\mathbf{z}_t^l))\bigr), \\
\mathbf{z}_t^l &\leftarrow \mathbf{z}_t^l 
  + \operatorname{FiLM}\!\bigl(\operatorname{SkelAttn}(\operatorname{LN}(\mathbf{z}_t^l))\bigr), \\
\mathbf{z}_t^l &\leftarrow \mathbf{z}_t^l 
  + \operatorname{FiLM}\!\bigl(\operatorname{MetricAttn}(\operatorname{LN}(\mathbf{z}_t^l), \mathbf{c}_m)\bigr), \\
\mathbf{z}_t^l &\leftarrow \mathbf{z}_t^l 
  + \operatorname{FiLM}\!\bigl(\operatorname{CrossAttn}(\operatorname{LN}(\mathbf{z}_t^l), \operatorname{CLIP}(\mathbf{c}_t))\bigr).
\end{align}
Text prompts are encoded by a frozen CLIP encoder~\cite{radford2021learning}.  

The denoiser predicts the diffusion velocity \(\mathbf{v}_t\)~\cite{salimans2022progressivedistillationfastsampling}:
\begin{equation}
\mathbf{v}_t = \alpha_t \boldsymbol{\epsilon} - \sigma_t \mathbf{x},
\end{equation}
where \(\boldsymbol{\epsilon}\) is sampled noise, \(\mathbf{x}\) is the clean motion sample, and \(\alpha_t, \sigma_t\) are schedule parameters. 
The training loss is expressed as
\begin{equation}
L_{\mathrm{denoiser}} = \lVert \hat{\mathbf{v}}_t - \mathbf{v}_t \rVert_2^2,
\end{equation}
where \(\hat{\mathbf{v}}_t\) is the predicted velocity.  

Classifier-free guidance~\cite{ho2022classifierfreediffusionguidance} with weight \(w\) is employed to balance conditional and unconditional signals:
\begin{align}
\hat{\mathbf{v}}_\theta(\mathbf{z}_t, t, \mathbf{c}_t, \mathbf{c}_m)
  &\coloneqq \mathbf{v}_\theta(\mathbf{z}_t, t, \varnothing, \mathbf{c}_m) \nonumber\\
  &\quad + w\!\left[
      \mathbf{v}_\theta(\mathbf{z}_t, t, \mathbf{c}_t, \mathbf{c}_m)
      - \mathbf{v}_\theta(\mathbf{z}_t, t, \varnothing, \mathbf{c}_m)
    \right].
\end{align}

During inference, users provide effort metrics and sequence length, which guide DDIM sampling~\cite{song2022ddim} to generate motion sequences with the desired dynamics.  
Baseline metrics corresponding to typical normal-paced motions are computed as averages over HumanML3D samples~(Table~\ref{tab:start_inference_metrics}).  

\begin{table}[t]
\centering
\caption{Baseline Effort Metrics from HumanML3D. 
Values show average peak and collective positional changes for each body region.}
\begin{tabular}{lcc}
\toprule
Body Part & Peak Change & Collective Change \\
\midrule
Root        & 0.010 & 1.256 \\
Left Lower  & 0.015 & 1.279 \\
Right Lower & 0.015 & 1.279 \\
Spine       & 0.010 & 1.252 \\
Left Upper  & 0.014 & 1.293 \\
Right Upper & 0.014 & 1.295 \\
Head        & 0.012 & 1.262 \\
\bottomrule
\end{tabular}
\label{tab:start_inference_metrics}
\end{table}

%---------------------------------------------------------------------------------
\subsection{Data Augmentation}
\label{ssec:data_augmentation}
%---------------------------------------------------------------------------------
HumanML3D~\cite{guo2022generating} provides a large dataset of everyday human activities obtained from motion capture, downsampled to 20 fps.  
However, most motions are moderate in speed and intensity, and the dataset includes few examples of extreme-effort dynamics.  
In addition, hardware constraints limit the ability to record fast movements~\cite{pan_romo_2024}, which reduces motion diversity.

To mitigate this limitation, we introduced pacing-based data augmentation at a fixed 20~fps to simulate both high- and low-effort actions.  
Two complementary strategies are applied:  
\begin{itemize}
    \item \textbf{Faster motions}: frames are accelerated by removing $k$ consecutive frames between sampled frames ($k \in \{1,2\}$ in our experiments), increasing perceived speed by compressing the sequence into the same playback duration, simulating rapid, high-effort dynamics.  
    \item \textbf{Slower motions}: intermediate frames are interpolated by inserting $m$ additional frames between consecutive frames ($m \in \{1,2\}$ in our experiments), reducing perceived speed by extending the sequence duration, simulating low-effort dynamics.
\end{itemize}

\joshua{
These augmentations systematically vary peak and collective positional changes, increasing the effective training set size by approximately $3\times$ without external data.
The added diversity improves effort-guided training and supports finer control of joint dynamics at inference.
An ablation study evaluating the impact of this data augmentation is reported in the supplementary material.
}

%---------------------------------------------------------------------------------
\section{Experiments}
\subsection{Experimental Setup}
%---------------------------------------------------------------------------------
We evaluate EMA through two complementary experiments.  
First, \textbf{Metric-to-Motion Consistency} evaluates whether generated motions follow user-specified effort metrics and compares the results with SALAD~\cite{hong2025salad}.  
Second, \textbf{Body-Part Effort Modulation} examines suppression and amplification of individual body parts.
\joshua{ In addition, we conduct an \textbf{Ablation Analysis} to isolate the contribution of Peak and Collective effort components, and a \textbf{User Study} to assess perceived effort controllability and motion plausibility from a human perspective.}

\joshua{ 
\textbf{Dataset.}
We use HumanML3D~\cite{guo2022generating}, which contains 14{,}616 motion sequences paired with 44{,}970 textual descriptions.
We split the data into 80/15/5 for train/validation/test.
We apply the augmentation described in Section~\ref{ssec:data_augmentation} to the training, validation, and test splits.
We use the test split only to report standard metrics in Table~\ref{tab:dist_metrics} and the effort metric MAE in Table~\ref{tab:merged_evaluation}.
}

\joshua{
\textbf{Training.}
The model is trained on a single NVIDIA A6000 for about two days using AdamW~\cite{loshchilov2019decoupledweightdecayregularization} with a learning rate $5\times10^{-4}$ and a decay factor of $0.1$ at 50k iterations.
}

%---------------------------------------------------------------------------------
\subsection{Metric-to-Motion Consistency}
%---------------------------------------------------------------------------------
\joshua{
\textbf{Evaluation protocol.}
We test whether generated motions follow user-specified effort scalars by measuring how peak and collective positional changes vary with controlled inputs.
We use all prompts in Table~\ref{tab:action_categories} and scale the baseline effort metrics (Table~\ref{tab:start_inference_metrics}) by multipliers from 0.7 to 1.3 (i.e., $\pm 30\%$).}

\joshua{
To ensure sufficiently separated test conditions, we use a step size of 0.1, guided by reported sensitivity ranges in motion perception~\cite{mckee1984,werkhoven1992}.
During inference, motions are generated using DDIM sampling with 50 steps and a classifier-free guidance weight of 7.5.
}

\joshua{
To validate controllability, we perform a two-part evaluation.
First, we quantify anatomical monotonicity by computing Spearman rank correlations between the input scalars and the resulting peak and collective positional changes for each of the seven body parts.
We count body parts that satisfy $p < 0.05$ and $r > 0.5$, and report the percentage of body parts that show strong monotonic alignment.
}

\joshua{
Second, we complement this structural analysis with an external validation using a computational Laban Movement Analysis (LMA) framework~\cite{samadani2020affectivemovementgenerationusing}.
Although our control signals are not direct LMA parameters, this analysis examines whether modulation in our kinematic proxy space induces consistent trends in LMA-derived descriptors for Weight, Time, and Flow.
We exclude the Space component, as our effort modifiers target dynamic intensity rather than spatial trajectory.
}

\begin{figure*}[t]
\centering
\trendrow{Lower body (Runs)}{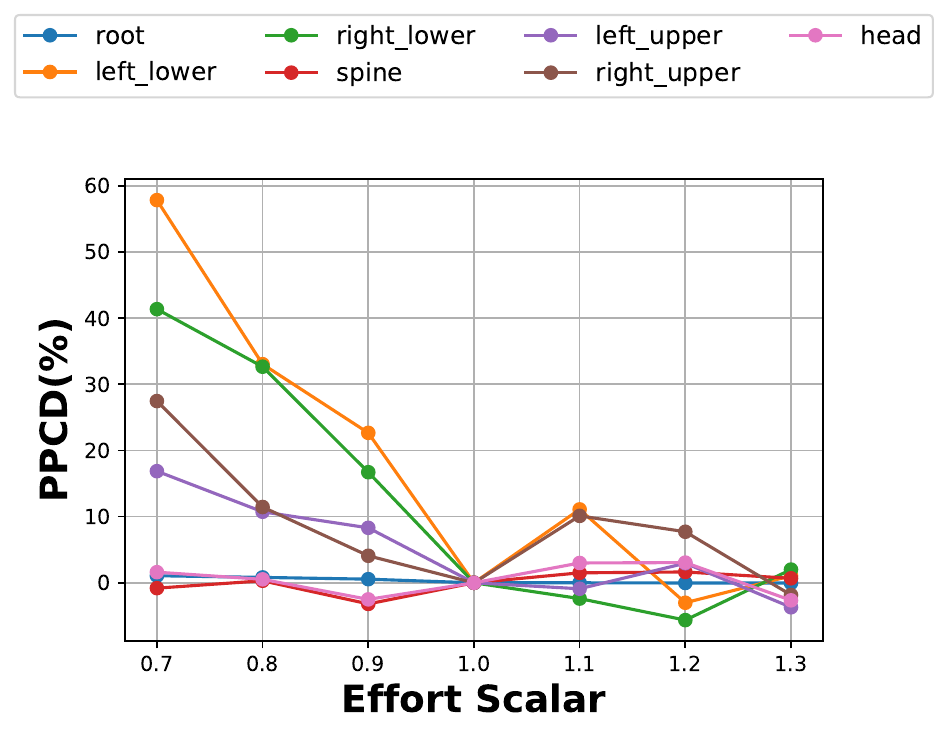}{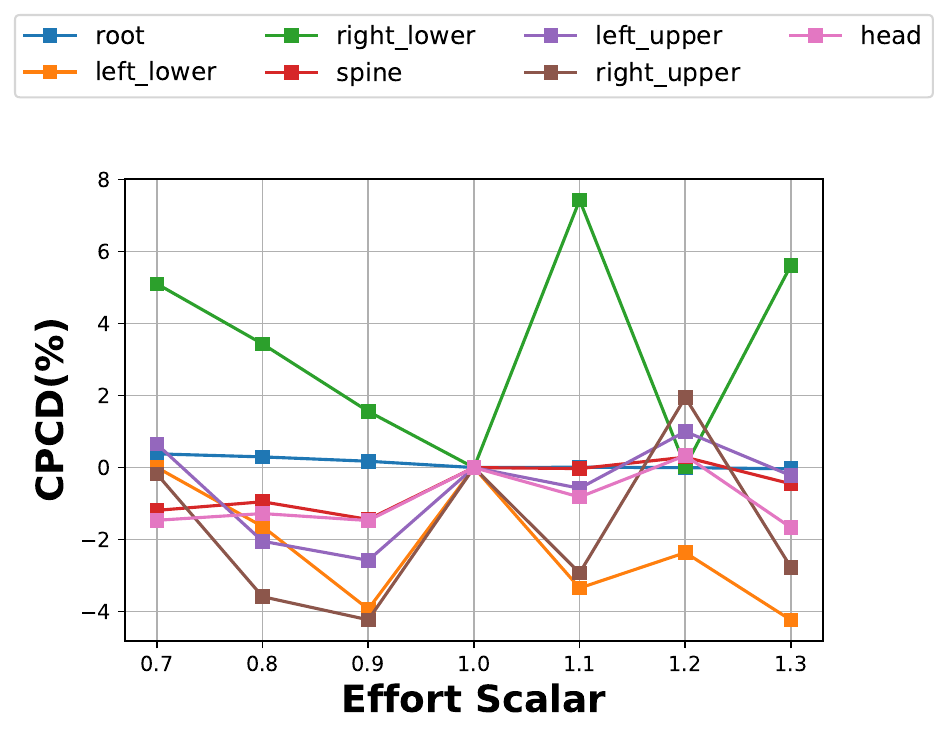}{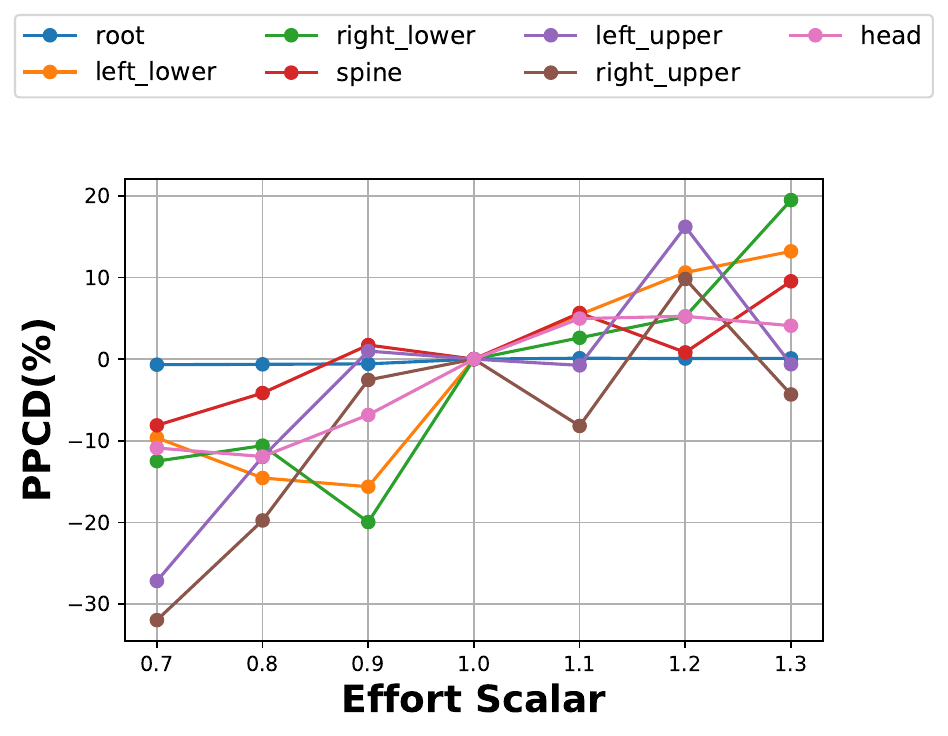}{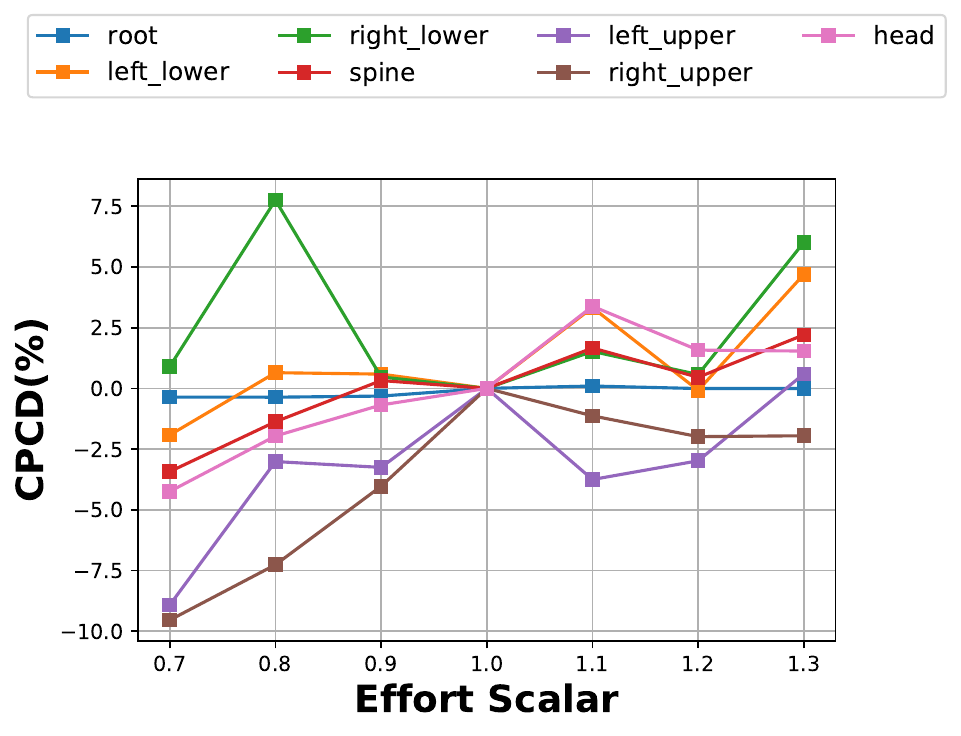}
\trendrow{Upper body (Swings)}{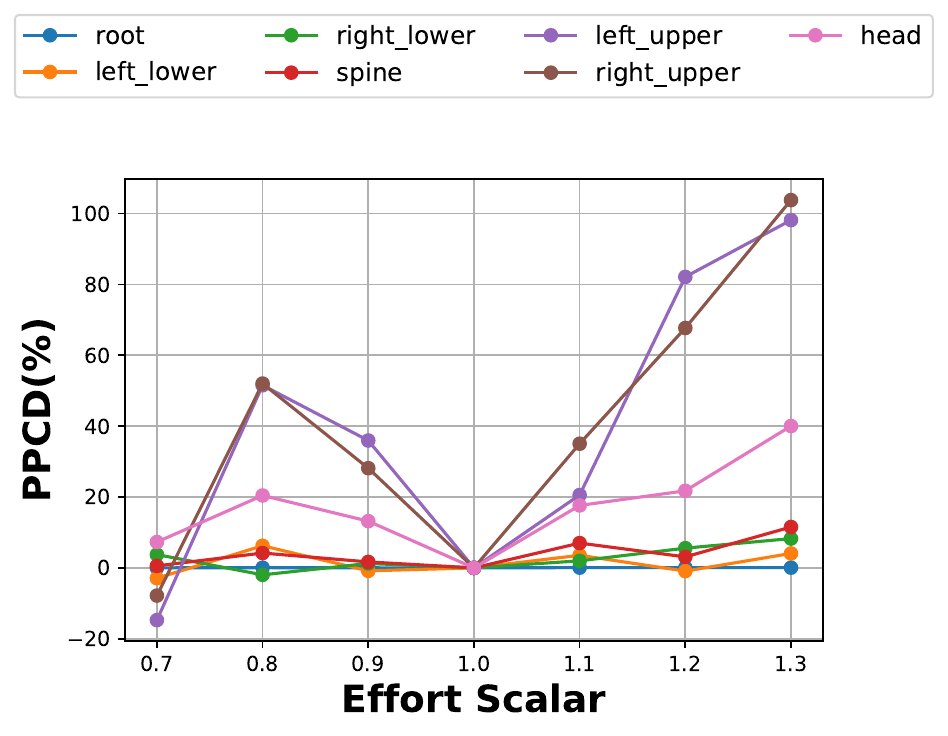}{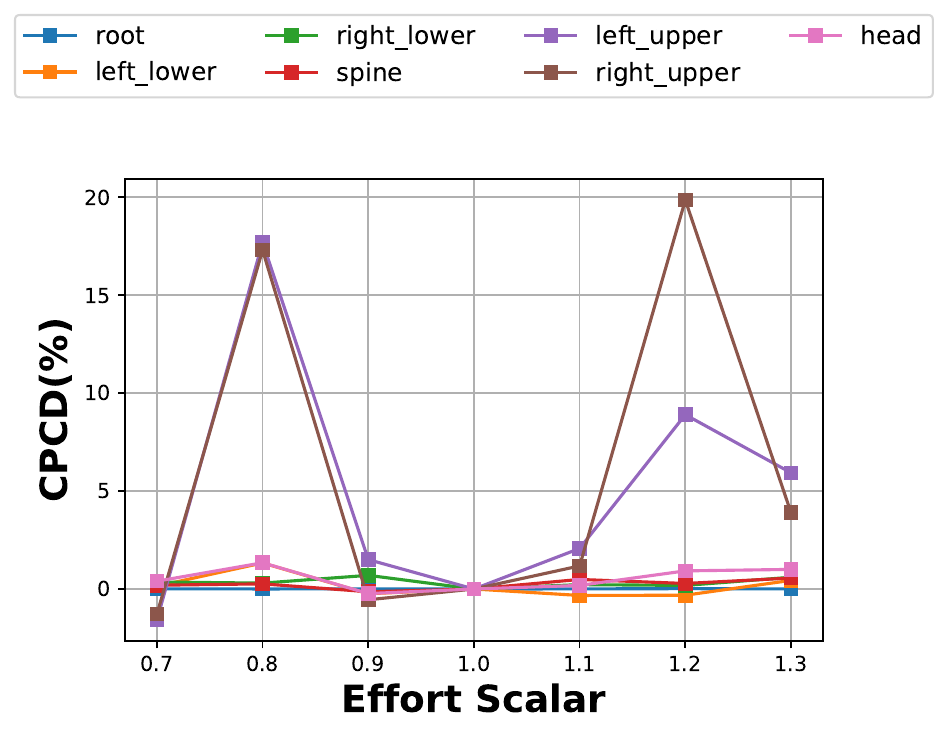}{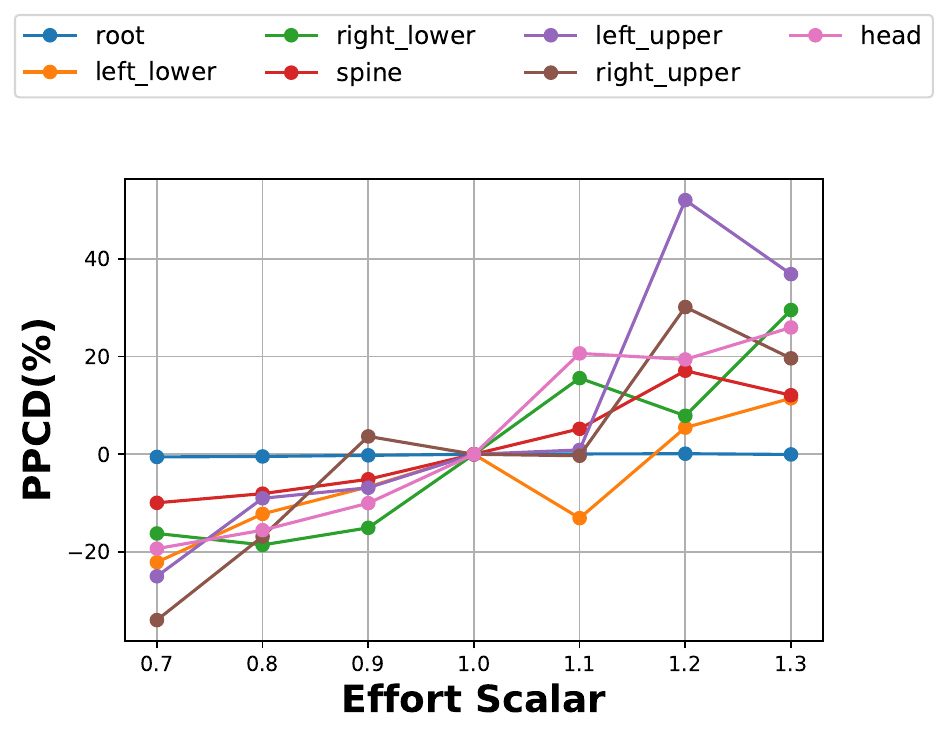}{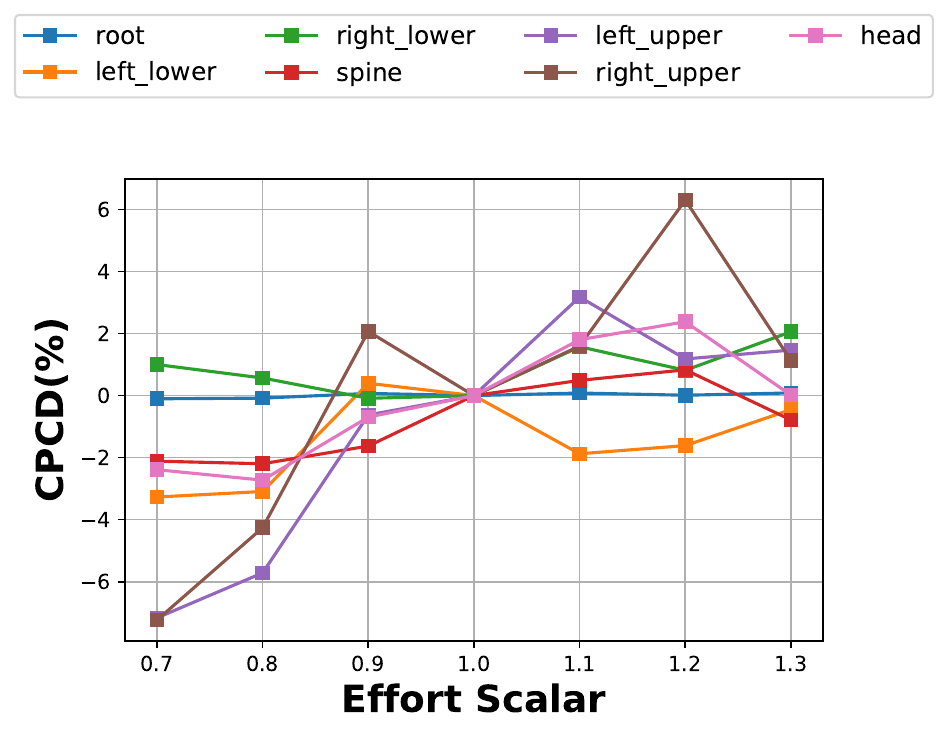}
\trendrow{Full body (Bends Over)}{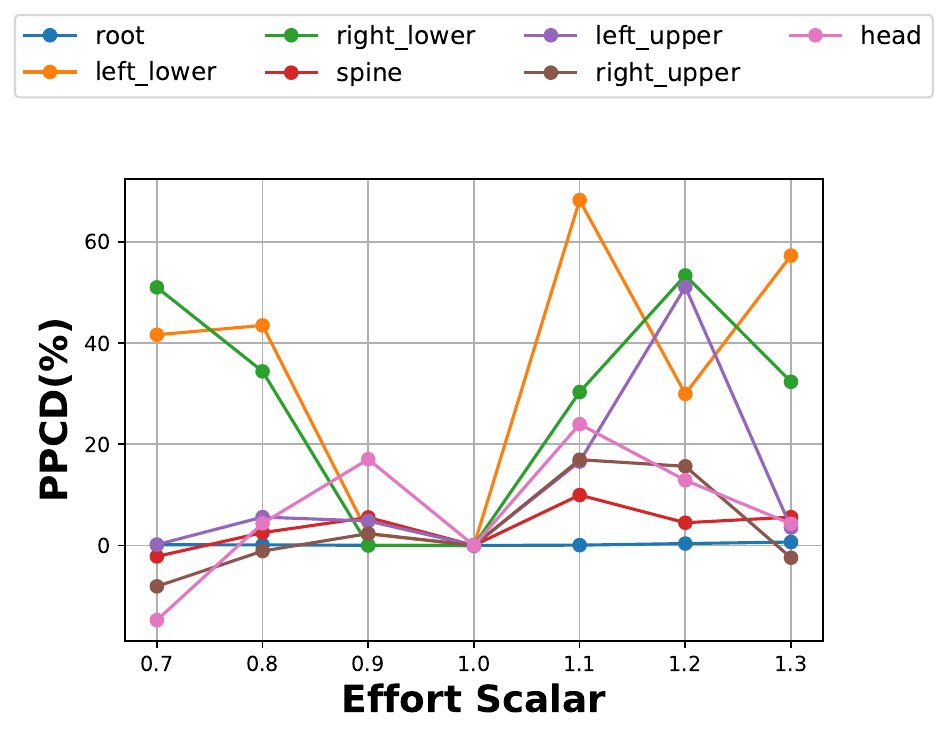}{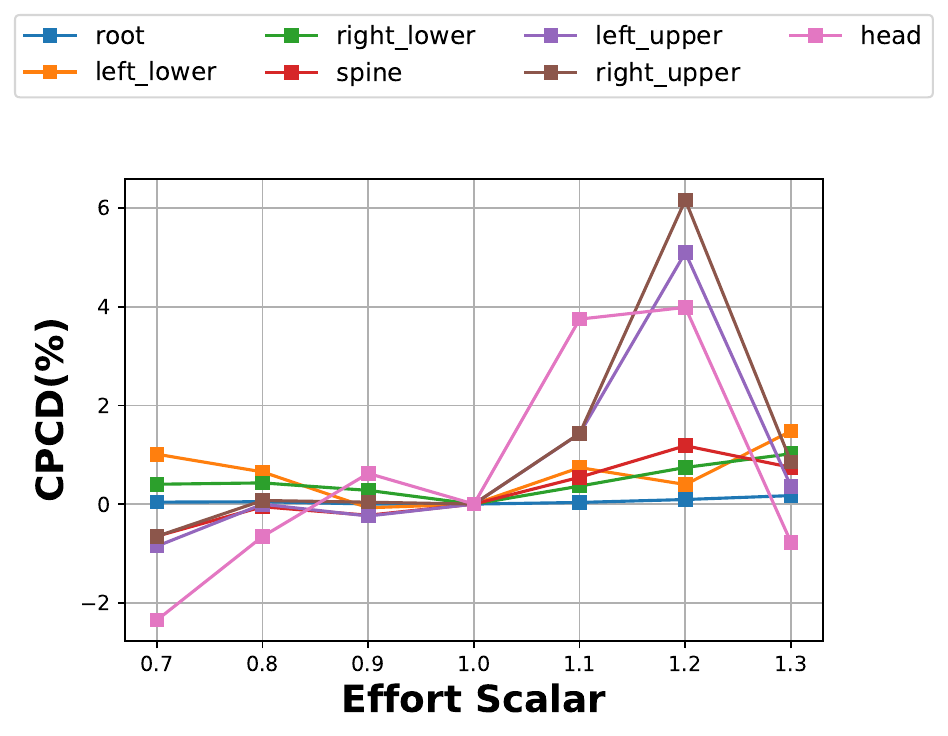}{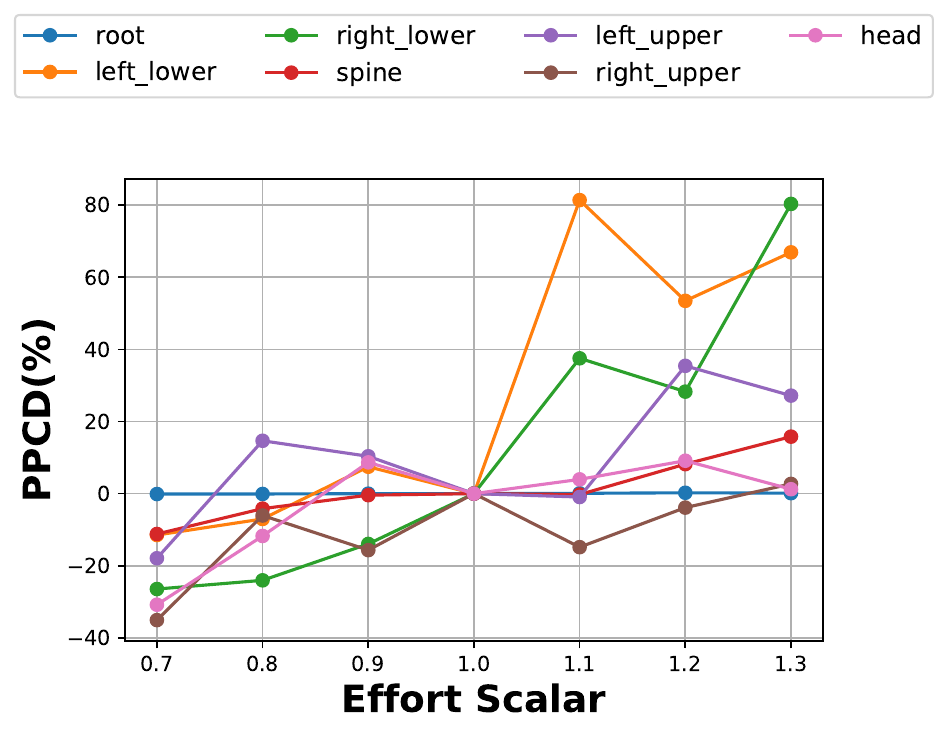}{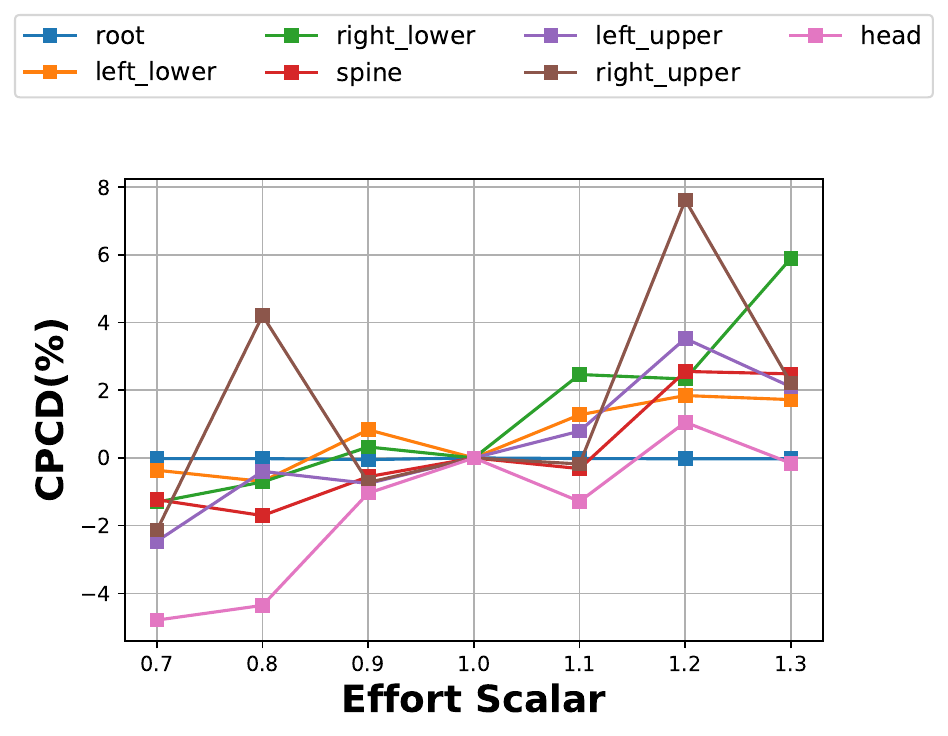}
\caption{\joshua{Trend comparison across representative actions for SALAD and EMA under 120 frames.  
Each row shows SALAD--PPCD, SALAD--CPCD, EMA--PPCD, and EMA--CPCD.  
PPCD = Peak Positional Change Difference, CPCD = Collective Positional Change Difference.  
EMA yields monotonic, near-proportional scaling (notably in PPCD), while SALAD exhibits irregular trends and asymmetries.  
For SALAD, the horizontal axis values (0.7--1.3) correspond to adverbs: \textit{Extremely Slow} (0.7), \textit{Very Slow} (0.8), \textit{Slow} (0.9), \textit{Normal} (1.0), \textit{Fast} (1.1), \textit{Very Fast} (1.2), and \textit{Extremely Fast} (1.3).}}
\label{fig:trend_compact}
\end{figure*}

\joshua{
\textbf{Results.}
EMA maintains strong monotonic alignment between input effort scalars and generated motion across diverse actions.
As shown in Fig.~\ref{fig:trend_compact} and Table~\ref{tab:merged_evaluation}, peak positional changes exhibit consistent, near-monotonic trends, while SALAD shows irregular and non-monotonic responses.
These trends also correspond to monotonic behavior in established LMA descriptors.
}

\begin{table}[t]
\centering
\caption{Categorization of action prompts.
Action classes are grouped into lower body, upper body, and full body categories for region-specific analysis of metric-to-motion consistency.}
\begin{tabular}{l p{5cm}}
\toprule
Category & Actions \\ \midrule
Lower body & lunges, walks, runs, kicks \\
Upper body & waves, waves an arm, punches, throws ball, \\
           & swings arms, shakes arms \\
Full body  & squats, dances, jumps, bends over \\
\bottomrule
\end{tabular}
\label{tab:action_categories}
\end{table}

\begin{table*}[t]
\centering
\setlength{\tabcolsep}{4pt}
\caption{\joshua{Unified quantitative evaluation. We report Mean Absolute Error (MAE) and Monotonicity Rates (\%) for structural metrics, and Monotonicity Accuracy (\%) for Laban kinetic metrics (Weight, Flow, Time).}}
\begin{tabular}{l c c c c c c c}
\toprule
\multirow{2}{*}{\textbf{Model}} &
\multicolumn{2}{c}{\textbf{Effort Metric MAE} $\downarrow$} &
\multicolumn{2}{c}{\textbf{Structural Mono. (\%)} $\uparrow$} &
\multicolumn{3}{c}{\textbf{Laban Mono. (\%)} $\uparrow$} \\
\cmidrule(lr){2-3}
\cmidrule(lr){4-5}
\cmidrule(lr){6-8}
 & \textbf{Peak} & \textbf{Coll.}
 & \textbf{Peak} & \textbf{Coll.}
 & \textbf{Weight} & \textbf{Flow} & \textbf{Time} \\
\midrule
EMA w/ Peak + Coll. & \textbf{0.0597} & \textbf{1.574} & 74.5 & 60.2 & \textbf{78.6} & \textbf{85.7} & \textbf{92.9} \\
EMA w/ Peak-only   & 0.0644 & 1.786          & \textbf{76.5} & \textbf{63.3} & \textbf{78.6} & \textbf{85.7} & \textbf{92.9} \\
EMA w/ Coll.-only  & 0.0646 & 1.743 & 33.7 & 36.7 & 42.9 & 57.1 & 50.0 \\
SALAD              & - & - & 22.4 & 17.3 & 21.4 & 14.3 & 28.6 \\
\bottomrule
\end{tabular}
\label{tab:merged_evaluation}
\end{table*}

%---------------------------------------------------------------------------------
\subsection{Body-Part Effort Modulation}
%---------------------------------------------------------------------------------
\textbf{Evaluation protocol.}
This experiment evaluated whether the proposed EMA-based denoiser can suppress or amplify motion dynamics at the level of individual body parts while remaining consistent with text descriptions.  
We also examined zero-shot scenarios in which ambiguous prompts are disambiguated through effort metrics.  
During inference, text prompts are provided together with customized effort metrics \joshua{$\mathbf{c}_m \in \mathbb{R}^{B \times N_g \times N_p}$.}
Motions are generated using DDIM sampling with 50 steps and a classifier-free guidance weight of $7.5$.  
We then analyze how altering or zeroing out subsets of metrics affects local body-part dynamics.

\joshua{
\textbf{Prompts.}
We evaluate two prompts:
(1) ``A person waves both hands.''
(2) ``A person waves their hands.''
Prompt (1) explicitly specifies a bilateral arm action and is used to evaluate targeted suppression and amplification.
Prompt (2) is linguistically ambiguous and is used to test whether effort metrics can bias or disambiguate the generated motion through controlled input manipulation.
}

\joshua{
\textbf{Effort metric settings.}
We initialize effort metrics from the dataset (S0) and set all anatomical regions to near-zero values except for the left upper region under prompt for suppression (S1) (1).
For suppression--amplification (S2), all anatomical regions are suppressed while the left upper region is assigned an atypical Peak/Collective pair (0.3, 1.0) under prompt (1) to emphasize localized dynamics.
For context reinforcement under prompt (2), we retain only the right upper region metrics (S3A) or only the left-upper region metrics (S3B), suppressing all others.
For the both-arms condition (S30), single-arm metrics are mirrored across the left and right upper regions  to promote symmetric involvement.
}

\begin{figure*}[t]
\centering
\includegraphics[width=0.9\textwidth]{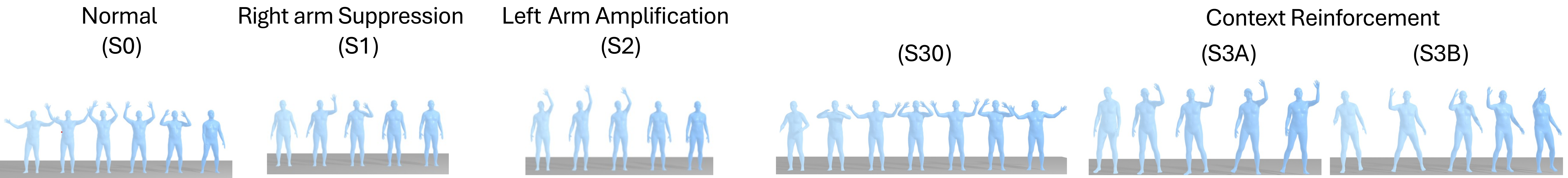}
\caption{\joshua{Examples of context manipulation via effort metrics.
Suppression (S1), amplification (S2), and context reinforcement (S30, S3A, S3B) are shown using color-coded meshes indicating temporal progression from start (light blue) to end (dark blue).
Each sequence illustrates motion generation under suppressive or context-reinforcing effort metric inputs.}}
\label{fig:suppressed_motion}
\end{figure*}

\joshua{
\textbf{Results.}
Figure~\ref{fig:suppressed_motion} summarizes the results.
\textit{Prompt (1): ``A person waves both hands.''}
Under the default metric setting, motion is correctly generated in both arms.
S1 suppresses movement in non-target body parts while preserving left-arm motion.
S2 further amplifies localized left-arm dynamics, while the remainder of the body remains largely static.
\textit{Prompt (2): ``A person waves their hands.''}
Due to linguistic ambiguity, the generated motion may involve either arm under the default metrics.
S30 produces symmetric bilateral waving by mirroring arm metrics.
S3A and S3B bias the motion toward exclusive right- or left-arm waving, respectively, resolving ambiguity through effort-metric control.
}

\joshua{
These results demonstrate that EMA enables precise body-part-level suppression and amplification of motion.
Moreover, numerical effort metrics can function as substitutes for, or reinforcements of, textual conditions in scenarios where language alone is ambiguous.
Additional isolation experiments analyzing effort-metric influence are provided in the supplementary material.
}

%---------------------------------------------------------------------------------
\joshua{\subsection{Ablation Analysis}}
%---------------------------------------------------------------------------------

\joshua{
\textbf{Evaluation protocol.}
To isolate the contribution of each effort component, we evaluate models conditioned on Peak-only and Collective-only effort metrics.
We report Effort Metric Mean Absolute Error (MAE), which measures deviations from the test set ground truth, and monotonicity scores evaluated on all prompts in Table~\ref{tab:action_categories} with scaled baseline effort metrics (Table~\ref{tab:start_inference_metrics}) by multipliers from 0.7 to 1.3 (i.e., $\pm 30\%$) in Table~\ref{tab:merged_evaluation}
}\joshuav{,to assess controllability, and FID to evaluate generative quality.}

\joshua{
\textbf{Peak-only.}
The Peak-only ablation primarily modulates instantaneous motion dynamics, functioning as a velocity-like scaling factor.
It achieves high peak monotonicity ($76.5$), indicating consistent control over maximum joint displacement.
However, it exhibits a higher collective MAE ($1.786$ compared to $1.574$ for the full model), demonstrating limited capacity to match cumulative displacement over extended sequences.
}

\joshua{
\textbf{Collective-only.}
In contrast, the Collective-only ablation more accurately captures total displacement, yielding a collective MAE of $1.743$.
However, its control over instantaneous intensity is weak, resulting in very low peak monotonicity ($33.7$) and an inability to reproduce sharp dynamic variations in high-effort motions.
}

% \joshua{
% \textbf{Peak + Collective.}
% Conditioning on both Peak and Collective metrics produces the most balanced behavior.
% Although the collective MAE ($1.581$) remains non-negligible, reflecting the intrinsic difficulty of modeling long-horizon trajectory accumulation, this configuration significantly improves upon the Peak-only baseline while aligning instantaneous dynamics with cumulative displacement.
% Consequently, it generates motion patterns that more closely match the combined temporal and spatial characteristics of the validation data.
% }

\joshuav{ \textbf{Peak + Collective.} 
Conditioning on both Peak and Collective metrics produces the most balanced behavior and the highest generative quality.
While the individual components struggle to maintain distributional fidelity ($FID \approx 0.12$), the combined EMA model achieves a significantly superior FID of 0.056. 
Furthermore, it yields the most stable structural control with a Collective MAE of 1.574, effectively aligning instantaneous dynamics with cumulative displacement. 
These results suggest that joint conditioning is not only necessary for precise effort modulation but also serves as a critical regularizer for generating natural, high-fidelity motion. }

%---------------------------------------------------------------------------------
\joshua{\subsection{User Study}}
\label{sec:user_study_results}
%---------------------------------------------------------------------------------

\joshua{
To evaluate the controllability and quality of the proposed effort parameters, we conducted a user study with 21 participants (16 males, 5 females, mean age 39.5 years, located in the US or the UK) recruited via Prolific. 
We hypothesized that the intended algorithmic physical effort is positively correlated with \textit{Perceived Effort} ($H_1$).
Second, we hypothesized that increasing effort is negatively correlated with motion plausibility ($H_2$). Motion plausibility was operationalized as \textit{Perceived Humanness} of the movement, as opposed to perceived AI-generated level. }

%---------------------------------------------------------------------------------
%%%%% EXPERIMENTAL DESIGN

%----------------------------------------------------------------------------------------
\subsubsection{\joshua{Experimental Design}}
%----------------------------------------------------------------------------------------
\joshua{
The study followed a 7x14 within-subjects design with seven effort levels (0.7, 0.8, 0.9, 1.0, 1.1, 1.2, 1.3) and 14 action classes. 
Each condition had three trials, generated with different random seeds, yielding 294 unique video stimuli for rating.
We evaluated EMA alone, without comparing it to other methods, because alternative methods modulate effort via natural-language adverbs, which would confound perceived physical effort with linguistic framing effects. 
Unlike prior work that relies on descriptive statistics to compare two methods, we conduct formal statistical testing to assess the relationship between intended effort encoded in the motion generation and perceived effort in the generated motion.
}

\joshua{
Participants answered one question for each video (``How is the motion?'') and then evaluated each stimulus using two Visual Analog Scales (VAS) to rate \textbf{Perceived Effort} with anchors ranging from Low to High effort; and \textbf{Perceived Humanness} with anchors ranging from Artificially Generated to Completely Human-like.
}

%---------------------------------------------------------------------------------
\subsubsection{\joshua{Analysis and Results}}
\begin{figure}[h]
    \centering
    % --- LEFT PLOT ---
    \begin{subfigure}[b]{0.45\linewidth}
        \centering
        % Width is relative to the subfigure, not the page
        \includegraphics[width=\linewidth]{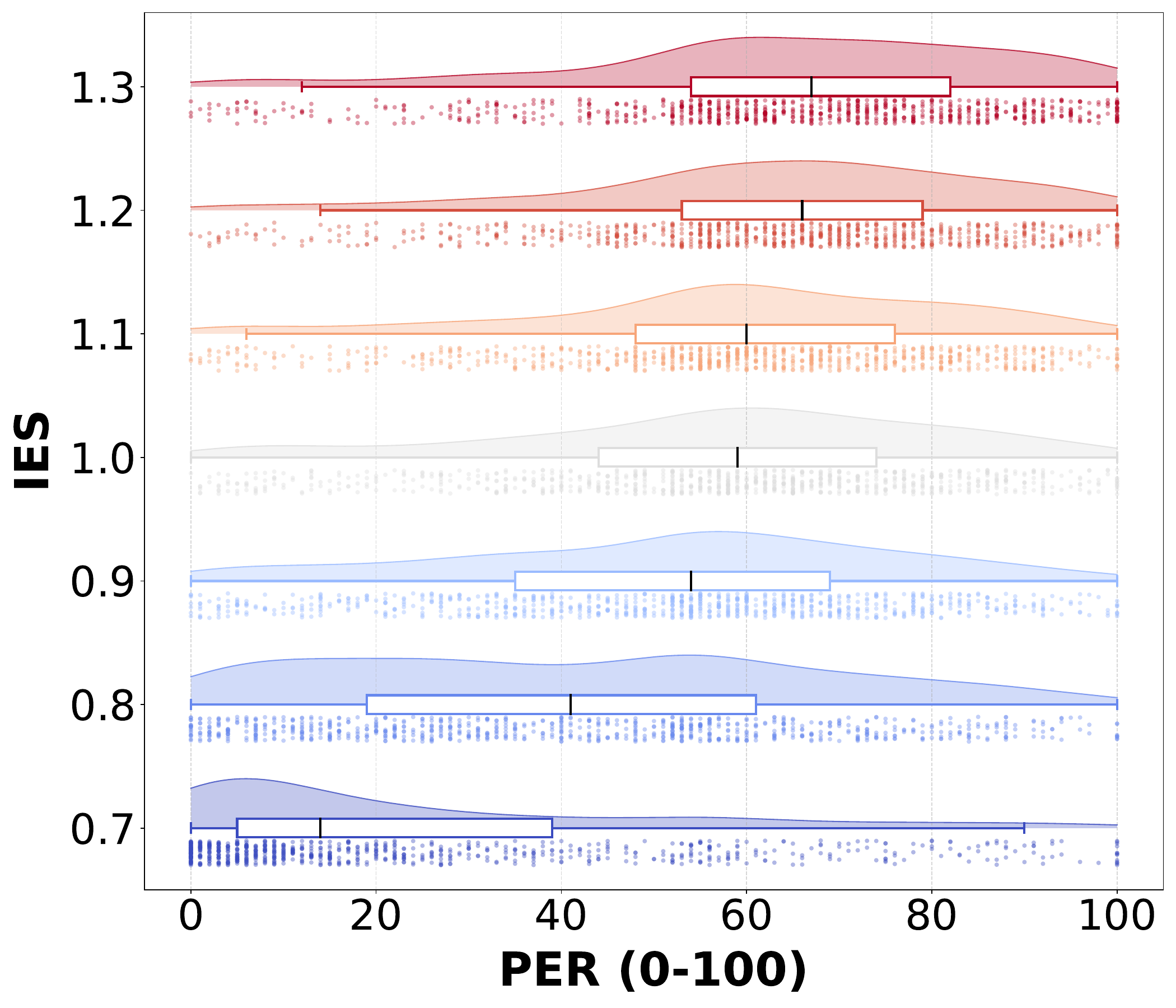}
        \caption{Perceived Effort} % Short caption for the subfigure
        \label{fig:speed_plot}
    \end{subfigure}
    \hfill % Adds space between the two images
    % --- RIGHT PLOT ---
    \begin{subfigure}[b]{0.45\linewidth}
        \centering
        \includegraphics[width=\linewidth]{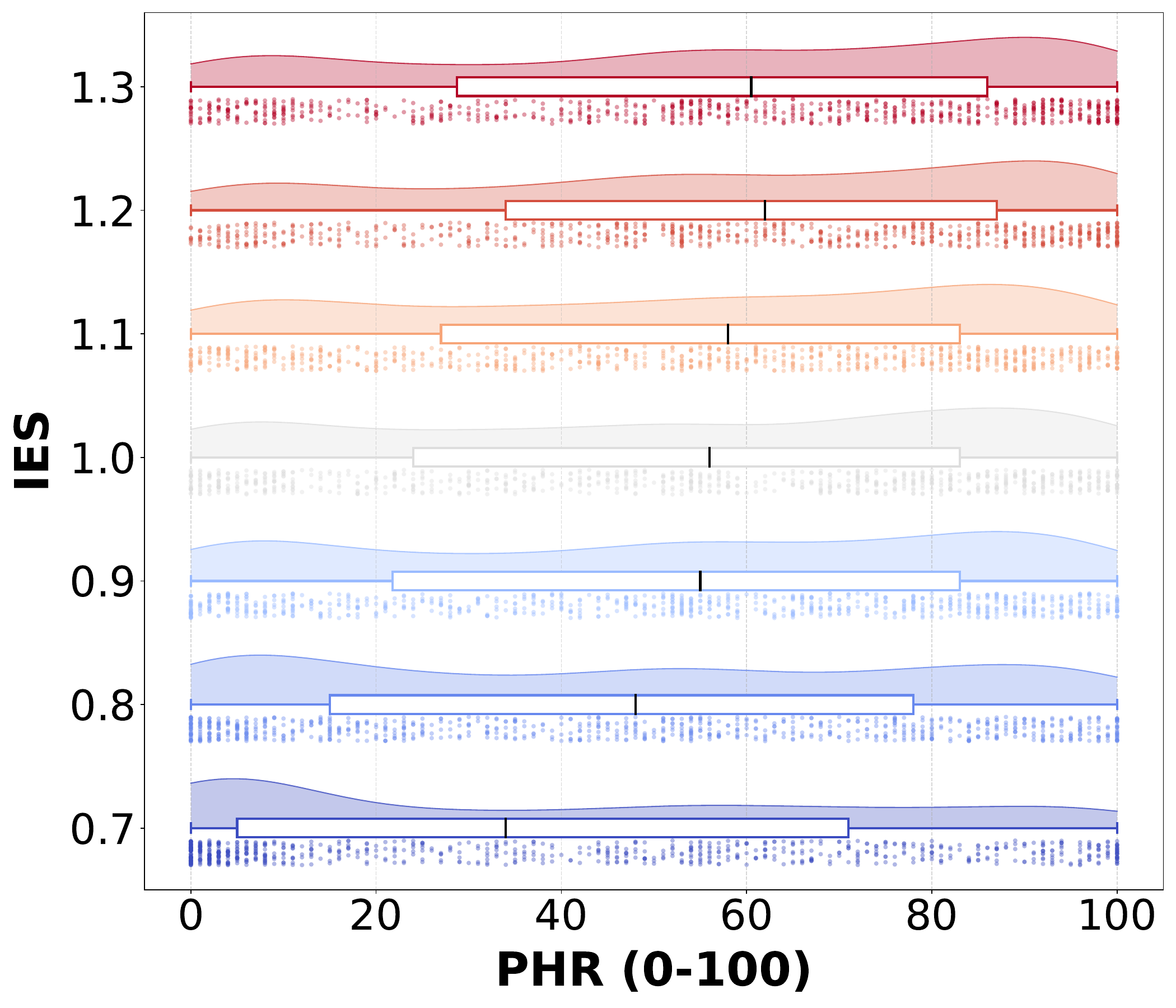}
        \caption{Perceived Humanness} % Short caption for the subfigure
        \label{fig:humanness_plot}
    \end{subfigure}
    
    \caption{\joshua{User study results ($N=21$).
    Raincloud plots \cite{allen2019raincloud} visualize raw responses (rain), boxplot statistics (umbrella), and probability densities (cloud) for the Intended Effort Scale (IES).
    (a) Perceived Effort Ratings (PER) are positively correlated with the effort parameter.
    (b) Perceived Humanness Ratings (PHR) exhibit a stable plausibility plateau across the functional effort range.}}
    \label{fig:user_study_plots}
\end{figure}

%---------------------------------------------------------------------------------
\paragraph{\joshua{$H_1$: Control of Perceived Effort}}
%---------------------------------------------------------------------------------x
\joshua{
A Spearman's rank correlation analysis confirmed a significant positive relationship between intended and perceived effort ($\rho = 0.41, p < 0.001$), supporting $H_1$.
Increasing the effort parameter consistently shifts the probability density distributions toward higher scores (Fig.~\ref{fig:user_study_plots}a).
}
%---------------------------------------------------------------------------------
\paragraph{\joshua{$H_2$: Perceived Humanness}}
%---------------------------------------------------------------------------------
\joshua{
A Spearman's rank correlation analysis revealed a weak positive correlation ($\rho = 0.16, p < 0.001$) between intended effort and Perceived Humanness, contrary to our expectation ($H_2$). We predicted that higher effort would degrade plausibility, but we observe a \textit{plausibility plateau}: while the artificially slow condition ($0.7\times$) received lower ratings, all functional conditions ($0.9\times$--$1.3\times$) clustered within a stable range (approx. 45--60)(Fig.~\ref{fig:user_study_plots}b). Therefore, EMA maintains human-like motion quality across the target effort range.
}

\begin{table*}[t]
\caption{\joshua{Results on standard metrics. Lower is better for Fréchet Inception Distance (FID), higher is better for the others. We report EMA after the 500th epoch of training.}}
\centering
\begin{tabular}{l c c c c c c}
\toprule
\textbf{Model}
& \textbf{FID} $\downarrow$
& \textbf{Diversity} $\uparrow$
& \textbf{Top-1} $\uparrow$
& \textbf{Top-2} $\uparrow$
& \textbf{Top-3} $\uparrow$
& \textbf{Multi-Modality} $\uparrow$ \\
\midrule

EMA
& $\textbf{0.056} \pm 0.002$
& $9.410 \pm 0.086$
& $0.517 \pm 0.002$
& $0.711 \pm 0.002$
& $0.808 \pm 0.001$
& $1.579 \pm 0.057$ \\

EMA (Peak-only)
& $0.128 \pm 0.003$
& $9.458 \pm 0.069$
& $0.530 \pm 0.002$
& $0.725 \pm 0.001$
& $0.821 \pm 0.001$
& $1.643 \pm 0.058$ \\

EMA (Collective-only)
& $0.124 \pm 0.002$
& $9.517 \pm 0.081$
& $0.531 \pm 0.001$
& $0.728 \pm 0.001$
& $0.823 \pm 0.001$
& $1.617 \pm 0.056$ \\

SALAD
& $0.115 \pm 0.003$
& $\textbf{9.586} \pm 0.070$
& $\textbf{0.553} \pm 0.001$
& $\textbf{0.749} \pm 0.001$
& $\textbf{0.842} \pm 0.001$
& $\textbf{1.869} \pm 0.069$ \\
\bottomrule
\end{tabular}
\label{tab:dist_metrics}
\end{table*}
%---------------------------------------------------------------------------------
\joshua{\subsection{Discussion and limitations}}
%---------------------------------------------------------------------------------

\joshua{
The EMA denoiser enables direct numerical control of motion by injecting LMA-inspired effort metrics into the \textit{MetricAttn} layer, allowing single-pass modulation of body-part dynamics.
Unlike global conditioning, EMA distributes effort locally via region-identity embeddings, thereby reducing unintended distortions in non-target body parts.
Additional comparisons between global and region-aware embeddings are provided in the supplementary material.
}

% \joshua{
% EMA exhibits a negative performance gap versus SALAD on standard metrics, while preserving multimodality, shown in Table~\ref{tab:dist_metrics}.
% This stems from EMA's trade-off in jointly optimizing textual alignment and numerical effort control, rather than text-motion alignment alone.
% As a result, EMA does not fully exploit dataset biases toward moderate-effort motions that favor lower FID and higher R-precision. Application-wise, this controllability outweighs minor FID gains, enabling user customization via data augmentation for out-of-distribution intensities.
% }

\joshuav{ As shown in Table~\ref{tab:dist_metrics}, EMA achieves a significantly lower FID (0.056) than the SALAD baseline (0.115), effectively matching the ground-truth distribution. The slight reduction in Diversity, R-precision (Top-1, Top-2, Top-3) and Multi-Modality reflects the model's shift from purely semantic-driven generation to a multi-objective optimization that includes numerical effort. While SALAD relies on textual biases within the dataset, EMA prioritizes the structural constraints of the effort metrics, a necessary trade-off for achieving the fine-grained controllability demonstrated in our qualitative results. }

To capture controllability beyond standard metrics, we introduced Metric-to-Motion Consistency and Body-Part Modulation as complementary evaluations.
On these tasks, EMA substantially outperformed SALAD in peak change control. Also, increasing effort scalars yielded near-monotonic scaling of peak displacements, while SALAD exhibited irregular responses.
This highlights that effort-based control is not a substitute for text-driven generation but a complementary capability, motivating new evaluation protocols that assess both fidelity and controllability.

Finally, EMA demonstrates strong control over peak positional changes \joshuav{(MAE $\approx$ 0.0597), but exhibits higher errors for collective changes (MAE $\approx$ 1.574)} due to cumulative per-frame deviations shown in Table~\ref{tab:merged_evaluation}.
Although trend plots (Fig.~\ref{fig:trend_compact}) confirm near-monotonic alignment with effort scalars, these results highlight a limitation in modeling long-horizon cumulative effects.
This behavior reflects the inherent bias of diffusion models toward local patterns, suggesting that improved loss formulations or trajectory-level constraints may be required for more accurate collective control.
\joshuav{Additionally, the user study is limited by a modest sample size (N=21); while the within-subjects design yielded 294 ratings per participant and both hypotheses were supported with high significance (p $<$ 0.001), future work should replicate findings with larger cohorts.}
Additional trend analyses are provided in the supplementary material.

%---------------------------------------------------------------------------------
\section{Conclusion}
%---------------------------------------------------------------------------------
We proposed Effort Metric Attention (EMA), a novel module for motion latent diffusion models that encodes quantified effort metrics inspired by Laban Movement Analysis.
By conditioning on peak and collective joint changes—corresponding to LMA’s \textit{Time} and \textit{Weight}—EMA enables interpretable, fine-grained, and single-pass intensity control, overcoming the limitations of adverb-based approaches such as SALAD~\cite{hong2025salad}.

Experiments show that EMA achieves metric-to-motion consistency while preserving semantic fidelity, with low peak errors, and supports localized modulation.
\joshuav{While EMA achieves superior FID (0.056) compared to the baseline (0.115), standard metrics like R-precision still do not capture controllability, underscoring the need for effort-specific evaluation.}
Limitations include difficulty in regulating collective changes and the underrepresentation of extreme motions in HumanML3D.

Future work should enhance supervision strategies, diversify datasets, and integrate EMA with real-time systems~\cite{tevet2025closd,wang2024move}.
Applications span virtual avatars~\cite{zhao2022metaverse}, video games~\cite{park2023generativeagentsinteractivesimulacra,kim_vtuber_2025}, and communication robots~\cite{serifi2024robotmdm,belpeame_guidelines_2018,chaudhry_user_2024}, where effort-aware control can enrich expressiveness and adaptability.

%%%%%%%%%%%%%%%%%%%%%%%%%%%%%%%%%%%%%%%%%%%%%%%%%%%%%%%%%%%%%%%%%%%%%%%%%%%%%%%%
\section{Acknowledgements}
M.P-H. was supported by JSPS KAKENHI 25K21250 and RIEC Nation-Wide Cooperative Research Project R05/A33.

%%%%%%%%%%%%%%%%%%%%%%%%%%%%%%%%%%%%%%%%%%%%%%%%%%%%%%%%%%%%%%%%%%%%%%%%%%%%%%%%

%%%%%%%%%%%%%%%%%%%%%%%%%%%%%%%%%%%%%%%%%%%%%%%%%%%%%%%%%%%%%%%%%%%%%%%%%%%%%%%%
\section*{Ethical Impact Statement}

EffortMetricAttention (EMA) generates expressive 3D human motions using the anonymized HumanML3D dataset, augmented via interpolation and frame reduction. 
Human participants were involved in this study to rate videos anonymously to create our data and analysis. 
Ethical approval\joshuav{(2025-I-44)} was obtained to run this study, and participants provided informed consent before the experiment.
The sample size used in this study is relatively small and is limited to raters in the USA and the UK. Furthermore, it is skewed to male raters. Future work should investigate the perceptual validity of EMA in other populations.

%%%%%%%%%%%%%%%%%%%%%%%%%%%%%%%%%%%%%%%%%%%%%%%%%%%%%%%%%%%%%%%%%%%%%%%%%%%%%%%%

{\small
\bibliographystyle{ieee}
\bibliography{references}

@misc{tevet2022human,
  author = {Tevet, Guy and Raab, Sigal and Gordon, Brian and Shafir, Yonatan and Cohen-Or, Daniel and Bermano, Amit H.},
  title = {Human Motion Diffusion Model},
  year = {2022},
  eprint = {2209.14916},
  archivePrefix = {arXiv},
  primaryClass = {cs.CV},
  url = {https://arxiv.org/abs/2209.14916}
}

@misc{tashakori2025flexmotion,
  author = {Tashakori, Arvin and Tashakori, Arash and Yang, Gongbo and Wang, Z. Jane},
  title = {FlexMotion: Lightweight, Physics-Aware, and Controllable Human Motion Generation},
  year = {2025},
  eprint = {2501.16778},
  archivePrefix = {arXiv},
  primaryClass = {cs.AI},
  url = {https://arxiv.org/abs/2501.16778}
}

@misc{hong2025salad,
  author = {Hong, S. and Kim, C. and Yoon, S. and Nam, J. and Cha, S. and Noh, J.},
  title = {SALAD: Skeleton-aware Latent Diffusion for Text-driven Motion Generation and Editing},
  year = {2025},
  eprint = {2503.13836},
  archivePrefix = {arXiv},
  primaryClass = {cs.CV},
  url = {https://arxiv.org/abs/2503.13836}
}

@inproceedings{petrovich2022temos,
  author = {Petrovich, Mathis and Black, Michael J. and Varol, Gül},
  title = {TEMOS: Generating Diverse Human Motions from Textual Descriptions},
  booktitle = {Computer Vision – ECCV 2022: 17th European Conference, Tel Aviv, Israel, October 23–27, 2022, Proceedings, Part XXII},
  year = {2022},
  pages = {480--497},
  doi = {10.1007/978-3-031-20047-2_28},
  url = {https://link.springer.com/chapter/10.1007/978-3-031-20047-2_28},
  eprint = {2204.14109},
  archivePrefix = {arXiv}
}

@misc{huang2024mvadapter,
  author = {Huang, Zehuan and Guo, Yuan-Chen and Wang, Haoran and Yi, Ran and Ma, Lizhuang and Cao, Yan-Pei and Sheng, Lu},
  title = {MV-Adapter: Multi-view Consistent Image Generation Made Easy},
  year = {2024},
  eprint = {2412.03632},
  archivePrefix = {arXiv},
  primaryClass = {cs.CV},
  url = {https://arxiv.org/abs/2412.03632}
}

@article{he2024mojito,
  title   = {Mojito: Motion Trajectory and Intensity Control for Video Generation},
  author  = {Xuehai He and Shuohang Wang and Jianwei Yang and Xiaoxia Wu and Yiping Wang and Kuan Wang and Zheng Zhan and Olatunji Ruwase and Yelong Shen and Xin Eric Wang},
  year    = {2024},
  journal = {arXiv preprint arXiv: 2412.08948}
}

@misc{ho2020denoising,
  author = {Ho, Jonathan and Jain, Ajay and Abbeel, Pieter},
  title = {Denoising Diffusion Probabilistic Models},
  year = {2020},
  eprint = {2006.11239},
  archivePrefix = {arXiv},
  primaryClass = {cs.LG},
  url = {https://arxiv.org/abs/2006.11239}
}

@article{aristidou2015folk,
  author = {Aristidou, Andreas and Stavrakis, Efstathios and Charalambous, Panayiotis and Chrysanthou, Yiorgos and Himona, Stephania Loizidou},
  title = {Folk Dance Evaluation Using Laban Movement Analysis},
  journal = {Journal on Computing and Cultural Heritage},
  year = {2015},
  volume = {8},
  number = {4},
  pages = {1--19},
  doi = {10.1145/2755566},
  url = {https://dl.acm.org/doi/10.1145/2755566}
}

@inproceedings{zacharatos2013emotion,
  author = {Zacharatos, Haris and Gatzoulis, Christos and Chrysanthou, Yiorgos and Aristidou, Andreas},
  title = {Emotion Recognition for Exergames using Laban Movement Analysis},
  booktitle = {Proceedings of Motion on Games},
  year = {2013},
  pages = {61--66},
  doi = {10.1145/2522628.2522651},
  url = {https://doi.org/10.1145/2522628.2522651}
}

@inproceedings{guo2020action2motion,
  author = {Guo, Chuan and Zuo, Xinxin and Wang, Sen and Zou, Shihao and Sun, Qingyao and Deng, Annan and Gong, Minglun and Cheng, Li},
  title = {Action2Motion: Conditioned Generation of 3D Human Motions},
  booktitle = {Proceedings of the 28th ACM International Conference on Multimedia},
  year = {2020},
  doi = {10.1145/3394171.3413635},
  url = {https://arxiv.org/abs/2007.15240},
  eprint = {2007.15240},
  archivePrefix = {arXiv}
}

@misc{zhang2023t2mgpt,
  author = {Zhang, Jianrong and Zhang, Yangsong and Cun, Xiaodong and Huang, Shaoli and Zhang, Yong and Zhao, Hongwei and Lu, Hongtao and Shen, Xi},
  title = {T2M-GPT: Generating Human Motion from Textual Descriptions with Discrete Representations},
  year = {2023},
  eprint = {2301.06052},
  archivePrefix = {arXiv},
  primaryClass = {cs.CV},
  url = {https://arxiv.org/abs/2301.06052}
}

@misc{wang2024move,
  author = {Wang, Zan and Chen, Yixin and Jia, Baoxiong and Li, Puhao and Zhang, Jinlu and Zhang, Jingze and Liu, Tengyu and Zhu, Yixin and Liang, Wei and Huang, Siyuan},
  title = {Move as You Say, Interact as You Can: Language-guided Human Motion Generation with Scene Affordance},
  year = {2024},
  eprint = {2403.18036},
  archivePrefix = {arXiv},
  primaryClass = {cs.CV},
  url = {https://arxiv.org/abs/2403.18036}
}

@misc{wan2024tlcontrol,
  author = {Wan, Weilin and Dou, Zhiyang and Komura, Taku and Wang, Wenping and Jayaraman, Dinesh and Liu, Lingjie},
  title = {TLControl: Trajectory and Language Control for Human Motion Synthesis},
  year = {2024},
  eprint = {2311.17135},
  archivePrefix = {arXiv},
  primaryClass = {cs.CV},
  url = {https://arxiv.org/abs/2311.17135}
}

@inproceedings{guo2022generating,
  author = {Guo, Chuan and Zou, Shihao and Zuo, Xinxin and Wang, Sen and Ji, Wei and Li, Xingyu and Cheng, Li},
  title = {Generating Diverse and Natural 3D Human Motions From Text},
  booktitle = {Proceedings of the IEEE/CVF Conference on Computer Vision and Pattern Recognition (CVPR)},
  year = {2022},
  pages = {5152--5161},
  url = {https://openaccess.thecvf.com/content/CVPR2022/html/Guo_Generating_Diverse_and_Natural_3D_Human_Motions_From_Text_CVPR_2022_paper.html}
}

@misc{rombach2022high,
  author = {Rombach, Robin and Blattmann, Andreas and Lorenz, Dominik and Esser, Patrick and Ommer, Björn},
  title = {High-Resolution Image Synthesis with Latent Diffusion Models},
  year = {2022},
  eprint = {2112.10752},
  archivePrefix = {arXiv},
  primaryClass = {cs.CV},
  url = {https://arxiv.org/abs/2112.10752}
}

@inproceedings{chen2023executing,
  title     = {Executing your Commands via Motion Diffusion in Latent Space},
  author    = {Chen, Xin and Jiang, Biao and Liu, Wen and Huang, Zilong and Fu, Bin and Chen, Tao and Yu, Gang},
  booktitle = {Proceedings of the IEEE/CVF Conference on Computer Vision and Pattern Recognition},
  pages     = {18000--18010},
  year      = {2023},
}

@misc{radford2021learning,
  author = {Radford, Alec and Kim, Jong Wook and Hallacy, Chris and Ramesh, Aditya and Goh, Gabriel and Agarwal, Sandhini and Sastry, Girish and Askell, Amanda and Mishkin, Pamela and Clark, Jack and Krueger, Gretchen and Sutskever, Ilya},
  title = {Learning Transferable Visual Models From Natural Language Supervision},
  year = {2021},
  eprint = {2103.00020},
  archivePrefix = {arXiv},
  primaryClass = {cs.CV},
  url = {https://arxiv.org/abs/2103.00020}
}

@misc{yuan2022physdiff,
  author = {Yuan, Ye and Song, Jiaming and Iqbal, Umar and Vahdat, Arash and Kautz, Jan},
  title = {PhysDiff: Physics-Guided Human Motion Diffusion Model},
  year = {2022},
  eprint = {2212.02500},
  archivePrefix = {arXiv},
  primaryClass = {cs.CV},
  url = {https://arxiv.org/abs/2212.02500}
}

@misc{zhang2022motiondiffuse,
  author = {Zhang, Mingyuan and Cai, Zhongang and Pan, Liang and Hong, Fangzhou and Guo, Xinying and Yang, Lei and Liu, Ziwei},
  title = {MotionDiffuse: Text-Driven Human Motion Generation with Diffusion Model},
  year = {2022},
  eprint = {2208.15001},
  archivePrefix = {arXiv},
  primaryClass = {cs.CV},
  url = {https://arxiv.org/abs/2208.15001}
}

@article{zhao2022metaverse,
  author = {Zhao, Yuheng and Jiang, Jinjing and Chen, Yi and Liu, Richen and Yang, Yalong and Xue, Xiangyang and Chen, Siming},
  title = {Metaverse: Perspectives from Graphics, Interactions and Visualization},
  journal = {Visual Informatics},
  year = {2022},
  volume = {6},
  number = {1},
  pages = {56--67},
  doi = {10.1016/j.visinf.2022.03.002},
  url = {https://www.sciencedirect.com/science/article/pii/S2468502X22000045}
}

@misc{yao2023moconvq,
  author = {Yao, Heyuan and Song, Zhenhua and Zhou, Yuyang and Ao, Tenglong and Chen, Baoquan and Liu, Libin},
  title = {MoConVQ: Unified Physics-Based Motion Control via Scalable Discrete Representations},
  year = {2023},
  eprint = {2310.14307},
  archivePrefix = {arXiv},
  primaryClass = {cs.CV},
  url = {https://arxiv.org/abs/2310.14307}
}

@misc{ho2022classifierfreediffusionguidance,
      title={Classifier-Free Diffusion Guidance}, 
      author={Jonathan Ho and Tim Salimans},
      year={2022},
      eprint={2207.12598},
      archivePrefix={arXiv},
      primaryClass={cs.LG},
      url={https://arxiv.org/abs/2207.12598}, 
}

@misc{salimans2022progressivedistillationfastsampling,
      title={Progressive Distillation for Fast Sampling of Diffusion Models}, 
      author={Tim Salimans and Jonathan Ho},
      year={2022},
      eprint={2202.00512},
      archivePrefix={arXiv},
      primaryClass={cs.LG},
      url={https://arxiv.org/abs/2202.00512}, 
}

@inproceedings{
		tevet2025closd,
		title={{CL}o{SD}: Closing the Loop between Simulation and Diffusion for multi-task character control},
		author={Guy Tevet and Sigal Raab and Setareh Cohan and Daniele Reda and Zhengyi Luo and Xue Bin Peng and Amit Haim Bermano and Michiel van de Panne},
		booktitle={The Thirteenth International Conference on Learning Representations},
		year={2025},
		url={https://openreview.net/forum?id=pZISppZSTv}
}

@misc{park2023generativeagentsinteractivesimulacra,
      title={Generative Agents: Interactive Simulacra of Human Behavior}, 
      author={Joon Sung Park and Joseph C. O'Brien and Carrie J. Cai and Meredith Ringel Morris and Percy Liang and Michael S. Bernstein},
      year={2023},
      eprint={2304.03442},
      archivePrefix={arXiv},
      primaryClass={cs.HC},
      url={https://arxiv.org/abs/2304.03442}, 
}

@misc{devlin2019bertpretrainingdeepbidirectional,
      title={BERT: Pre-training of Deep Bidirectional Transformers for Language Understanding}, 
      author={Jacob Devlin and Ming-Wei Chang and Kenton Lee and Kristina Toutanova},
      year={2019},
      eprint={1810.04805},
      archivePrefix={arXiv},
      primaryClass={cs.CL},
      url={https://arxiv.org/abs/1810.04805}, 
}

@inproceedings{serifi2024robotmdm,
author = {Serifi, Agon and Grandia, Ruben and Knoop, Espen and Gross, Markus and B\"{a}cher, Moritz},
title = {Robot Motion Diffusion Model: Motion Generation for Robotic Characters},
year = {2024},
isbn = {9798400711312},
publisher = {Association for Computing Machinery},
address = {New York, NY, USA},
url = {https://doi.org/10.1145/3680528.3687626},
doi = {10.1145/3680528.3687626},
booktitle = {SIGGRAPH Asia 2024 Conference Papers},
articleno = {50},
}

@article{belpeame_guidelines_2018,
author = {Belpaeme, Tony and Vogt, Paul and van den Berghe, Rianne and Bergmann, Kirsten and Goksun, Tilbe and de Haas, Mirjam and Kanero, Junko and Kennedy, James and Küntay, Aylin and Oudgenoeg-Paz, Ora and Papadopoulos, Fotios and Schodde, Thorsten and Verhagen, Josje and Wallbridge, Christopher and Willemsen, Bram and de Wit, Jan and Geckin, Vasfiye and Kunold Neé Hoffmann, Laura and Kopp, Stefan and Pandey, Amit Kumar},
year = {2018},
month = {06},
pages = {},
title = {Guidelines for Designing Social Robots as Second Language Tutors},
volume = {10},
journal = {International Journal of Social Robotics},
doi = {10.1007/s12369-018-0467-6}
}

@article{chaudhry_user_2024,
	title = {User perceptions and experiences of an {AI}-driven conversational agent for mental health support},
	volume = {10},
	issn = {2306-9740},
	url = {https://www.ncbi.nlm.nih.gov/pmc/articles/PMC11304096/},
	doi = {10.21037/mhealth-23-55},
	urldate = {2025-09-10},
	journal = {mHealth},
	author = {Chaudhry, Beenish Moalla and Debi, Happy Rani},
	month = jul,
	year = {2024},
	pmid = {39114462},
	pmcid = {PMC11304096},
	pages = {22},
}

@inproceedings{kim_vtuber_2025, series={CHI ’25},
   title={VTuber’s Atelier: The Design Space, Challenges, and Opportunities for VTubing},
   url={http://dx.doi.org/10.1145/3706598.3714107},
   DOI={10.1145/3706598.3714107},
   booktitle={Proceedings of the 2025 CHI Conference on Human Factors in Computing Systems},
   publisher={ACM},
   author={Kim, Daye and Lee, Sebin and Jun, Yoonseo and Shin, Yujin and Lee, Jungjin},
   year={2025},
   month=apr, pages={1–23},
   collection={CHI ’25} }

@misc{karunratanakul2023guidedmotiondiffusioncontrollable,
      title={Guided Motion Diffusion for Controllable Human Motion Synthesis}, 
      author={Korrawe Karunratanakul and Konpat Preechakul and Supasorn Suwajanakorn and Siyu Tang},
      year={2023},
      eprint={2305.12577},
      archivePrefix={arXiv},
      primaryClass={cs.CV},
      url={https://arxiv.org/abs/2305.12577}, 
}

@article{sawdayee2025dance,
  title={Dance Like a Chicken: Low-Rank Stylization for Human Motion Diffusion},
  author={Sawdayee, Haim and Guo, Chuan and Tevet, Guy and Zhou, Bing and Wang, Jian and Bermano, Amit H},
  journal={arXiv preprint arXiv:2503.19557},
  year={2025}
}

@article{Jang_motion_puzzle_2022,
   title={Motion Puzzle: Arbitrary Motion Style Transfer by Body Part},
   volume={41},
   ISSN={1557-7368},
   url={http://dx.doi.org/10.1145/3516429},
   DOI={10.1145/3516429},
   number={3},
   journal={ACM Transactions on Graphics},
   publisher={Association for Computing Machinery (ACM)},
   author={Jang, Deok-Kyeong and Park, Soomin and Lee, Sung-Hee},
   year={2022},
   month=jun, pages={1–16} }

@inproceedings{habibie2017recurrent,
author = {Habibie, Ikhsanul and Holden, Daniel and Schwarz, Jonathan and Yearsley, Joe and Komura, Taku},
year = {2017},
month = {01},
pages = {},
title = {A Recurrent Variational Autoencoder for Human Motion Synthesis},
doi = {10.5244/C.31.119}
}

@article{park_2021_STGraph,
author = {Park, Soomin and Jang, Deok-Kyeong and Lee, Sung-Hee},
title = {Diverse Motion Stylization for Multiple Style Domains via Spatial-Temporal Graph-Based Generative Model},
year = {2021},
issue_date = {September 2021},
publisher = {Association for Computing Machinery},
address = {New York, NY, USA},
volume = {4},
number = {3},
url = {https://doi.org/10.1145/3480145},
doi = {10.1145/3480145},
month = sep,
articleno = {36},
numpages = {17}
}

@misc{meng2022sdeditguidedimagesynthesis,
      title={SDEdit: Guided Image Synthesis and Editing with Stochastic Differential Equations}, 
      author={Chenlin Meng and Yutong He and Yang Song and Jiaming Song and Jiajun Wu and Jun-Yan Zhu and Stefano Ermon},
      year={2022},
      eprint={2108.01073},
      archivePrefix={arXiv},
      primaryClass={cs.CV},
      url={https://arxiv.org/abs/2108.01073}, 
}

@misc{hertz2022prompttopromptimageeditingcross,
      title={Prompt-to-Prompt Image Editing with Cross Attention Control}, 
      author={Amir Hertz and Ron Mokady and Jay Tenenbaum and Kfir Aberman and Yael Pritch and Daniel Cohen-Or},
      year={2022},
      eprint={2208.01626},
      archivePrefix={arXiv},
      primaryClass={cs.CV},
      url={https://arxiv.org/abs/2208.01626}, 
}

@misc{huang2024comocontrollablemotiongeneration,
      title={CoMo: Controllable Motion Generation through Language Guided Pose Code Editing}, 
      author={Yiming Huang and Weilin Wan and Yue Yang and Chris Callison-Burch and Mark Yatskar and Lingjie Liu},
      year={2024},
      eprint={2403.13900},
      archivePrefix={arXiv},
      primaryClass={cs.CV},
      url={https://arxiv.org/abs/2403.13900}, 
}

@misc{shi2024fgmdmzeroshothumanmotion,
      title={FG-MDM: Towards Zero-Shot Human Motion Generation via ChatGPT-Refined Descriptions}, 
      author={Xu Shi and Wei Yao and Chuanchen Luo and Junran Peng and Hongwen Zhang and Yunlian Sun},
      year={2024},
      eprint={2312.02772},
      archivePrefix={arXiv},
      primaryClass={cs.CV},
      url={https://arxiv.org/abs/2312.02772}, 
}

@misc{fan2025zerozeroshotmotiongeneration,
      title={Go to Zero: Towards Zero-shot Motion Generation with Million-scale Data}, 
      author={Ke Fan and Shunlin Lu and Minyue Dai and Runyi Yu and Lixing Xiao and Zhiyang Dou and Junting Dong and Lizhuang Ma and Jingbo Wang},
      year={2025},
      eprint={2507.07095},
      archivePrefix={arXiv},
      primaryClass={cs.CV},
      url={https://arxiv.org/abs/2507.07095}, 
}

@inproceedings{witkin1988spacetime,
author = {Witkin, Andrew and Kass, Michael},
title = {Spacetime constraints},
year = {1988},
isbn = {0897912756},
publisher = {Association for Computing Machinery},
address = {New York, NY, USA},
url = {https://doi.org/10.1145/54852.378507},
doi = {10.1145/54852.378507},
booktitle = {Proceedings of the 15th Annual Conference on Computer Graphics and Interactive Techniques},
pages = {159–168},
numpages = {10},
series = {SIGGRAPH '88}
}

@article{long2023wonder3d,
  title={Wonder3D: Single Image to 3D using Cross-Domain Diffusion},
  author={Long, Xiaoxiao and Guo, Yuan-Chen and Lin, Cheng and Liu, Yuan and Dou, Zhiyang and Liu, Lingjie and Ma, Yuexin and Zhang, Song-Hai and Habermann, Marc and Theobalt, Christian and others},
  journal={arXiv preprint arXiv:2310.15008},
  year={2023}
}

@misc{perez2017filmvisualreasoninggeneral,
      title={FiLM: Visual Reasoning with a General Conditioning Layer}, 
      author={Ethan Perez and Florian Strub and Harm de Vries and Vincent Dumoulin and Aaron Courville},
      year={2017},
      eprint={1709.07871},
      archivePrefix={arXiv},
      primaryClass={cs.CV},
      url={https://arxiv.org/abs/1709.07871}, 
}

@misc{ba2016layernormalization,
      title={Layer Normalization}, 
      author={Jimmy Lei Ba and Jamie Ryan Kiros and Geoffrey E. Hinton},
      year={2016},
      eprint={1607.06450},
      archivePrefix={arXiv},
      primaryClass={stat.ML},
      url={https://arxiv.org/abs/1607.06450}, 
}

@misc{he2015deepresiduallearningimage,
      title={Deep Residual Learning for Image Recognition}, 
      author={Kaiming He and Xiangyu Zhang and Shaoqing Ren and Jian Sun},
      year={2015},
      eprint={1512.03385},
      archivePrefix={arXiv},
      primaryClass={cs.CV},
      url={https://arxiv.org/abs/1512.03385}, 
}

@misc{song2022ddim,
      title={Denoising Diffusion Implicit Models}, 
      author={Jiaming Song and Chenlin Meng and Stefano Ermon},
      year={2022},
      eprint={2010.02502},
      archivePrefix={arXiv},
      primaryClass={cs.LG},
      url={https://arxiv.org/abs/2010.02502}, 
}

@inproceedings{pan_romo_2024, series={SA ’24},
   title={RoMo: A Robust Solver for Full-body Unlabeled Optical Motion Capture},
   url={http://dx.doi.org/10.1145/3680528.3687615},
   DOI={10.1145/3680528.3687615},
   booktitle={SIGGRAPH Asia 2024 Conference Papers},
   publisher={ACM},
   author={Pan, Xiaoyu and Zheng, Bowen and Jiang, Xinwei and Zeng, Zijiao and Kou, Qilong and Wang, He and Jin, Xiaogang},
   year={2024},
   month=dec, pages={1–11},
   collection={SA ’24} }

@article{kim_comprehensive_2025,
	title = {A comprehensive survey of deep learning for time series forecasting: architectural diversity and open challenges},
	volume = {58},
	issn = {1573-7462},
	shorttitle = {A comprehensive survey of deep learning for time series forecasting},
	url = {https://doi.org/10.1007/s10462-025-11223-9},
	doi = {10.1007/s10462-025-11223-9},
	language = {en},
	number = {7},
	urldate = {2025-09-12},
	journal = {Artificial Intelligence Review},
	author = {Kim, Jongseon and Kim, Hyungjoon and Kim, HyunGi and Lee, Dongjun and Yoon, Sungroh},
	month = apr,
	year = {2025},
	pages = {216},
}

@misc{voas_what_2023,
    title = {What is the {Best} {Automated} {Metric} for {Text} to {Motion} {Generation}?},
    author = {Voas, Jordan and Wang, Yili and Huang, Qixing and Mooney, Raymond},
    url = {http://arxiv.org/abs/2309.10248},
    doi = {10.1145/3588432.3591550},
    month = sep,
    year = {2023},
    note = {arXiv:2309.10248 [cs]},
}

@article{mehrabian1967Decoding,
  author = {A. Mehrabian and M. Wiener},
  title = {Decoding of Inconsistent Communications},
  journal = {Journal of Personality and Social Psychology},
  volume = {6},
  number = {1},
  pages = {109--114},
  year = {1967},
  doi = {10.1037/h0024532},
  publisher = {American Psychological Association}
}

@misc{li2022hybrikhybridanalyticalneuralinverse,
      title={HybrIK: A Hybrid Analytical-Neural Inverse Kinematics Solution for 3D Human Pose and Shape Estimation}, 
      author={Jiefeng Li and Chao Xu and Zhicun Chen and Siyuan Bian and Lixin Yang and Cewu Lu},
      year={2022},
      eprint={2011.14672},
      archivePrefix={arXiv},
      primaryClass={cs.CV},
      url={https://arxiv.org/abs/2011.14672}, 
}

@misc{loshchilov2019decoupledweightdecayregularization,
      title={Decoupled Weight Decay Regularization}, 
      author={Ilya Loshchilov and Frank Hutter},
      year={2019},
      eprint={1711.05101},
      archivePrefix={arXiv},
      primaryClass={cs.LG},
      url={https://arxiv.org/abs/1711.05101}, 
}

@misc{turab2025dancestylerecognitionusing,
      title={Dance Style Recognition Using Laban Movement Analysis}, 
      author={Muhammad Turab and Philippe Colantoni and Damien Muselet and Alain Tremeau},
      year={2025},
      eprint={2504.21166},
      archivePrefix={arXiv},
      primaryClass={cs.CV},
      url={https://arxiv.org/abs/2504.21166}, 
}

@article{mckee1984,
  author = {McKee, S. P. and Nakayama, K.},
  title = {The detection of motion in the peripheral visual field},
  journal = {Vision Research},
  volume = {24},
  number = {1},
  pages = {25--32},
  year = {1984},
  doi = {10.1016/0042-6989(84)90040-8}
}

@article{werkhoven1992,
  author = {Werkhoven, P. and Snippe, H. P. and Toet, A.},
  title = {Visual processing of optic acceleration},
  journal = {Vision Research},
  volume = {32},
  number = {12},
  pages = {2313--2329},
  year = {1992},
  doi = {10.1016/0042-6989(92)90095-Z}
}

@misc{samadani2020affectivemovementgenerationusing,
      title={Affective Movement Generation using Laban Effort and Shape and Hidden Markov Models}, 
      author={Ali Samadani and Rob Gorbet and Dana Kulic},
      year={2020},
      eprint={2006.06071},
      archivePrefix={arXiv},
      primaryClass={cs.HC},
      url={https://arxiv.org/abs/2006.06071}, 
}

@article{allen2019raincloud,
  title={Raincloud plots: a multi-platform tool for robust data visualization},
  author={Allen, Micah and Poggiali, Davide and Whitaker, Kirstie and Marshall, Tom R and Kievit, Rogier A},
  journal={Wellcome open research},
  volume={4},
  year={2019},
  publisher={The Wellcome Trust}
}

@article{xie2024omnicontrol,
      title={OmniControl: Control Any Joint at Any Time for Human Motion Generation},
      author={Yiming Xie and Varun Jampani and Lei Zhong and Deqing Sun and Huaizu Jiang},
      booktitle={The Twelfth International Conference on Learning Representations},
      year={2024}
}

@article{zhong2024smoodistylizedmotiondiffusion,
      title={SMooDi: Stylized Motion Diffusion Model}, 
      author={Lei Zhong and Yiming Xie and Varun Jampani and Deqing Sun and Huaizu Jiang},
      year={2024},
      eprint={2407.12783},
      archivePrefix={arXiv},
      primaryClass={cs.CV},
      url={https://arxiv.org/abs/2407.12783}, 
}

@article{liang2024intergen,
  title={Intergen: Diffusion-based multi-human motion generation under complex interactions},
  author={Liang, Han and Zhang, Wenqian and Li, Wenxuan and Yu, Jingyi and Xu, Lan},
  journal={International Journal of Computer Vision},
  pages={1--21},
  year={2024},
  publisher={Springer}
}

@inproceedings{primal:iccv:2025,
  author = {Zhang, Yan and Feng, Yao and Cseke, Alpár and Saini, Nitin and Bajandas, Nathan and Heron, Nicolas and Black, Michael J.},  
   title = {{PRIMAL:} Physically Reactive and Interactive Motor Model for Avatar Learning},
  booktitle = {Proceedings of the IEEE/CVF International Conference on Computer Vision (ICCV)},
  month = oct,
  year = {2025}
}

@misc{raab2024monkeyseemonkeydo,
      title={Monkey See, Monkey Do: Harnessing Self-attention in Motion Diffusion for Zero-shot Motion Transfer}, 
      author={Sigal Raab and Inbar Gat and Nathan Sala and Guy Tevet and Rotem Shalev-Arkushin and Ohad Fried and Amit H. Bermano and Daniel Cohen-Or},
      year={2024},
      eprint={2406.06508},
      archivePrefix={arXiv},
      primaryClass={cs.CV},
      url={https://arxiv.org/abs/2406.06508}, 
}

@article{motionclr,
  title={MotionCLR: Motion Generation and Training-free Editing via Understanding Attention Mechanisms},
  author={Chen, Ling-Hao and Dai, Wenxun and Ju, Xuan and Lu, Shunlin and Zhang, Lei},
  journal={arxiv:2410.18977},
  year={2024}
}

@inproceedings{tao2022style,
  title={Style-ERD: Responsive and Coherent Online Motion Style Transfer},
  author={Tao, Tianxin and Zhan, Xiaohang and Chen, Zhongquan and van de Panne, Michiel},
  booktitle={Proceedings of the IEEE/CVF Conference on Computer Vision and Pattern Recognition},
  pages={6593--6603},
  year={2022}
}
}

% ===== SUPPLEMENTARY MATERIAL =====
\clearpage
\setcounter{section}{0}
\setcounter{figure}{0}
\setcounter{table}{0}
\renewcommand{\thesection}{S\arabic{section}}
\renewcommand{\thefigure}{S\arabic{figure}}
\renewcommand{\thetable}{S\arabic{table}}

\twocolumn[{
\begin{center}
{\LARGE \bf EMA: Effort Metric Attention for Anatomical Effort-Guided Human Motion Diffusion}\\[0.5em]
{\large\normalfont Supplementary Material}\\[0.5em]
\hrule
\end{center}
\vspace{1em}
}]

% \thispagestyle{fancy}
% \renewcommand{\headrulewidth}{0pt}
% \fancyhf{}
% \fancyhead[C]{2026 International Conference on Automatic Face and Gesture Recognition (FG)}
% \fancyfoot[L]{979-8-3315-7231-0/26/\$31.00 \copyright 2026 IEEE}

%----------------------------------------------------------------------------------------
\section{Section 3.D: Implementation Details and Design Rationale}
%----------------------------------------------------------------------------------------

%----------------------------------------------------------------------------------------
\subsection{Denoiser Architecture Overview}
%----------------------------------------------------------------------------------------
While the main paper describes the Effort Metric Attention (EMA) module in detail, this section provides a comprehensive overview of the full denoiser architecture for completeness.

\begin{figure*}[t]
\centering
\includegraphics[width=0.9\textwidth]{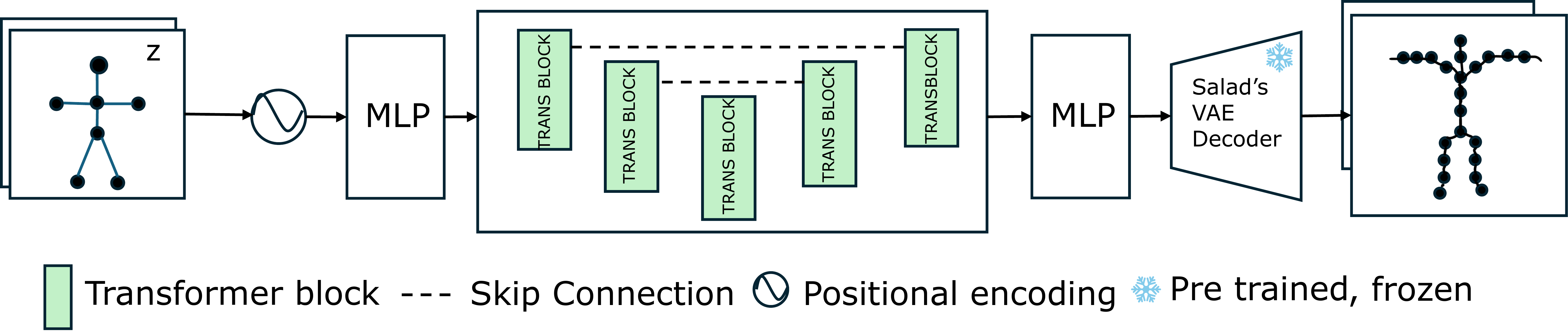}
\caption{Effort-guided skeleton-aware denoiser architecture. Motion latents encoded by SALAD’s frozen VAE are refined through sequential transformer blocks integrating temporal attention, skeletal attention, EMA, and text-based cross-attention. The denoiser predicts diffusion velocity conditioned on effort metrics and text prompts, enabling controllable motion synthesis during sampling.}
\label{fig:overview}
\end{figure*}

As illustrated in Fig.~\ref{fig:overview}, our model builds upon the skeleton-aware diffusion framework introduced in SALAD~\cite{hong2025salad}. 
Motion latents $\mathbf{z}$ are processed through $L$ stacked transformer layers.
In contrast to standard diffusion transformers, the proposed architecture explicitly decomposes motion representations into temporal and skeletal components before applying effort-based modulation at each layer.

%----------------------------------------------------------------------------------------
\subsection{Hierarchical Kinematic Conditioning}
%----------------------------------------------------------------------------------------
To justify the attention ordering adopted in the main model, we describe the hierarchical kinematic conditioning strategy that enforces the following strict processing order within every transformer block:
\emph{Temporal $\rightarrow$ Skeletal $\rightarrow$ EMA $\rightarrow$ Text Cross-Attention.}

\begin{enumerate}
    \item \textbf{Post-Spatio-Temporal Injection.}  
    EMA is applied strictly after both temporal and skeletal self-attention layers. 
    This design leverages the disentangled latent space to ensure that effort metrics modulate topologically coherent joint representations rather than geometrically unstructured features.
    By deferring effort conditioning until spatio-temporal structure is established, scalar energy signals are prevented from interfering with the formation of motion trajectories and inter-joint coordination patterns.
    
    \item \textbf{Pre-Text Injection.}  
    EMA is applied strictly before text cross-attention.
    This ensures that the text encoder operates on latent features that have already been modulated by effort characteristics.
    As a result, semantic guidance from text is conditioned on the physical energy state of the motion.
    Reversing this order would allow textual semantics to override effort constraints, potentially yielding motions that are semantically plausible but physically inconsistent with the specified effort metrics.
\end{enumerate}

Overall, this hierarchical design—progressing from structure to effort and finally to semantics—treats effort as a foundational physical constraint rather than a post-hoc adjustment, which is essential for achieving direct and stable numerical control.

%----------------------------------------------------------------------------------------
\subsection{Hyperparameter Configurations}
%----------------------------------------------------------------------------------------
To ensure reproducibility, Table~\ref{tab:supp_params} reports the architectural specifications and training hyperparameters used in our experiments, as implemented in our PyTorch framework.

\begin{table}[tb]
\centering
\caption{Architectural and training hyperparameters. We use the following hyperparameters to train our model.}
\label{tab:supp_params}
\resizebox{\columnwidth}{!}{
\begin{tabular}{l|c}
\toprule
\textbf{Parameter} & \textbf{Value} \\
\midrule
\multicolumn{2}{l}{\textit{Architecture Configuration}} \\
Latent Dimension ($D$) & 256 \\
Number of Attention Heads ($H$) & 8 \\
Number of Transformer Layers ($L$) & 5 \\
Effort Metric Dimension ($N_p$) & 2 (Peak, Collective) \\
Text Encoder & CLIP ViT-B/32 \\
\midrule
\multicolumn{2}{l}{\textit{Optimization and Diffusion}} \\
Prediction Type & Velocity ($v$-prediction) \\
Beta Schedule & Scaled linear ($0.00085 \rightarrow 0.012$) \\
Guidance Scale ($w$) & 7.5 \\
Optimizer & AdamW \\
Learning Rate & $5 \times 10^{-4}$ \\
Batch Size & 64 \\
Training Epochs & 500 \\
\bottomrule
\end{tabular}
}
\end{table}

%----------------------------------------------------------------------------------------
\section{Section 4.B: Quantitative Analysis Details}
\label{supp:quantitative}
%----------------------------------------------------------------------------------------

%----------------------------------------------------------------------------------------
\subsection{Overview}
%----------------------------------------------------------------------------------------
We provide formal definitions of the metrics used to evaluate the physical accuracy and controllability of generated motions.
While standard metrics such as FID and R-Precision assess generation quality and semantic alignment, they do not capture motion intensity or dynamics.
We therefore introduce metrics to quantify:
\begin{itemize}
    \item \textbf{Effort Metric MAE:} deviation from target effort values,
    \item \textbf{Structural Monotonicity:} consistency of intensity scaling across ordered effort levels,
    % \todo{\item \textbf{Laban Effort Dynamics:} alignment with high-level movement qualities.}
    \item \textbf{Laban Monotonicity:} consistency of the higher-order dynamic qualities (Weight, Time, and Flow) intensity scaling across ordered effort levels.

\end{itemize}

%----------------------------------------------------------------------------------------
\subsection{Evaluation Metric Definitions}
\label{supp:metric_defs}
%----------------------------------------------------------------------------------------

%----------------------------------------------------------------------------------------
\subsubsection{Effort Metric MAE}
%----------------------------------------------------------------------------------------
Given joint position $\mathbf{p}_{t,j} \in \mathbb{R}^3$ at frame $t$, we define the instantaneous positional change of skeletal group $G$ with $N_G$ joints as:
\begin{equation}
\delta_t^G = \frac{1}{N_G} \sum_{j \in G} \| \mathbf{p}_{t+1,j} - \mathbf{p}_{t,j} \|_2
\end{equation}
From this signal, we define:
\begin{align}
\text{Peak Change (PC)} &= \max_t \delta_t^G, \\
\text{Collective Change (CC)} &= \sum_t \delta_t^G.
\end{align}
Effort Metric MAE is computed as the absolute difference between generated and ground truth target metric values, averaged over all test samples.

%----------------------------------------------------------------------------------------
\subsubsection{Structural Monotonicity ($\rho$)}
%----------------------------------------------------------------------------------------
We evaluate monotonic intensity scaling across seven ordered effort levels $S=\{0.7,\dots,1.3\}$ using Spearman’s rank correlation.
Correlations are computed separately for Peak Change ($M_{\text{peak}}$) and Collective Change ($M_{\text{coll}}$), for each action and anatomical region:
\begin{equation}
\rho = \frac{\mathrm{cov}(R_S, R_M)}{\sigma_{R_S}\sigma_{R_M}}
\end{equation}
where $R_S$ and $R_M$ denote ranked effort scalars and metric values.
A limb is considered successfully controlled if its motion trend satisfies $p < 0.05$ and $\rho > 0.5$.
We report the percentage of successfully controlled limbs aggregated over all action--body-part pairs, separately for Peak and Collective metrics.

%----------------------------------------------------------------------------------------
\subsubsection{Laban Monotonicity}
%----------------------------------------------------------------------------------------
% \todo{We evaluate dynamic motion qualities}\joshua{We evaluate monotonic
% intensity scaling of dynamic quality} using computational approximations of Laban Effort factors based on higher-order joint derivatives:
%\todo{We evaluate dynamic motion qualities}
We evaluate monotonic
intensity scaling of dynamic quality using computational approximations of Laban Effort factors adapted from Samadani et al.~\cite{samadani2020affectivemovementgenerationusing}, standardized here to capture peak intensity magnitudes:
\begin{itemize}
    \item \textbf{Weight:} peak kinetic energy,
    \begin{equation}
    E_{\text{weight}} = \max_t \sum_j \|\mathbf{v}_{t,j}\|_2^2
    \end{equation}
    \item \textbf{Time:} peak net acceleration,
    \begin{equation}
    E_{\text{time}} = \max_t \sum_j \|\mathbf{a}_{t,j}\|_2
    \end{equation}
    \item \textbf{Flow:} peak jerk magnitude,
    \begin{equation}
    E_{\text{flow}} = \max_t \sum_j \|\mathbf{j}_{t,j}\|_2
    \end{equation}
\end{itemize}

For each action category, we evaluate the monotonic relationship between the input effort scalars and the extracted LMA metrics using Spearman’s rank correlation.
An action is considered to exhibit consistent Laban dynamics if the trend satisfies $p < 0.05$ and $\rho > 0.5$.
We report the percentage of actions meeting these criteria for \textit{Weight}, \textit{Time}, and \textit{Flow}.

%----------------------------------------------------------------------------------------
\subsection{Ablation Studies and Component Analysis}
\label{supp:ablation_studies}
%----------------------------------------------------------------------------------------

To isolate the contributions of our architectural choices, we evaluated the impact of the Anatomical Region embedding (ARE) and the Data Augmentation strategy on effort consistency and physical fidelity.
The corresponding quantitative results are detailed in Tables \ref{tab:merged_evaluation}, \ref{tab:dist_metrics}, and \ref{tab:mobert_metrics}.

\begin{table*}[t]
\caption{Unified Quantitative Evaluation Table. We report Mean Absolute Error (MAE) of effort metrics, Monotonicity Rates (\%) for structural metrics, and Monotonicity Accuracy (\%) for Laban kinetic metrics (Weight, Flow, Time). Arrows indicate whether lower (↓) or higher (↑) values are better.} 
\centering
\setlength{\tabcolsep}{4pt} % Adjust column spacing to fit
\begin{tabular}{l c c c c c c c}
\toprule
\multirow{2}{*}{\textbf{Model}} & \multicolumn{2}{c}{\textbf{Effort Metric MAE} $\downarrow$} & \multicolumn{2}{c}{\textbf{Structural Mono.} $\uparrow$} & \multicolumn{3}{c}{\textbf{Laban Mono. (\%)} $\uparrow$} \\
\cmidrule(lr){2-3} \cmidrule(lr){4-5} \cmidrule(lr){6-8}
 & \textbf{Peak} & \textbf{Coll.} & \textbf{Peak} & \textbf{Coll.} & \textbf{Weight} & \textbf{Flow} & \textbf{Time} \\
\midrule
Global Embeddings & 0.0626 & 1.614 & 78.6 & 64.3 & 71.4 & 85.7 & 85.7 \\
ARE (Ours) & 0.0597 & 1.574 & 74.5 & 60.2 & 78.6 & 85.7 & 92.9 \\
ARE  (Orig$\rightarrow$Orig) & 0.0570 & 2.033 & 63.3 & 55.1 & 71.4 & 71.4 & 92.9 \\
ARE  (Orig$\rightarrow$Aug) & 0.0684 & 1.892 & 63.3 & 55.1 & 71.4 & 71.4 & 92.9 \\
SALAD & - & - & 22.4 & 17.3 & 21.4 & 14.3 & 28.6 \\
\bottomrule
\end{tabular}
\label{tab:merged_evaluation}
\end{table*}

\begin{table*}[t]
\caption{Standard Evaluation Metrics Table. We report the Evaluation of distribution quality (FID, Diversity) and retrieval performance (Top-$k$ R-precision and Multi-Modality). Arrows indicate whether lower ($\downarrow$) or higher ($\uparrow$) values are better.}
\centering
\begin{tabular}{l c c c c c c }
\hline
\textbf{Model} 
& \textbf{FID} $\downarrow$
& \textbf{Diversity} $\uparrow$
& \textbf{Top-1} $\uparrow$
& \textbf{Top-2} $\uparrow$
& \textbf{Top-3} $\uparrow$
& \textbf{Multi-Modality} $\uparrow$\\
\hline

Global Embeddings & $0.143 \pm 0.003$ & $9.392 \pm 0.071$ & $0.521 \pm 0.001$ & $0.715 \pm 0.001$ & $0.812 \pm 0.001$ & $1.605 \pm 0.040$ \\
ARE (Ours) & $0.056 \pm 0.002$ & $9.410 \pm 0.086$ & $0.517 \pm 0.002$ & $0.711 \pm 0.002$ & $0.808 \pm 0.001$ & $1.579 \pm 0.057$ \\% PBP Token Pk & -- & -- & -- & -- & -- & --  & -- & -- & --\\
ARE [Orig→Orig] & $0.055 \pm 0.003$ & $9.462 \pm 0.075$ & $0.558 \pm 0.004$ & $0.749 \pm 0.003$ & $0.837 \pm 0.002$ & $1.506 \pm 0.073$ \\

ARE [Orig→Aug] & $0.231 \pm 0.005$ & $9.392 \pm 0.065$ & $0.483 \pm 0.001$ & $0.674 \pm 0.001$ & $0.775 \pm 0.001$ & $1.941 \pm 0.057$ \\
SALAD & $0.115 \pm 0.003$ & $9.586 \pm 0.070$ & $0.553 \pm 0.001$ & $0.749 \pm 0.001$ & $0.842 \pm 0.001$ & $1.869 \pm 0.069$ \\

\hline
\end{tabular}
\label{tab:dist_metrics}
\end{table*}

\begin{table*}[bt]
\caption{MoBERT-based Semantic Motion Evaluation. We report the evaluation of semantic motion quality (Alignment, Faithfulness, and Naturalness). Arrows indicate that higher ($\uparrow$) values are better.}
\centering
\begin{tabular}{l c c c}
\hline
\textbf{Model}
& \textbf{Alignment} $\uparrow$
& \textbf{Faithfulness} $\uparrow$
& \textbf{Naturalness} $\uparrow$ \\
\hline
Global Embeddings & $0.107 \pm 0.001$ & $0.535 \pm 0.000$ & $0.565 \pm 0.000$ \\
ARE (Ours) & $0.100 \pm 0.001$ & $0.529 \pm 0.000$ & $0.611 \pm 0.000$ \\
ARE[Orig→Orig] & $0.138 \pm 0.001$ & $0.552 \pm 0.001$ & $0.573 \pm 0.000$ \\
ARE[Orig→Aug] & $0.111 \pm 0.001$ & $0.535 \pm 0.000$ & $0.560 \pm 0.000$ \\
SALAD & $0.120 \pm 0.001$ & $0.537 \pm 0.000$ & $0.564 \pm 0.000$ \\
\hline
\end{tabular}
\label{tab:mobert_metrics}
\end{table*}

%----------------------------------------------------------------------------------------
\subsubsection{Impact of Conditioning Strategy: ARE vs. Global Embedding}
\begin{figure*}[tb]
\centering
\includegraphics[width=0.9\linewidth]{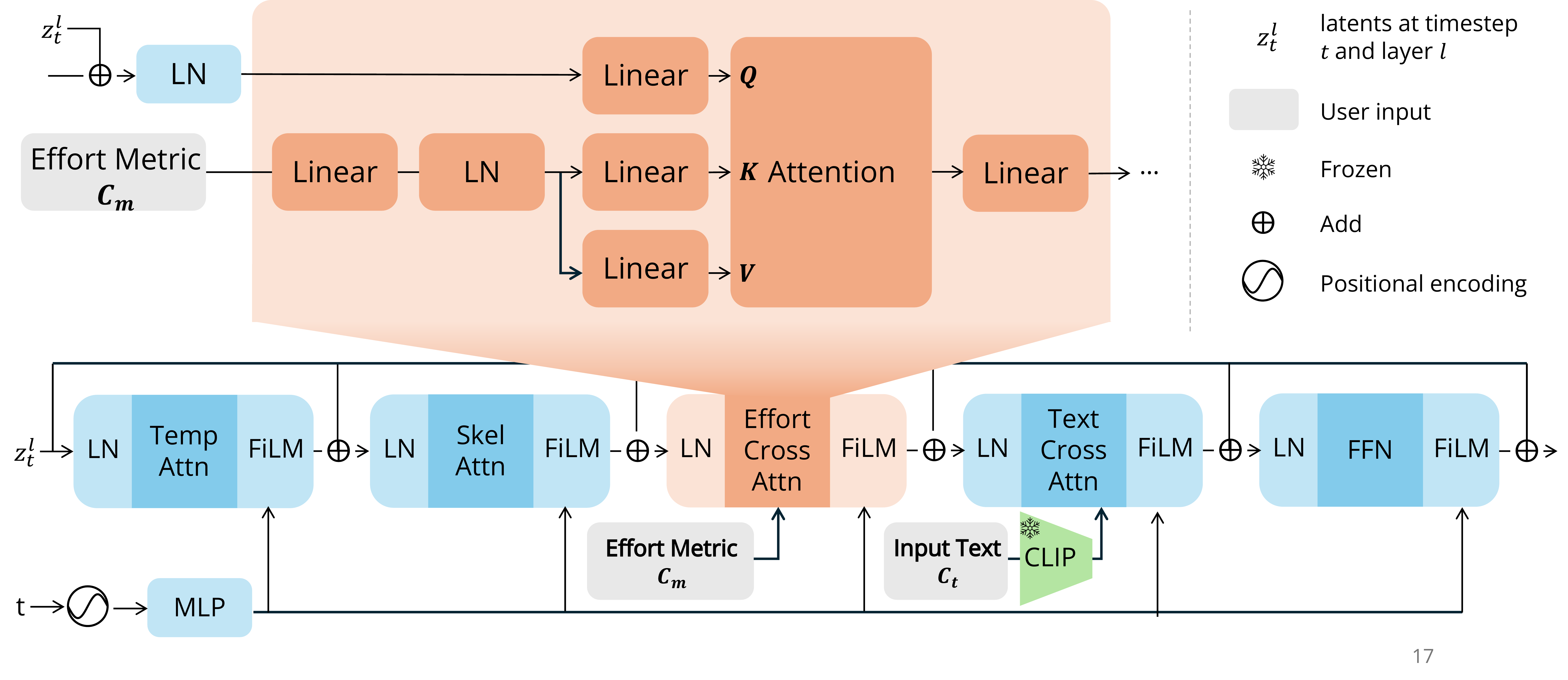}
\caption{Global embedding setup }
\label{fig:global_architecture}
\end{figure*}

%----------------------------------------------------------------------------------------
Table~\ref{tab:merged_evaluation} presents a comparative analysis between the Global Embedding strategy (Fig.~\ref{fig:global_architecture}) and our Anatomical Region Embedding (ARE). 
The main difference of the Global Embedding strategy is that it flattens the input metrics into $\mathbf{c}_m \in \mathbb{R}^{B \times (N_g \cdot N_p)}$ and lacks the region-identity embedding ($\mathbf{P}_{\mathrm{id}}$).

The results reveal a clear trade-off between crude responsiveness and structural precision. 
%While Global conditioning achieves marginally higher raw structural monotonicity, its \joshuav{higher Collective MAE ($1.614$ vs. $1.574$)} indicates that a global signal creates a uniform bias that amplifies motion across the entire skeleton indiscriminately.
% \joshuav{While Global conditioning achieves marginally higher monotonicity, it suffers from a higher Collective MAE (1.614 versus 1.574) and a worse FID (0.143 versus 0.056). 
% This indicates that global signals cause uniform changes across the entire skeleton regardless of the target region, which distorts the motion distribution.
% In contrast, our ARE setup achieves superior distributional fidelity and structural precision.
% This proves that localized conditioning is essential to maintain natural motion while following effort constraints.}
\joshuav{Although the numerical improvement in Collective MAE (1.574 vs. 1.614) appears incremental, it is coupled with a reduction in FID (0.056 vs. 0.143). This indicates that while Global conditioning can approximate the target effort, it does so at the cost of significant structural distortion. ARE, however, achieves higher precision without compromising the underlying motion manifold.}

Furthermore, the Local setup better captures high-level Laban descriptors, particularly in Weight ($78.6\%$ vs. $71.4\%$) and Time ($92.9\%$ vs. $85.7\%$), confirming that spatially distributed conditioning enables precise modulation without degrading the underlying motion structure.

%----------------------------------------------------------------------------------------
\subsubsection{Necessity of Data Augmentation}
%----------------------------------------------------------------------------------------
Standard datasets such as HumanML3D are inherently biased toward motions of moderate intensity, which limits a model’s ability to generalize to extreme-effort regimes.
We evaluate a baseline model trained solely on the original dataset (Orig$\to$Aug) under out-of-distribution effort conditions.
Compared to our full model, ARE(Ours), this baseline exhibits a substantial degradation in performance: \joshuav{the Collective MAE is significantly higher ($1.892$ vs.\ $1.574$)}, and Peak Monotonicity drops from $74.5\%$ to $63.3\%$, indicating reduced ability to appropriately scale effort.

Notably, while the baseline can capture simple speed variations—achieving comparable performance on Laban Time ($92.9\%$)—it lacks the diversity required to learn more complex effort dynamics.
As a result, it significantly underperforms on Flow ($71.4\%$ vs.\ $85.7\%$) and Weight ($71.4\%$ vs.\ $78.6\%$), which depend on subtler physical properties beyond raw speed.

These results demonstrate that data augmentation plays a critical role not only in extending the effective effort range but also in enabling the disentanglement of nuanced physical attributes such as fluidity and heaviness that are underrepresented in standard motion capture datasets.

%----------------------------------------------------------------------------------------
\section{Section 4.C: Qualitative Results}
%----------------------------------------------------------------------------------------
%----------------------------------------------------------------------------------------
\subsection{ Extending Body-Part Effort Modulation}
%----------------------------------------------------------------------------------------

\begin{figure*}[t]
\centering
  \includegraphics[width=\textwidth]{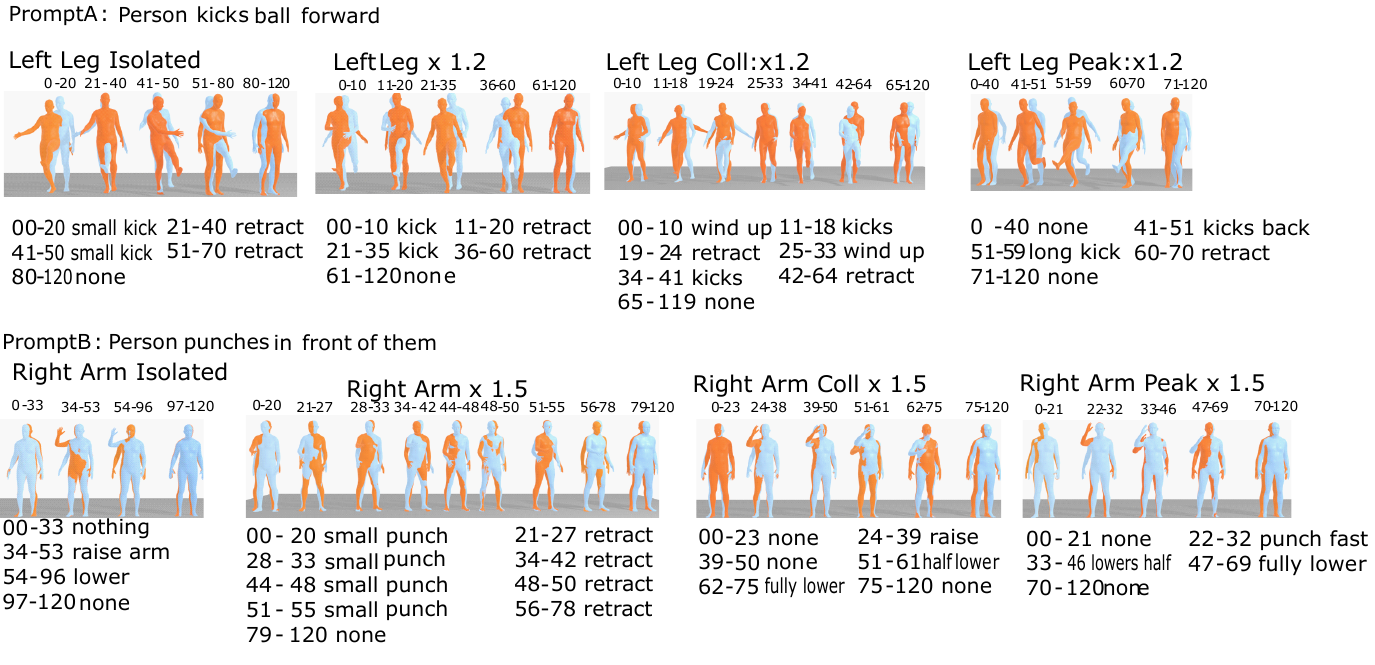}
\caption{Temporal comparison of ``Person kicks ball forward'' (Left Lower, x1.2) and ``Person punches in front of them'' (Right Upper, x1.5) under Baseline, Scaled, Collective-only, and Peak-only conditions.
Superimposed blue (onset) and orange (offset) meshes indicate key motion transitions, annotated with frame ranges and phase descriptions.
Peak scaling primarily affects motion velocity and sharpness, whereas Collective scaling increases spatial extent and overall range of motion.}
\label{fig:pnc2}
\end{figure*}

We analyze the effects of the Peak and Collective effort metrics on motion generation using two representative prompts: ``Person kicks ball forward'' and ``Person punches in front of them.'' 
As shown in Figure~\ref{fig:pnc2}, we compare isolated baseline generations with edited variants in which Peak and Collective parameters are selectively scaled.
This comparison enables a direct examination of how each component influences motion kinematics.

%----------------------------------------------------------------------------------------
\subsection{Component Analysis}
%----------------------------------------------------------------------------------------
\textbf{The Isolated Right Arm Punch.}
The baseline motion consists of a standard arm raise executed over approximately 20 frames (34--53) with a bent elbow configuration.
Peak-dominant editing (Peak $\times$1.5) concentrates the exertion into a short 11-frame interval (22--32), compressing the temporal span to roughly half of the baseline.
This temporal compression produces a sharp, uppercut-like motion.
In contrast, Collective-dominant editing (Coll $\times$1.5) emphasizes spatial reach rather than speed.
The arm extends further forward, forming a slower 15-frame raise (24--38) followed by a prolonged, segmented retraction phase (51--75) that extends well beyond the baseline termination.

\textbf{The Isolated Left Leg Kick.}
The baseline motion comprises two kicks (0--20 and 41--50) separated by retraction phases.
Peak-dominant editing (Left Lower Peak $\times$1.2) tightens this structure by delaying the onset until frame 40 and executing a single compact kick within a brief 19-frame window.
This edit collapses the multi-kick sequence into a focused, high-intensity action.
By contrast, Collective-only editing preserves the original multi-kick structure while introducing preparatory dynamics.
Specifically, it inserts an explicit wind-up phase (0--10) before the first kick and a combined retraction and wind-up phase (21--33) before the second.
This behavior indicates that the Collective parameter promotes fuller, more anticipatory motions rather than merely accelerating existing trajectories.

%----------------------------------------------------------------------------------------
\subsection{Complementary Relationship of Peak and Collective}
%----------------------------------------------------------------------------------------
Our observations indicate that realistic high-intensity motion editing benefits from the simultaneous scaling of both Peak and Collective parameters.
Peak supplies the acceleration necessary to convey impact, while Collective contributes the total action volume required to convey magnitude.

When both parameters are scaled concurrently, the model exhibits distinct behavioral adaptations:

\begin{itemize}
    \item \textbf{Right Upper (x1.5):}
    Joint scaling produces a sequence of five compact punches within the first 55 frames.
    This outcome differs from the isolated cases, avoiding both the single slow extension observed under Collective-only scaling and the solitary compact strike produced by Peak-only scaling.
    The increased action frequency reflects an attempt to satisfy the dual constraints of high velocity and large cumulative displacement.
    
    \item \textbf{Left Lower (x1.2):}
    Combined scaling substantially alters the initial posture and temporal structure.
    Whereas the baseline begins from a neutral stance and distributes two small kicks across the sequence, the x1.2 condition initializes in a pre-wound posture.
    The model immediately executes a long kick (0--10), retracts rapidly (11--20), and launches a second fast kick (21--35).
    This pre-wound initialization maximizes early displacement and compresses the timeline to accommodate multiple high-velocity actions.
\end{itemize}

Overall, scaling both parameters compels the model to resolve the competing demands of speed and displacement through increased action density rather than through speed or reach alone.

%----------------------------------------------------------------------------------------
\section{Section 4.E: User Study Set-up}
\label{sec:supp_user_study_setup}
%----------------------------------------------------------------------------------------
%----------------------------------------------------------------------------------------
\subsection{Experimental Design}
%----------------------------------------------------------------------------------------

\begin{figure}[h]
\centering
\includegraphics[width=0.9\linewidth]{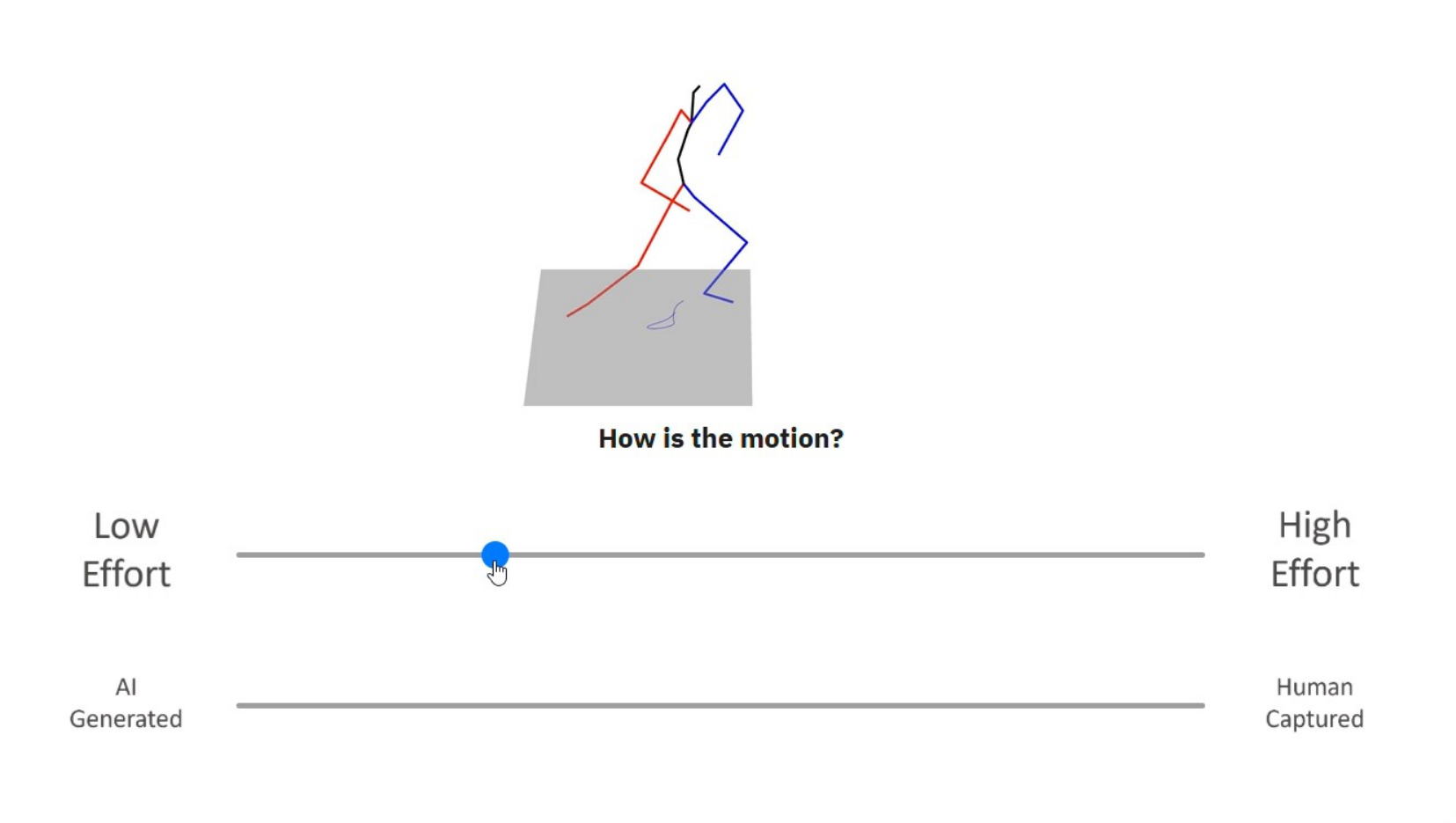}
\caption{Example of a survey question presented to participants.}
\label{fig:user_study_setup}
\end{figure}

The study followed a within-subjects design with a single-blind protocol.
Participants were recruited via Prolific, and the experiment was administered using LimeSurvey.
Under this setup, researchers remained blind to participant identities, and participants were naive to the generation method, effort metrics, and condition labels.
All stimuli were presented without textual identifiers or effort indicators.
An example of the survey interface is shown in Fig.~\ref{fig:user_study_setup}.

%----------------------------------------------------------------------------------------
\subsection{Dependent Variables}
%----------------------------------------------------------------------------------------

Participants were provided with detailed definitions of the evaluation criteria during the consent and instruction phase.
During each trial, a single video stimulus was presented together with the prompt: ``How is the motion?''
Participants evaluated each stimulus using two continuous Visual Analog Scales (VAS), as illustrated in Fig.~\ref{fig:user_study_setup}.

\begin{itemize}
    \item \textbf{Perceived Effort:} A scale ranging from 0 (Low Effort) to 100 (High Effort).
    \item \textbf{Perceived Humanness:} A scale ranging from 0 (Artificially Generated) to 100 (Completely Human-like).
\end{itemize}

%----------------------------------------------------------------------------------------
\subsection{Procedure and Filtering}
%----------------------------------------------------------------------------------------

To mitigate initialization noise, the session was divided into three blocks, each corresponding to a unique random seed.
Within each block, stimuli were presented in a randomized order.

To ensure data quality, attention checks were embedded throughout the experiment.
These checks consisted of videos with unambiguous speed characteristics, accompanied by an explicit instruction prompting participants to adjust the slider to either 0 or 100.
Participants who failed more than one attention check were excluded from the analysis.

Out of 29 recruited participants, 8 were excluded based on this criterion, resulting in a final sample size of $N=21$.

%----------------------------------------------------------------------------------------
\section{Section 4.D: Further Trend Analysis}
%----------------------------------------------------------------------------------------

\begin{figure*}[t]
\centering
\trendrow{Full body (Kicks)}{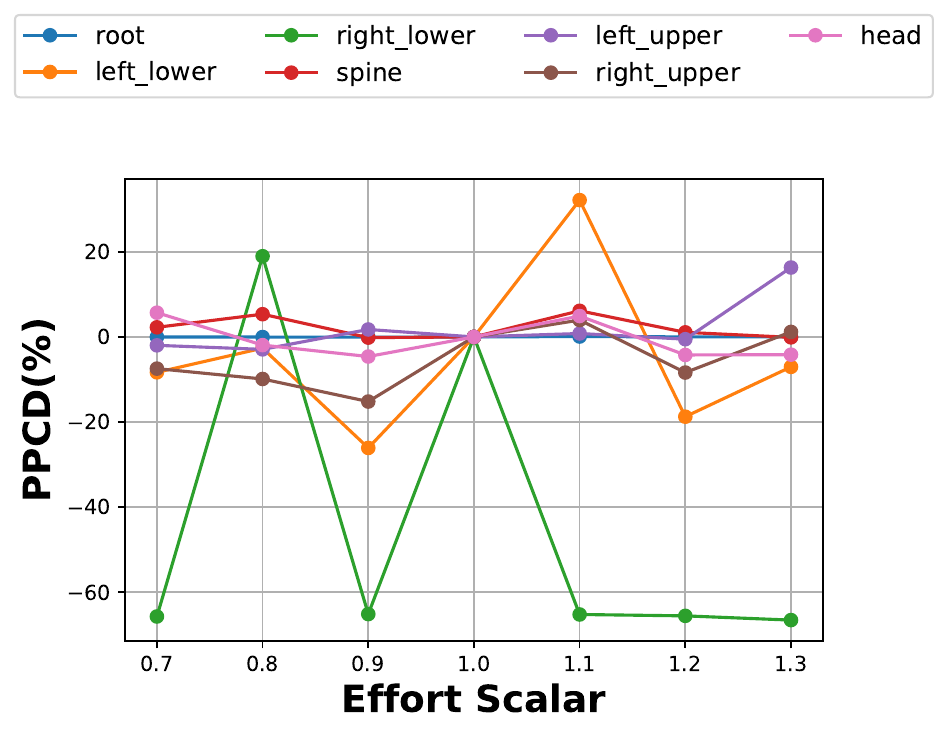}{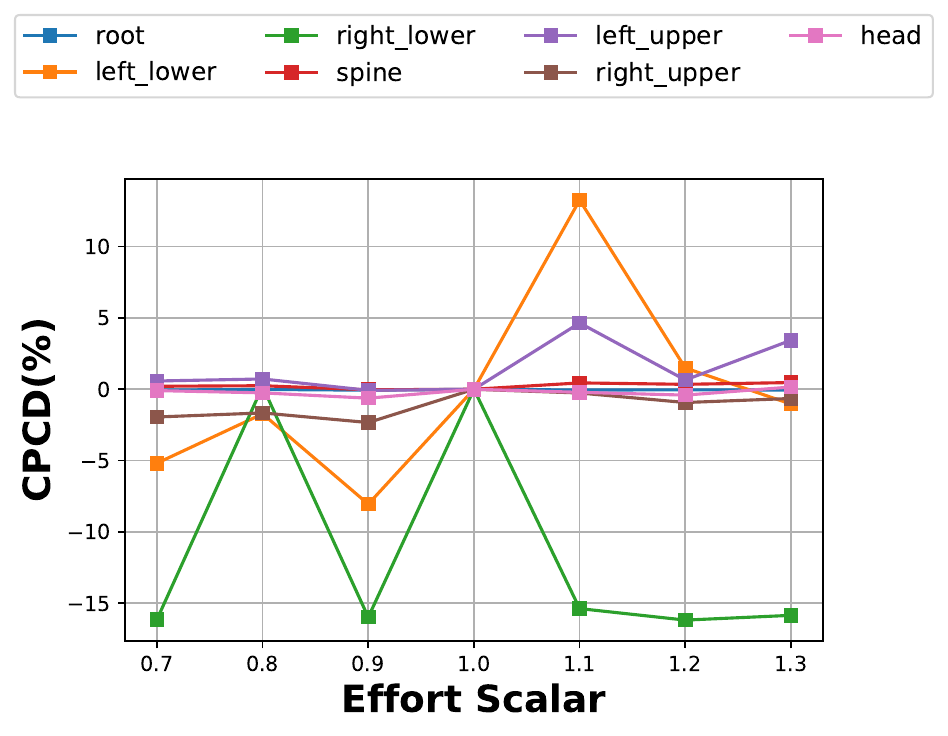}{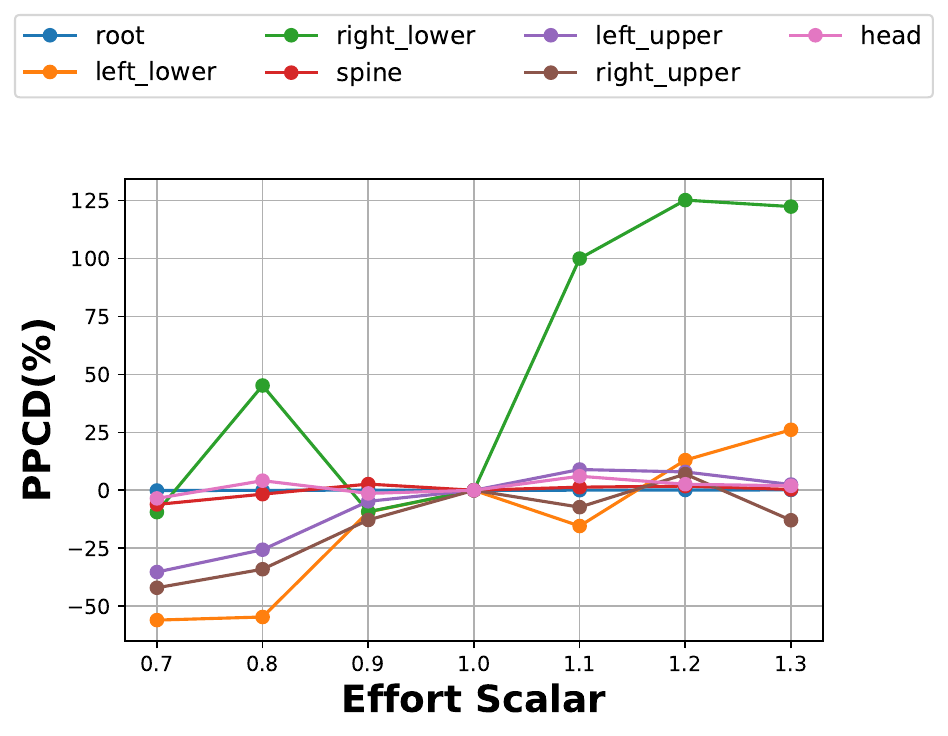}{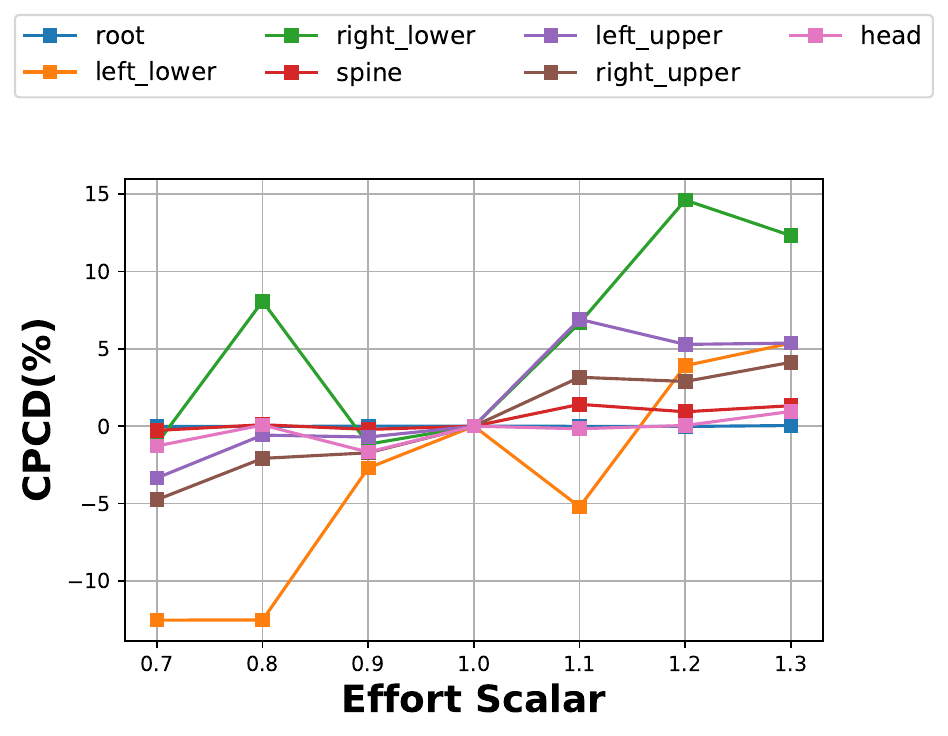}
\trendrow{Lower body (Lunges Forward)}{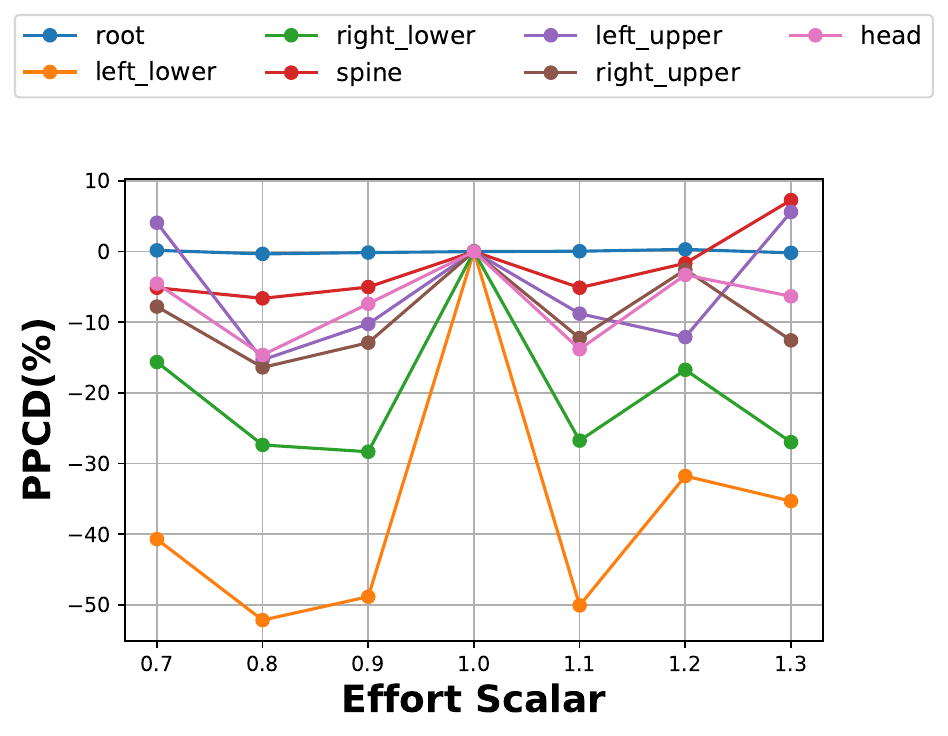}{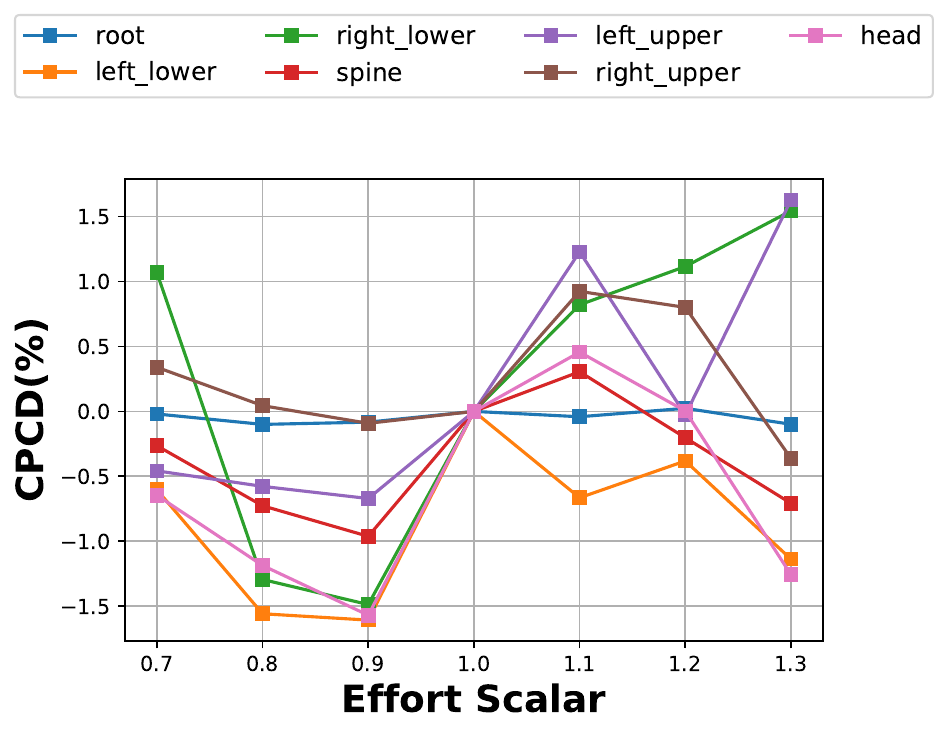}{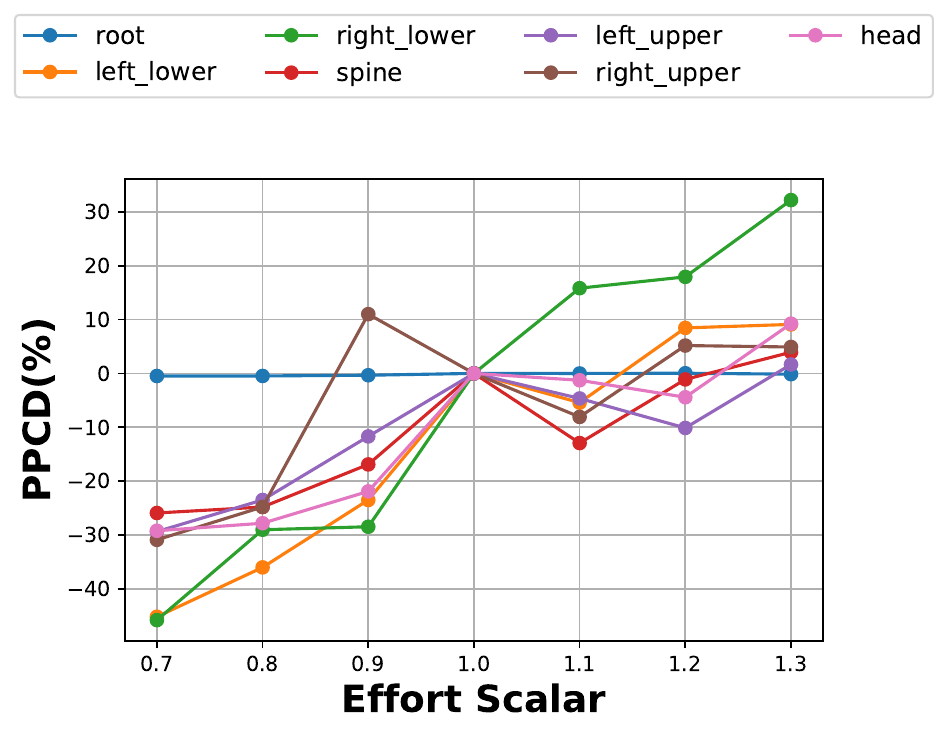}{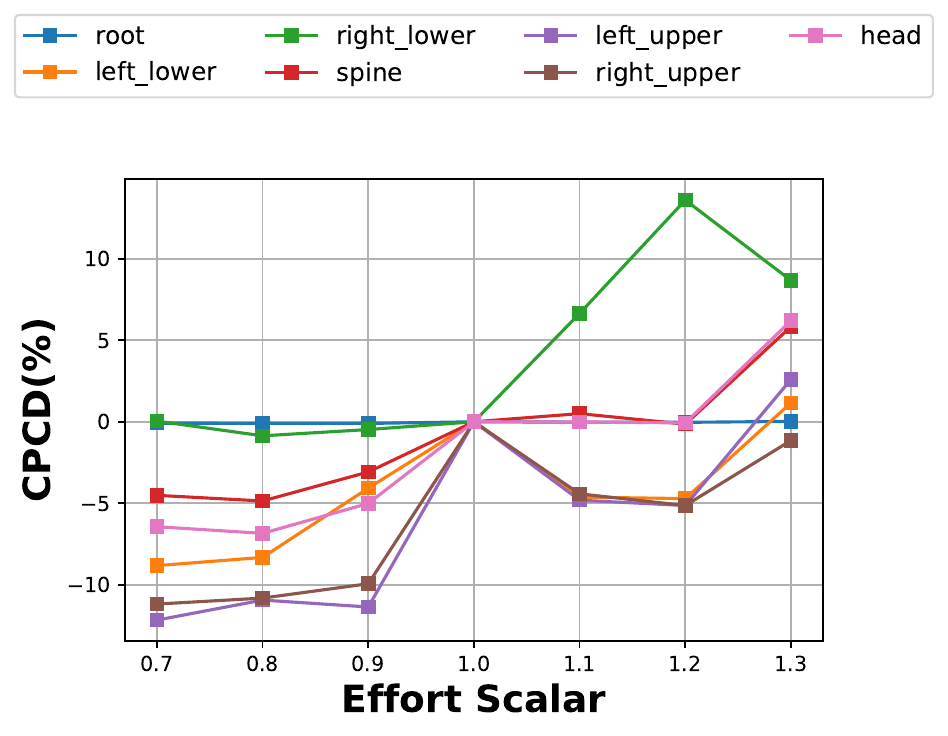}
\trendrow{Upper body (Punches)}{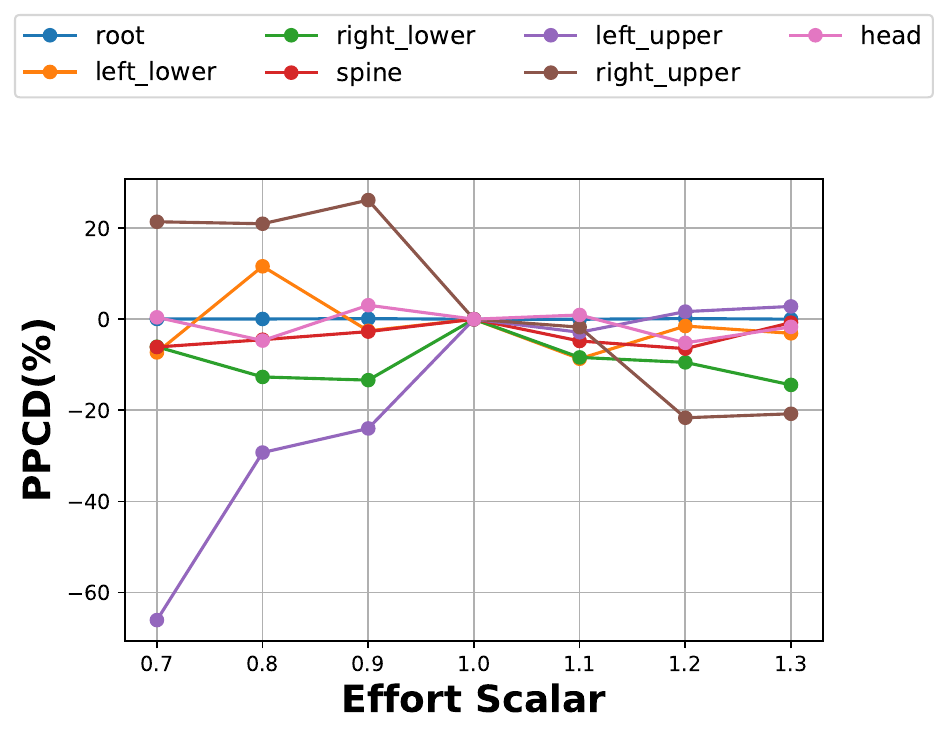}{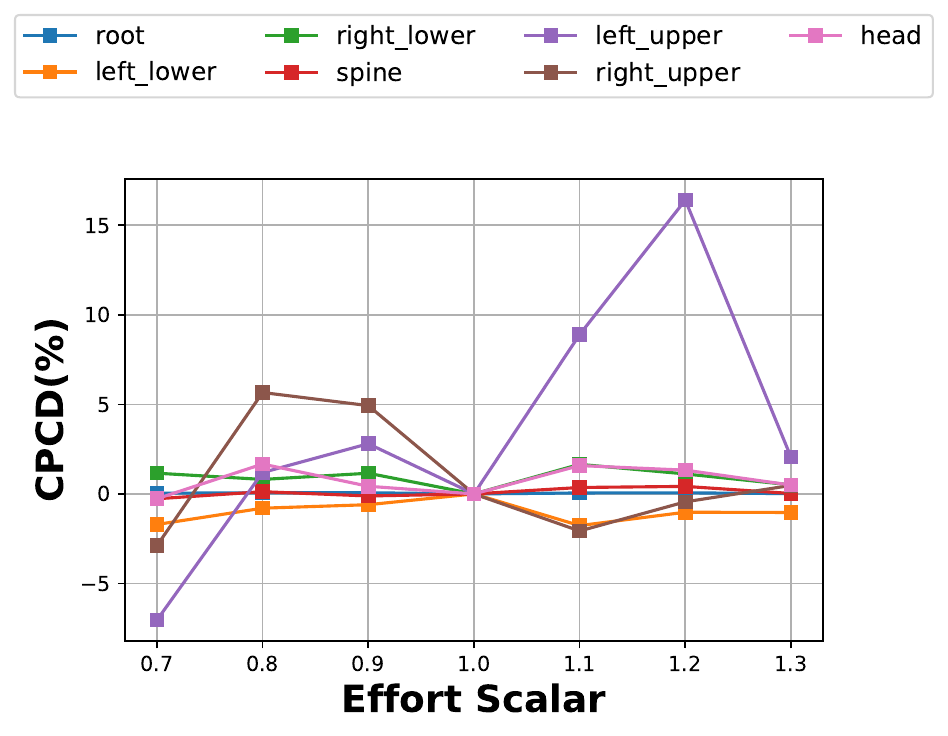}{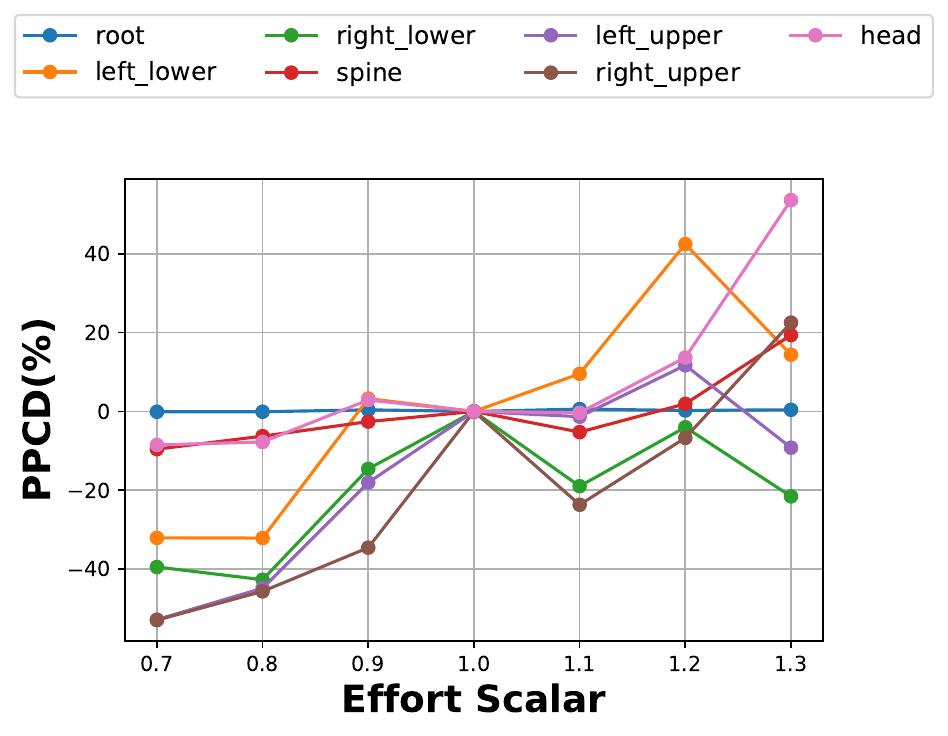}{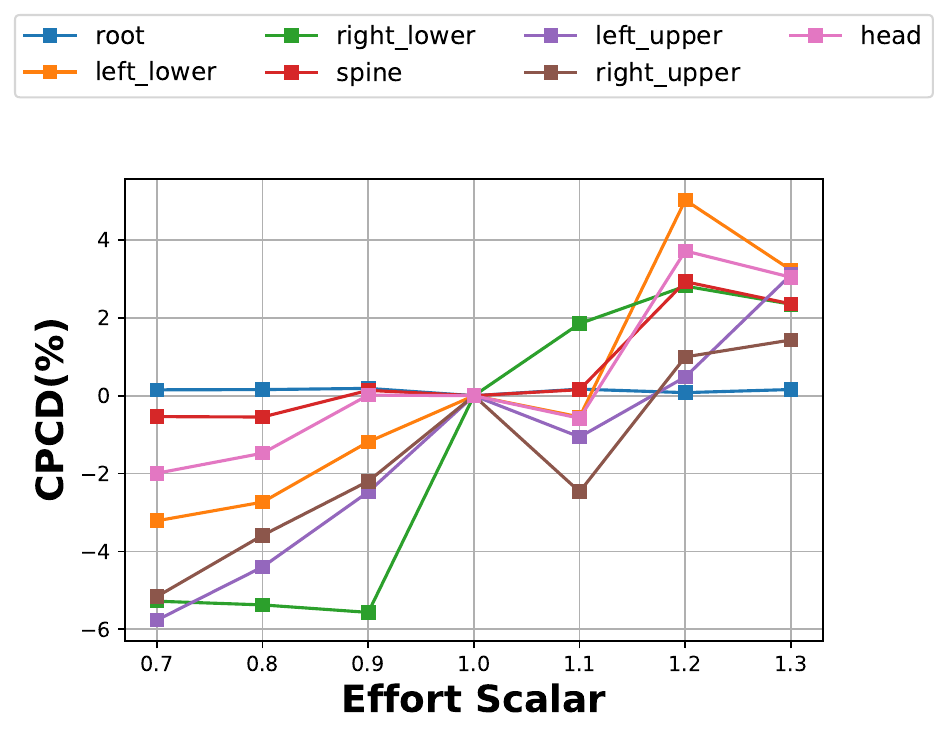}
\caption{Trend comparison across representative actions for SALAD and EMA (Part 2 of 5).}
\label{fig:trend_compact_1}

\end{figure*}

\begin{figure*}[t]
\centering
\trendrow{Upper body (Throws a Ball)}{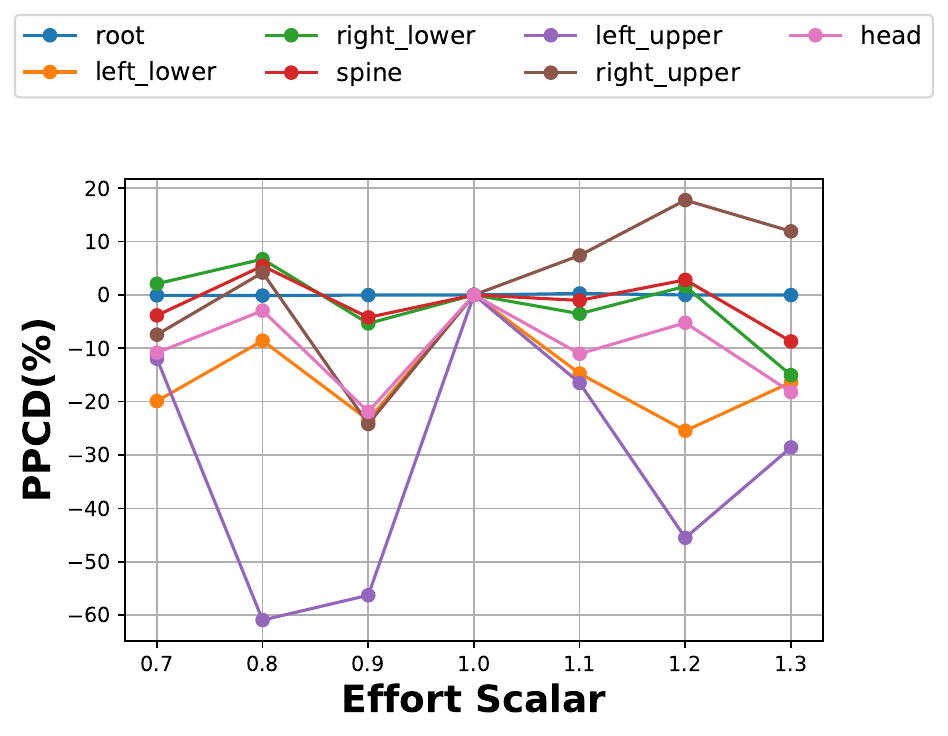}{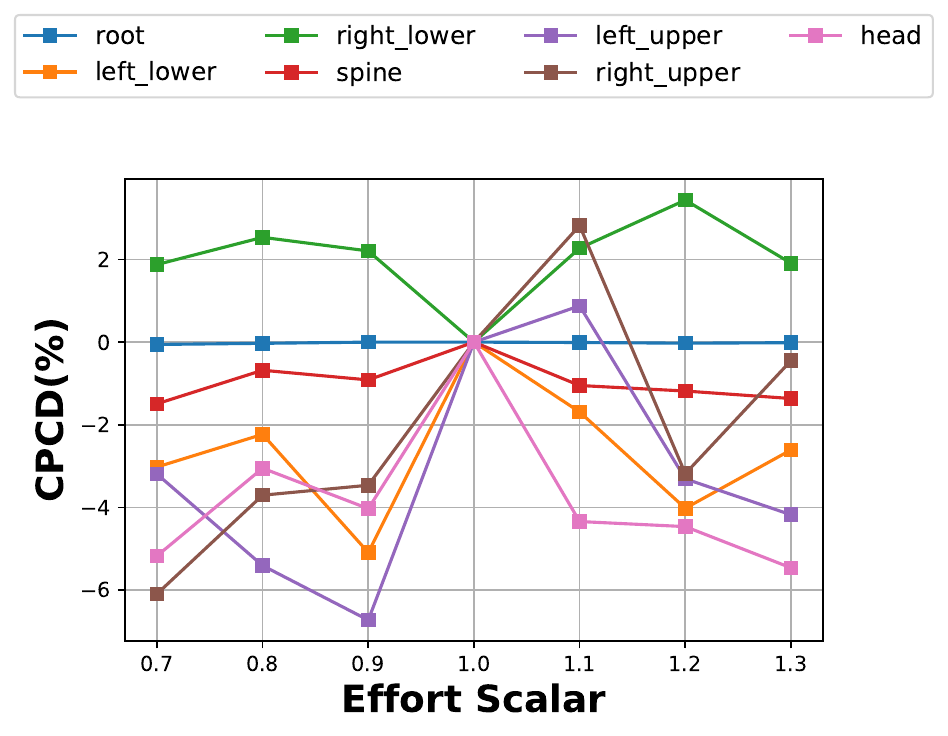}{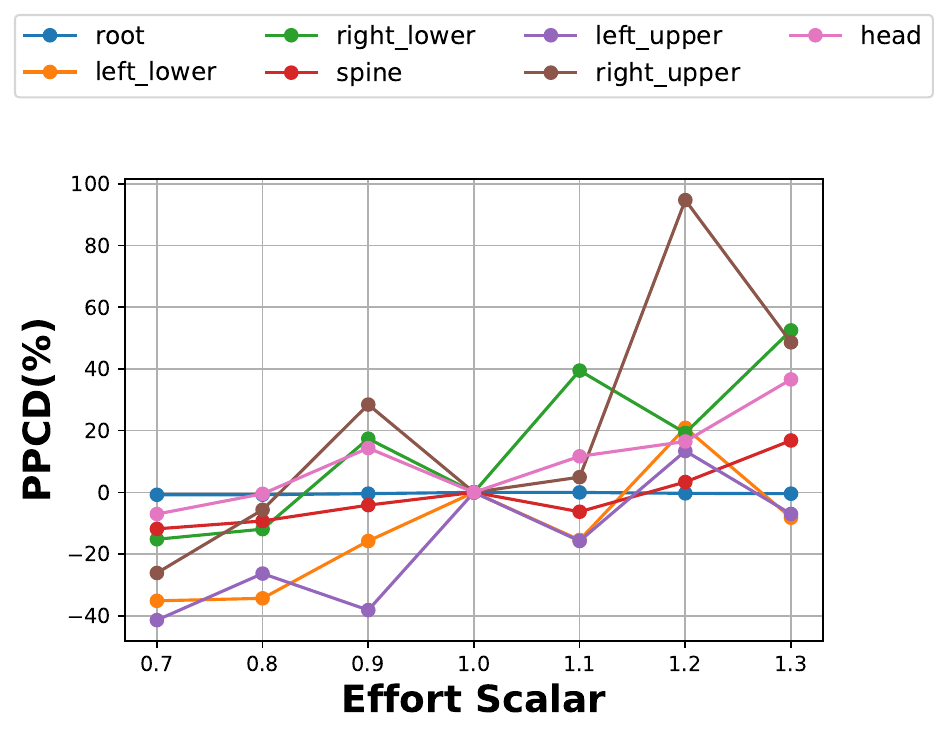}{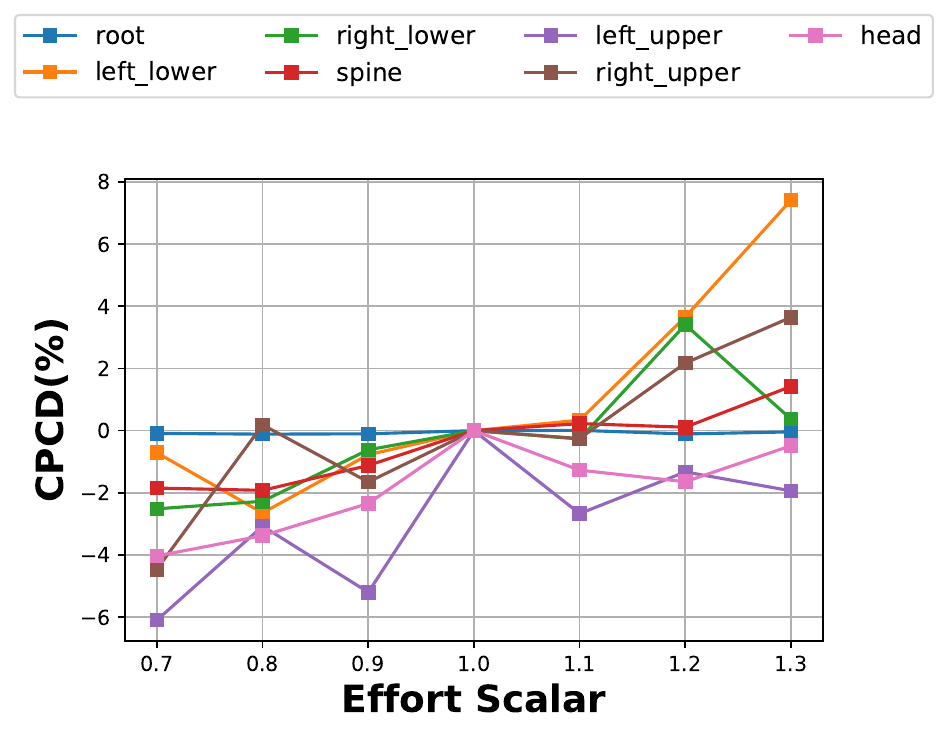}
\trendrow{Lower body (Walks)}{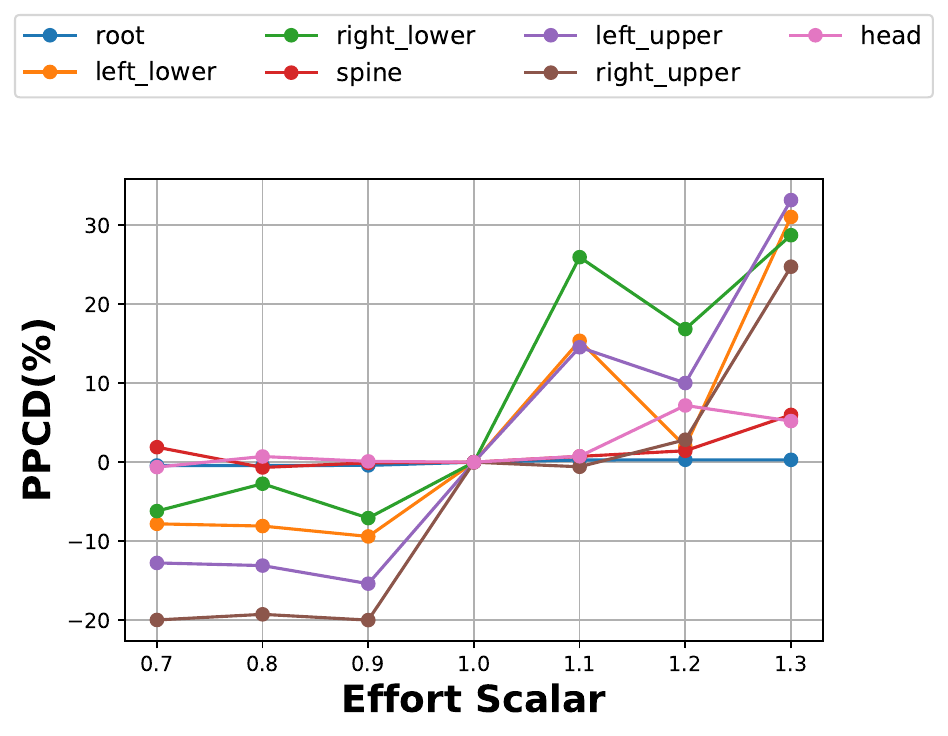}{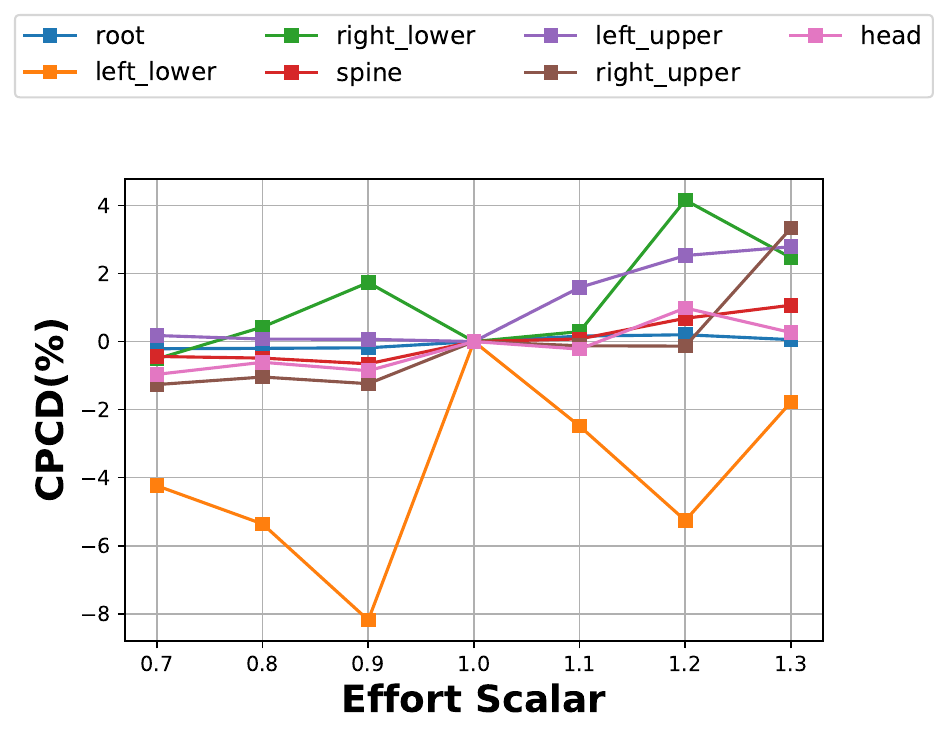}{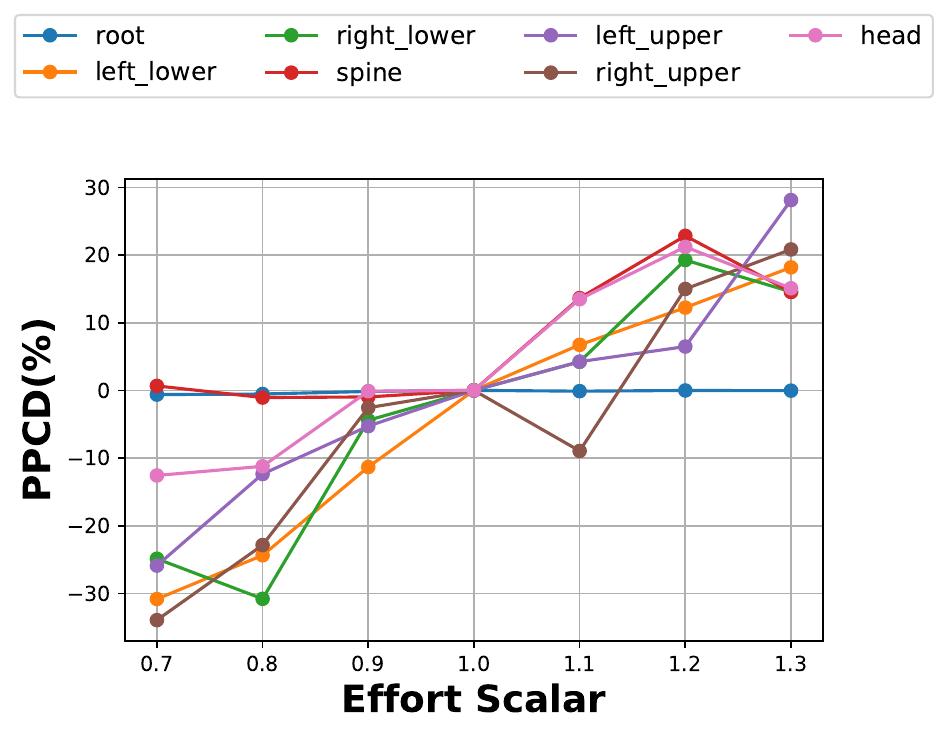}{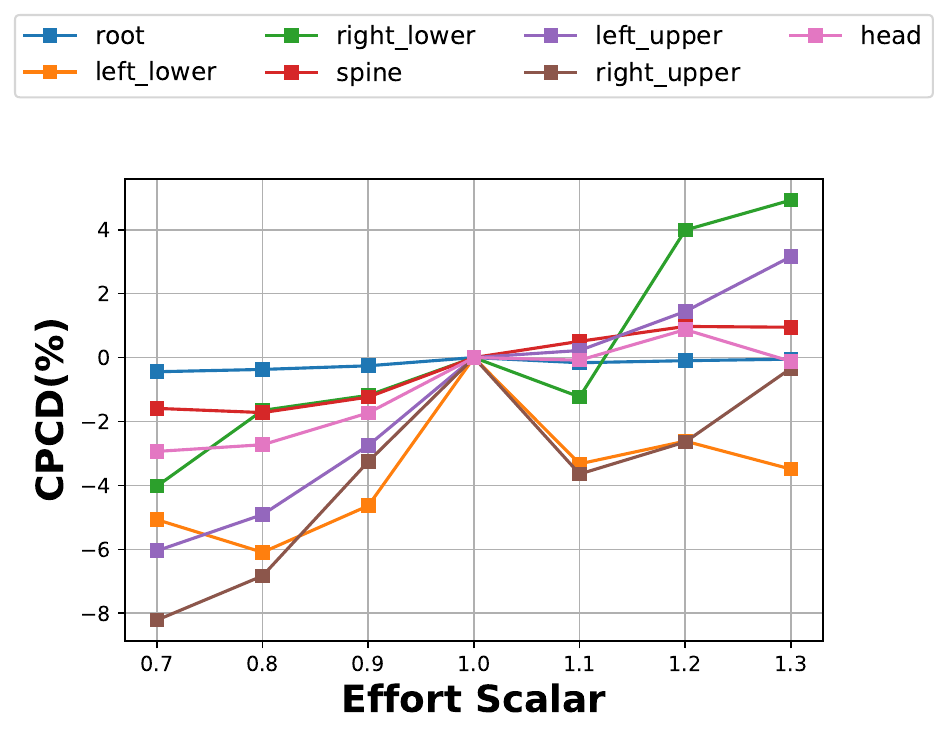}
\trendrow{Upper body (Waves Arms)}{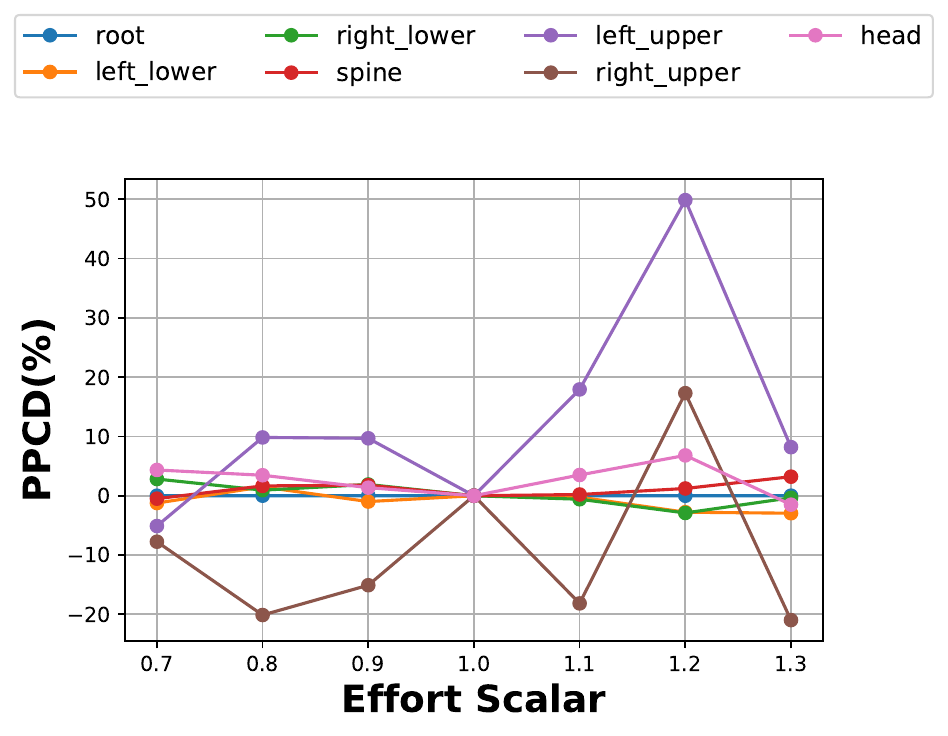}{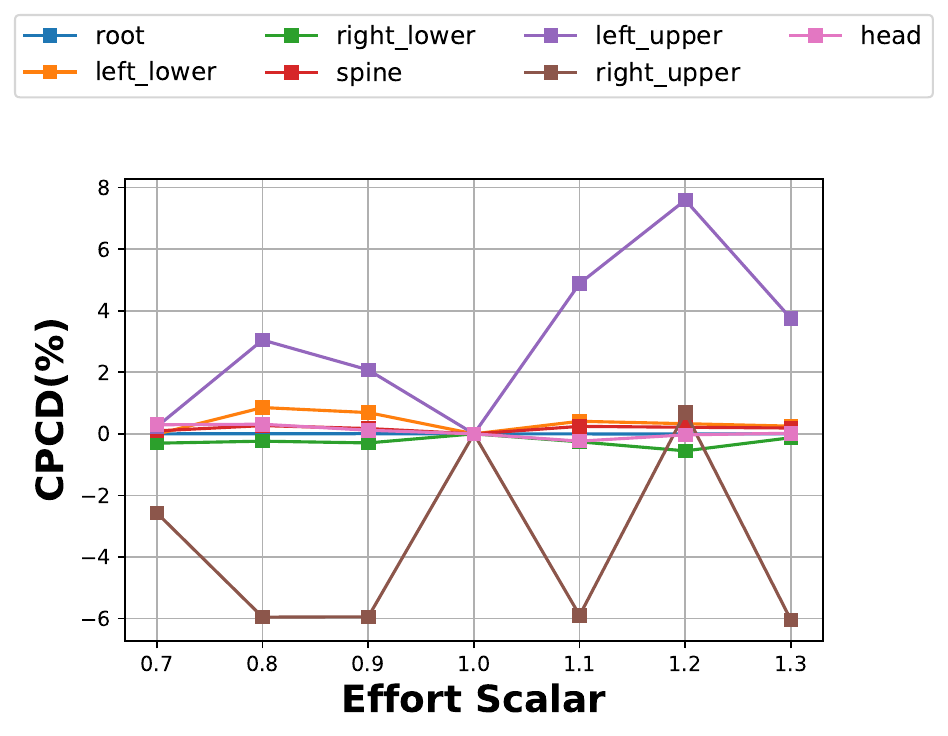}{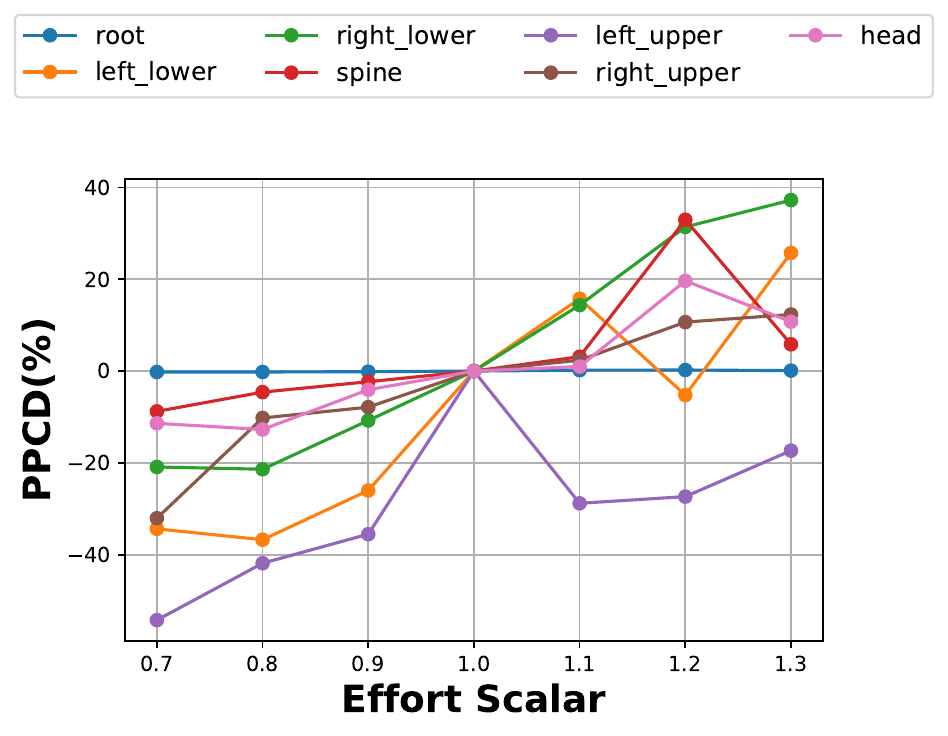}{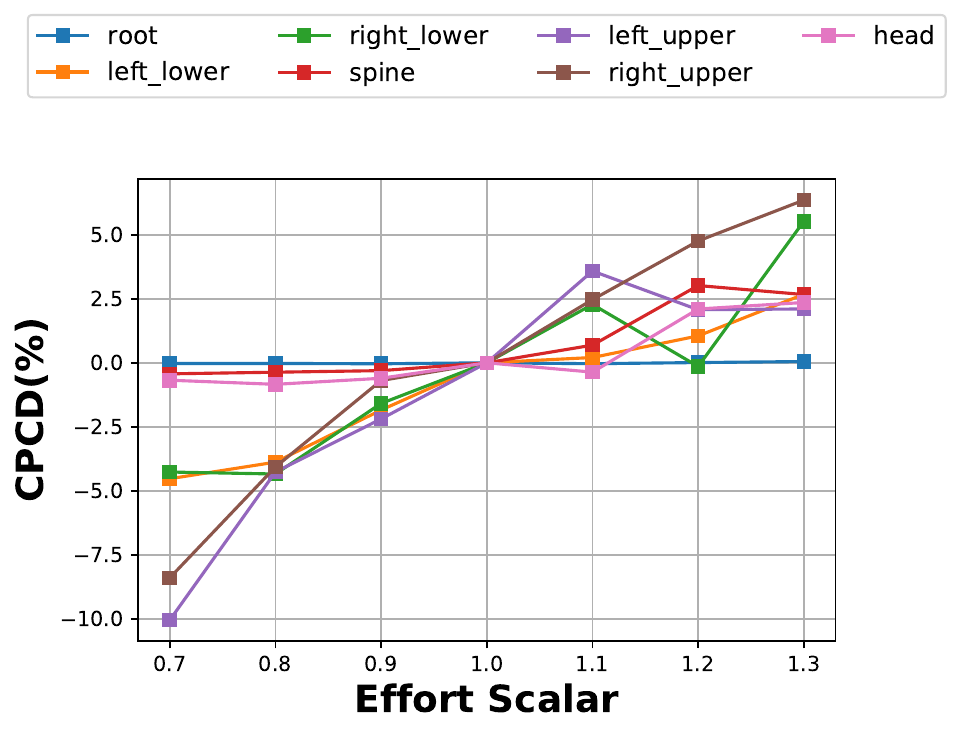}
\caption{Trend comparison across representative actions for SALAD and EMA (Part 3 of 5).}
\label{fig:trend_compact_2}

\end{figure*}

\begin{figure*}[t]
\centering
\trendrow{Full body (Dances)}{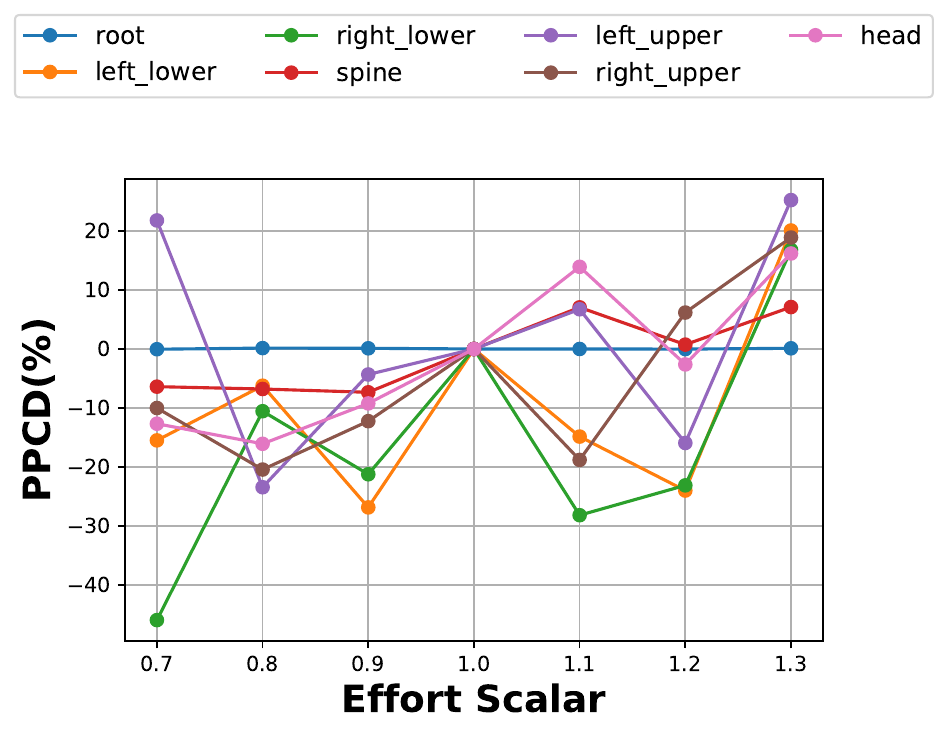}{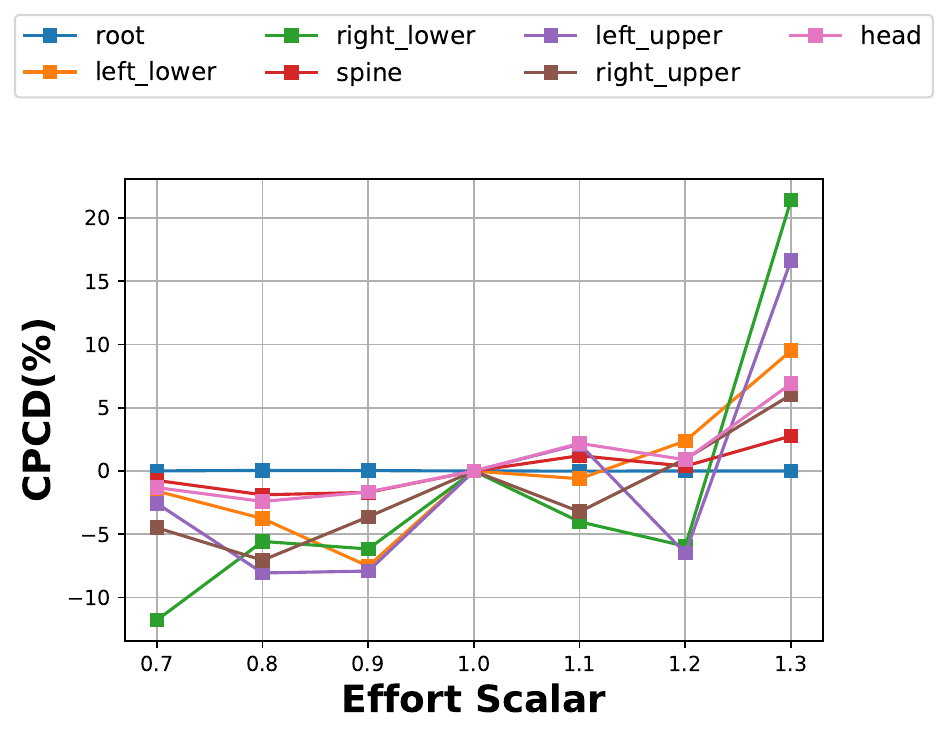}{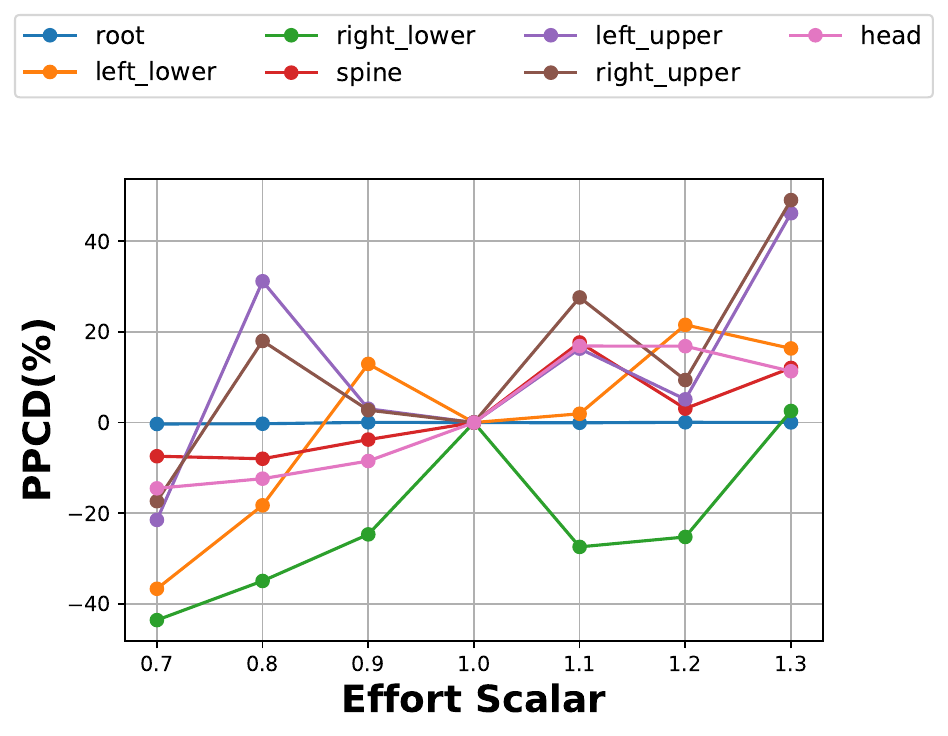}{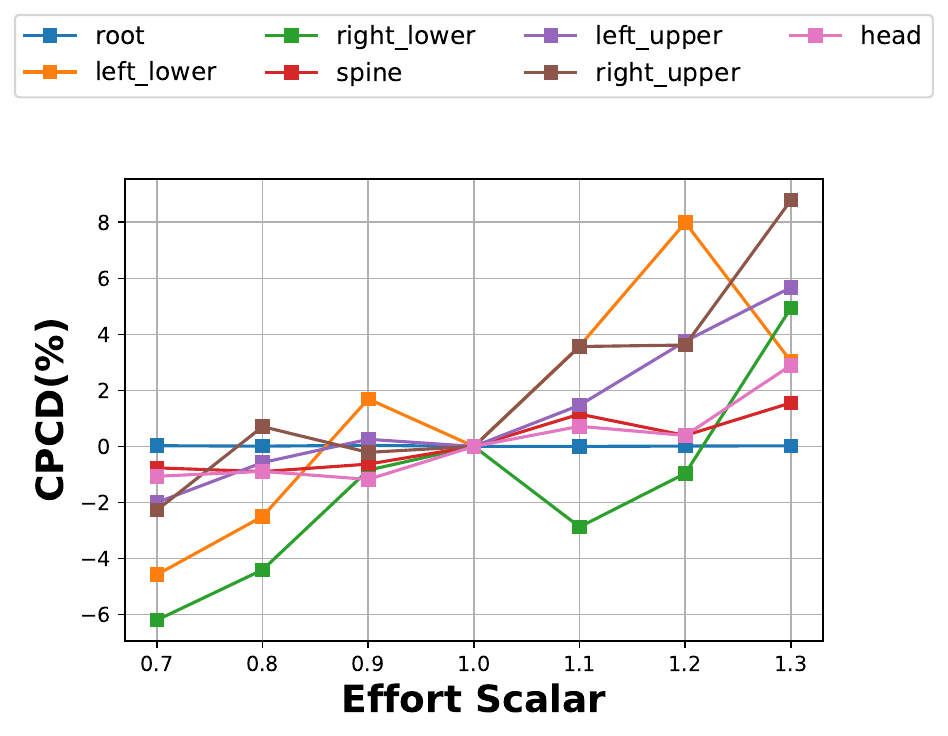}
\trendrow{Full body (Jumps)}{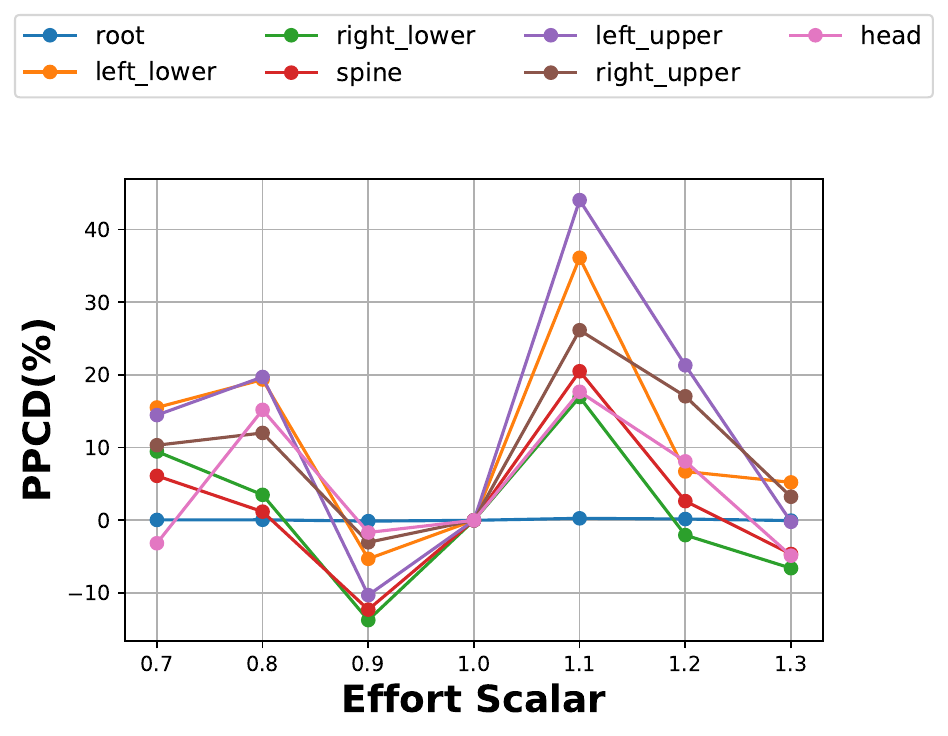}{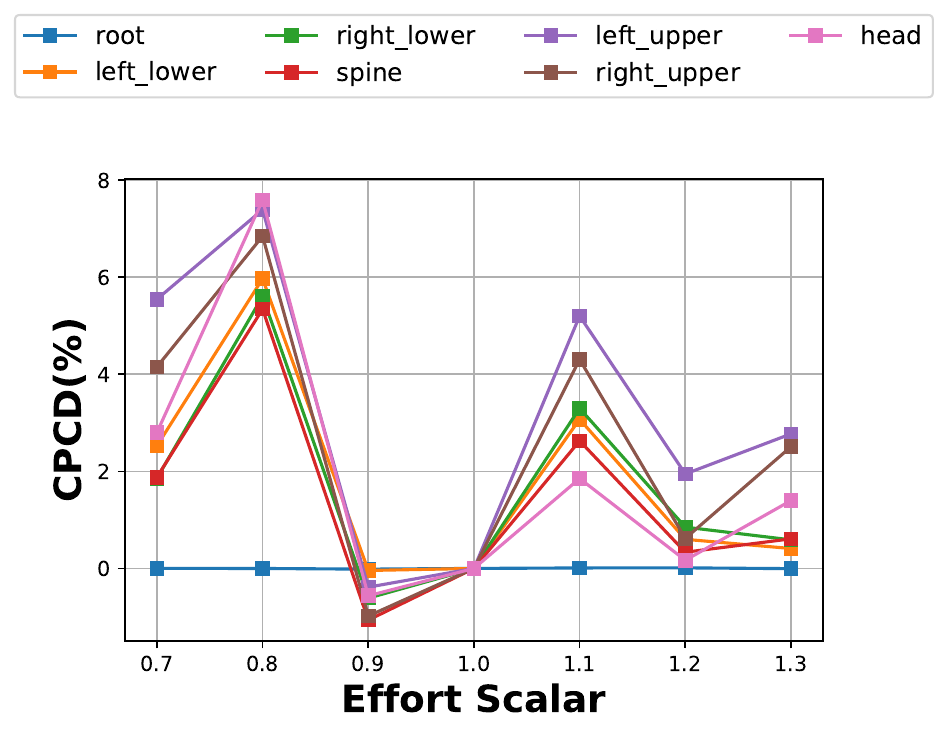}{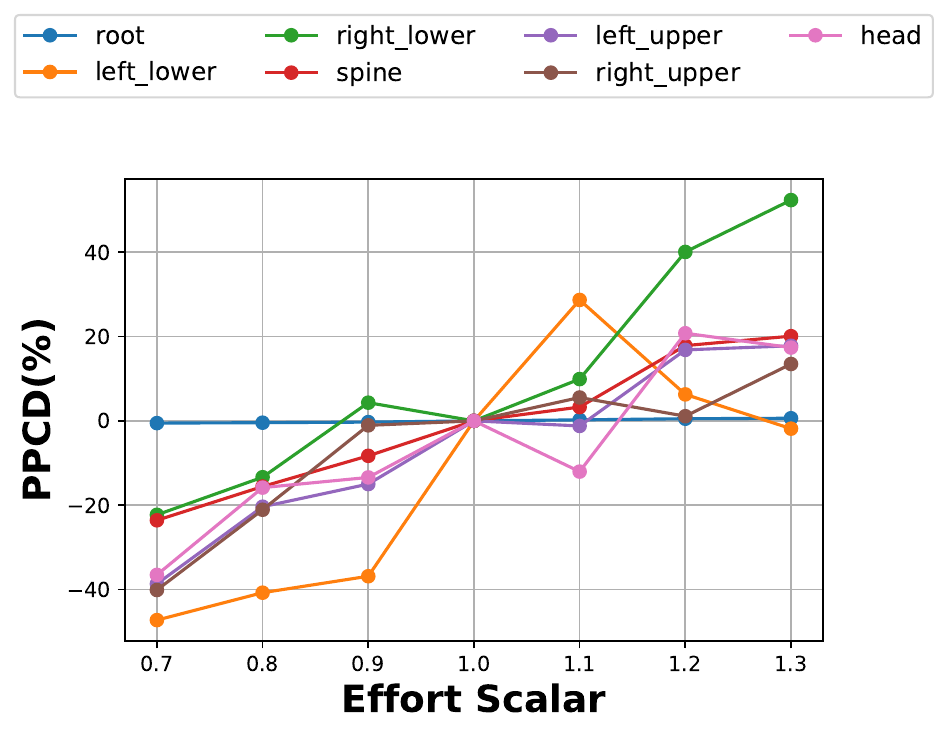}{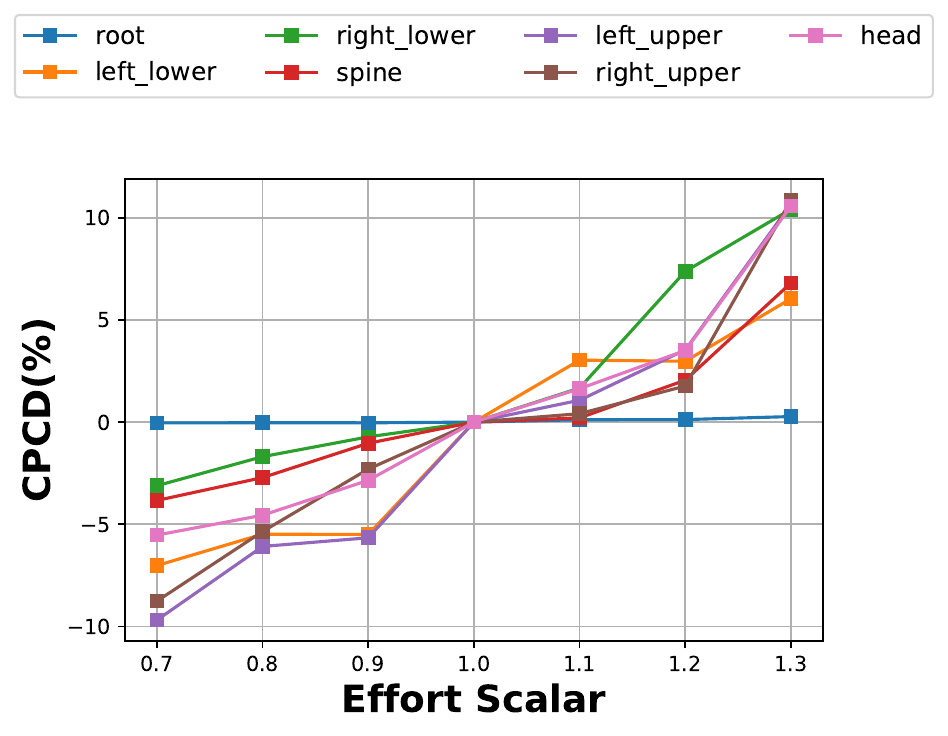}
\trendrow{Full body (Squats)}{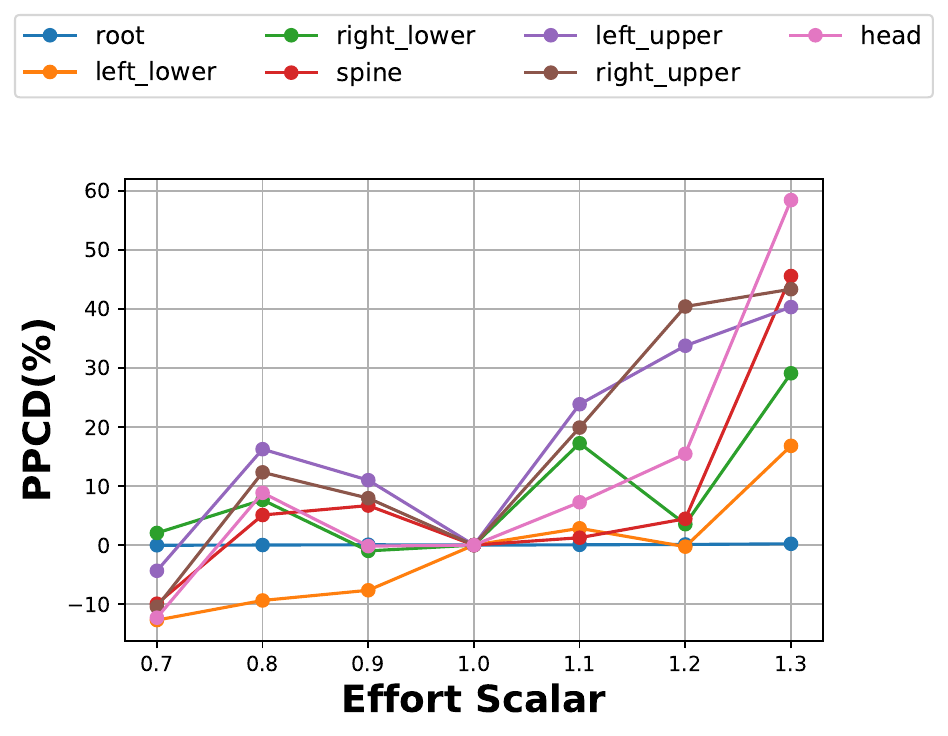}{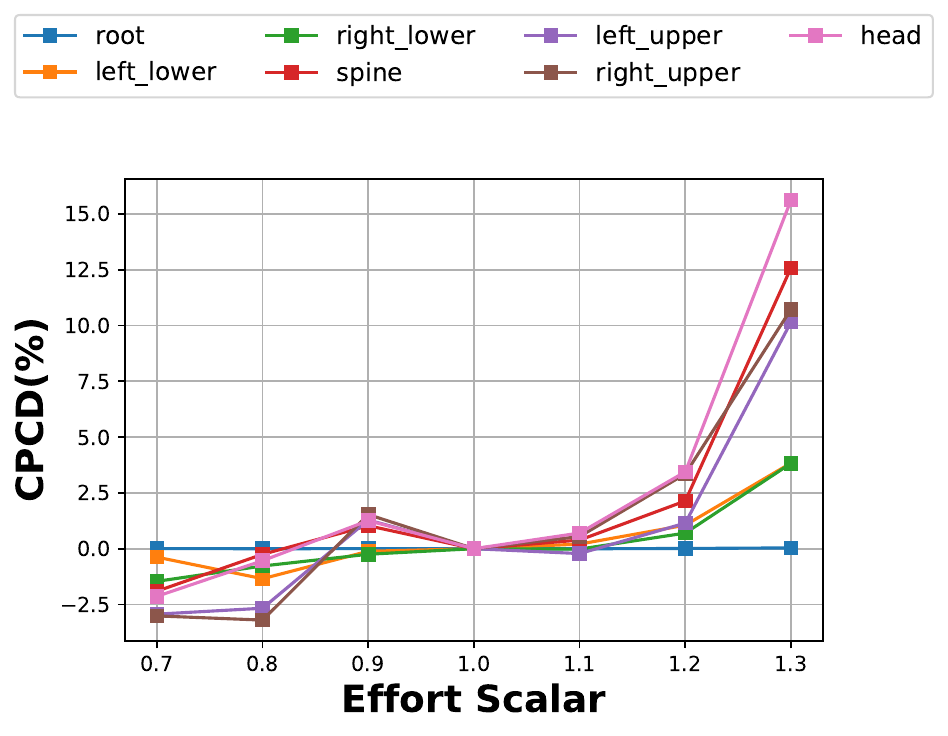}{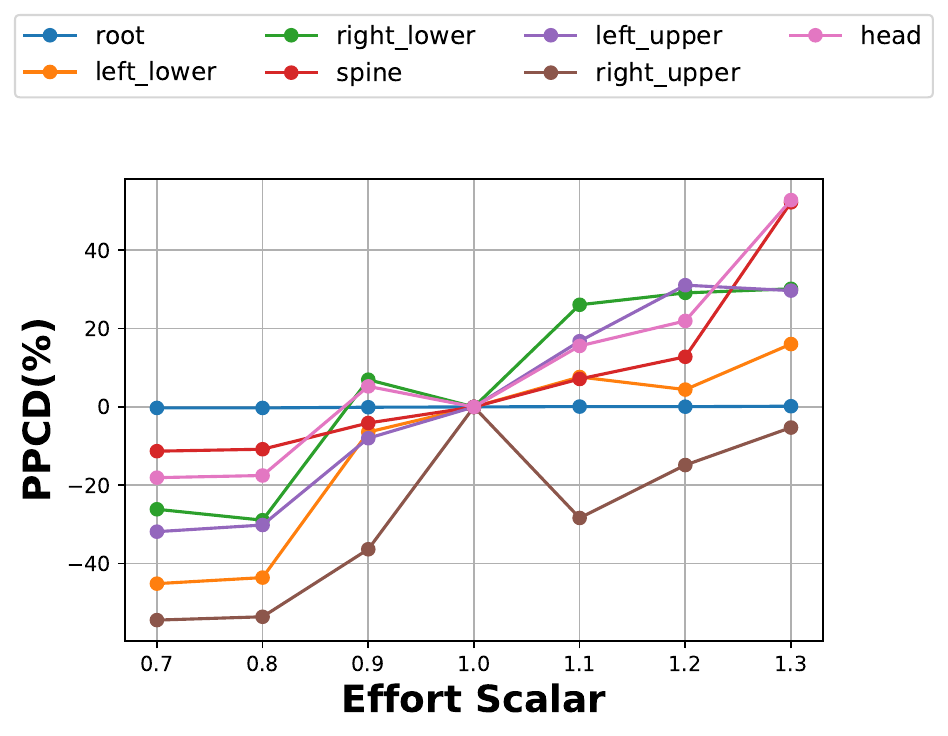}{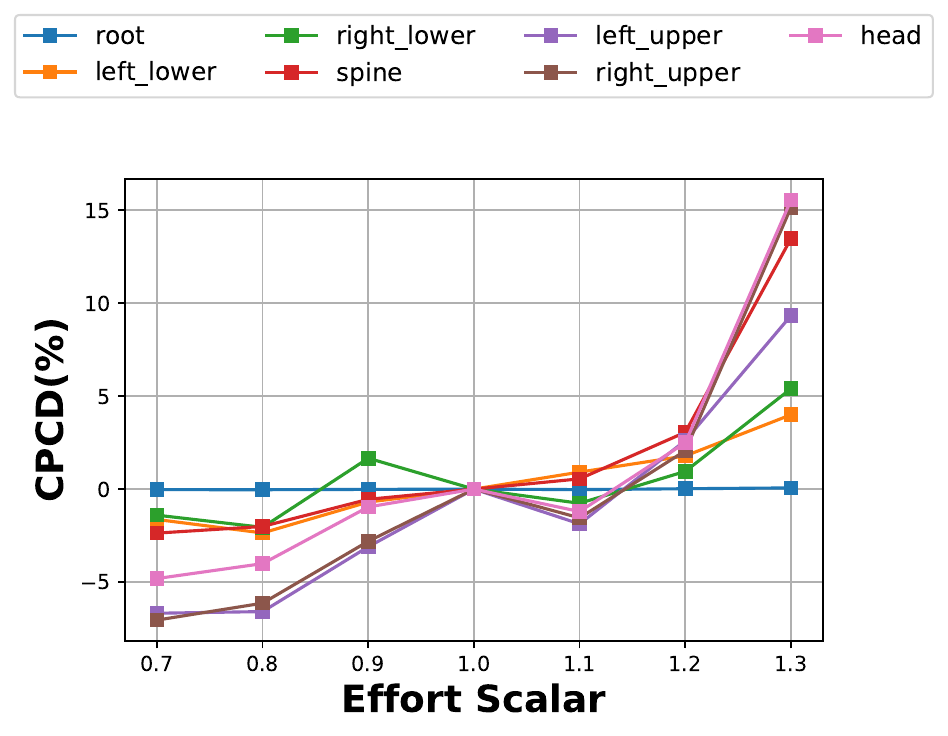}
\caption{Trend comparison across representative actions for SALAD and EMA (Part 4 of 5).}
\label{fig:trend_compact_4}
\end{figure*}

\begin{figure*}[t]
\centering
\trendrow{Upper body (Shakes Arms)}{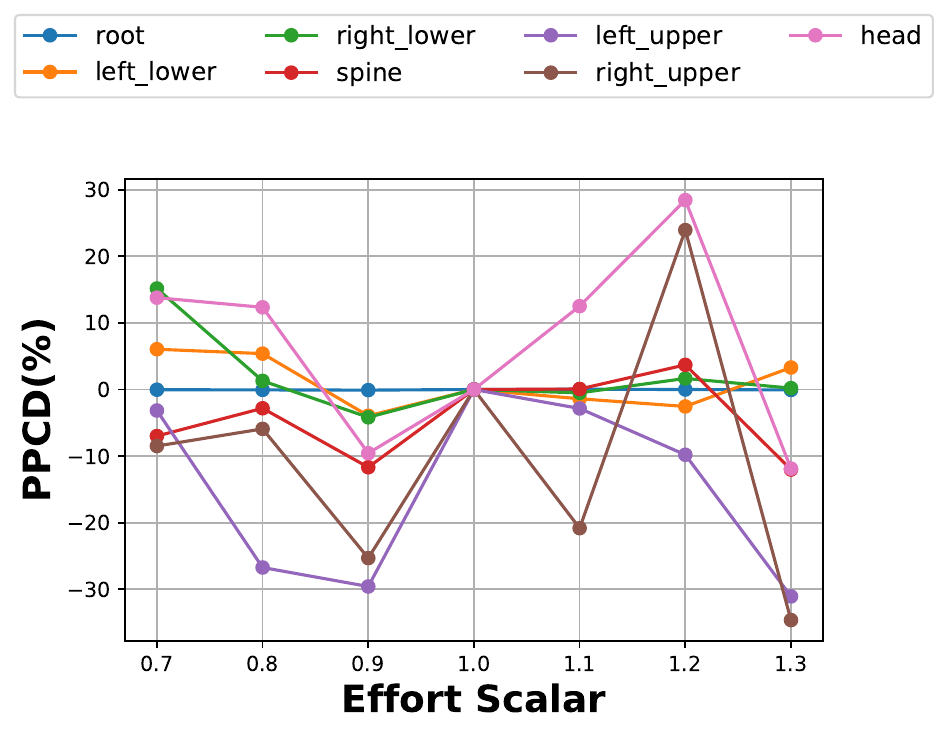}{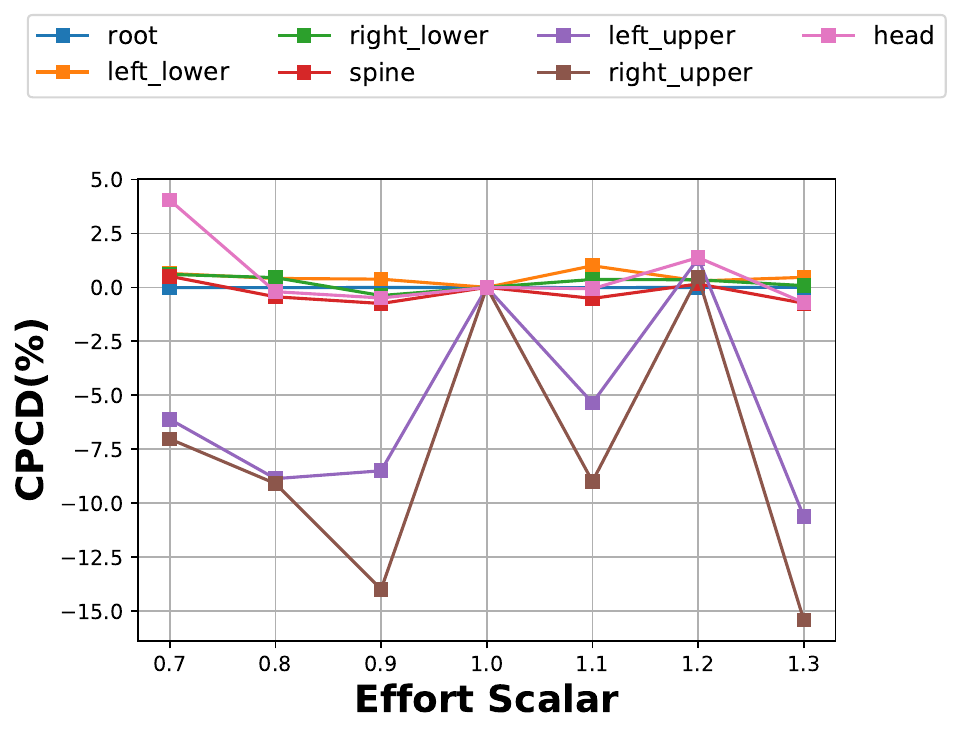}{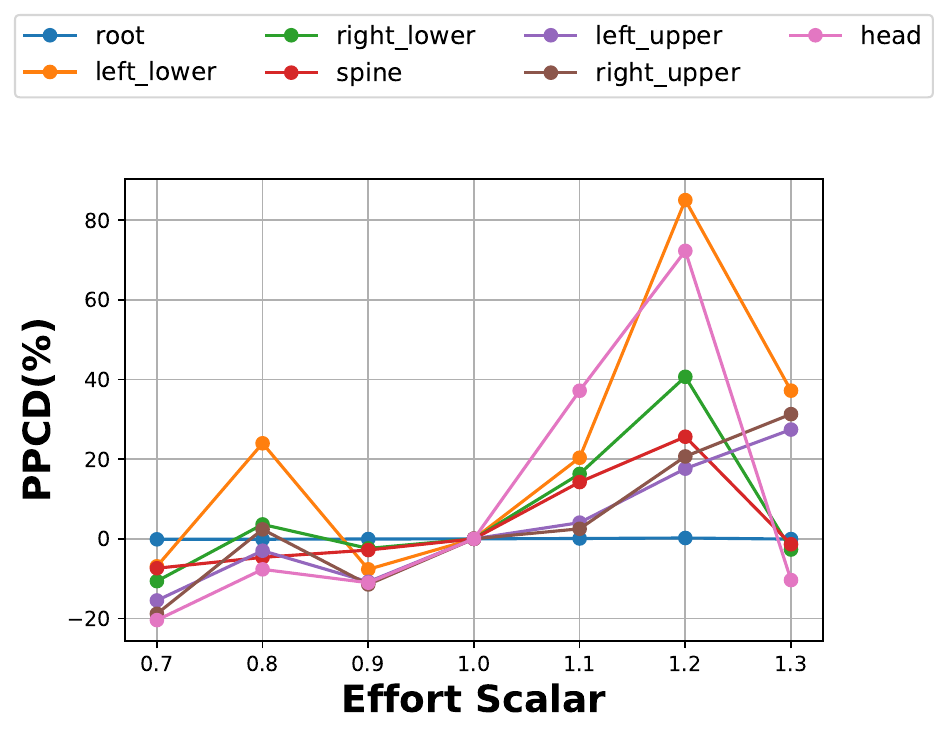}{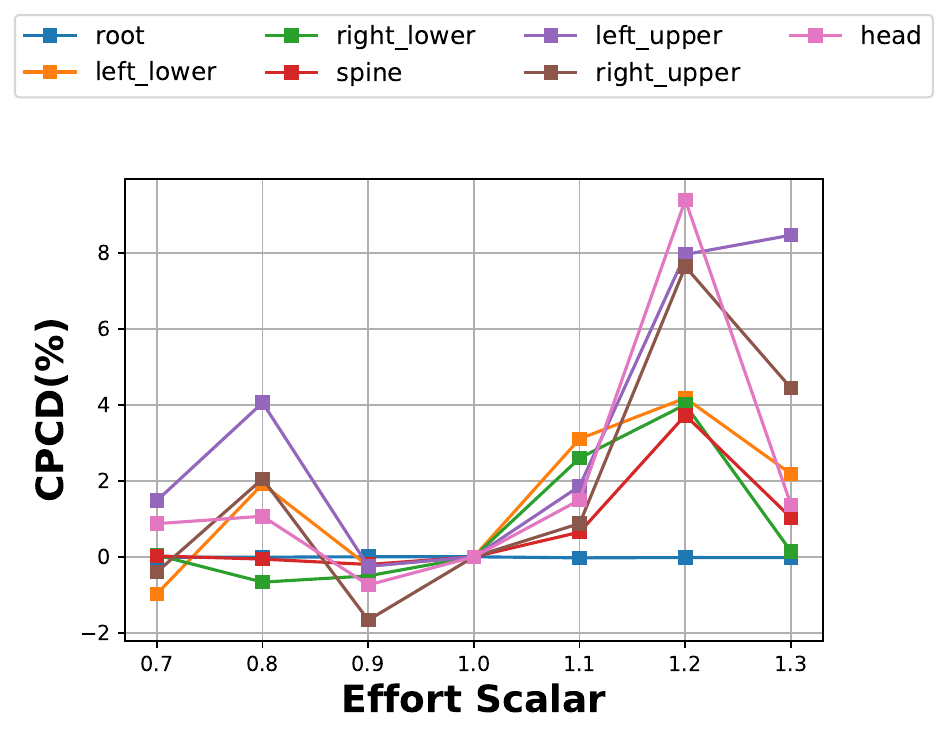}
\trendrow{Full body (Waves)}{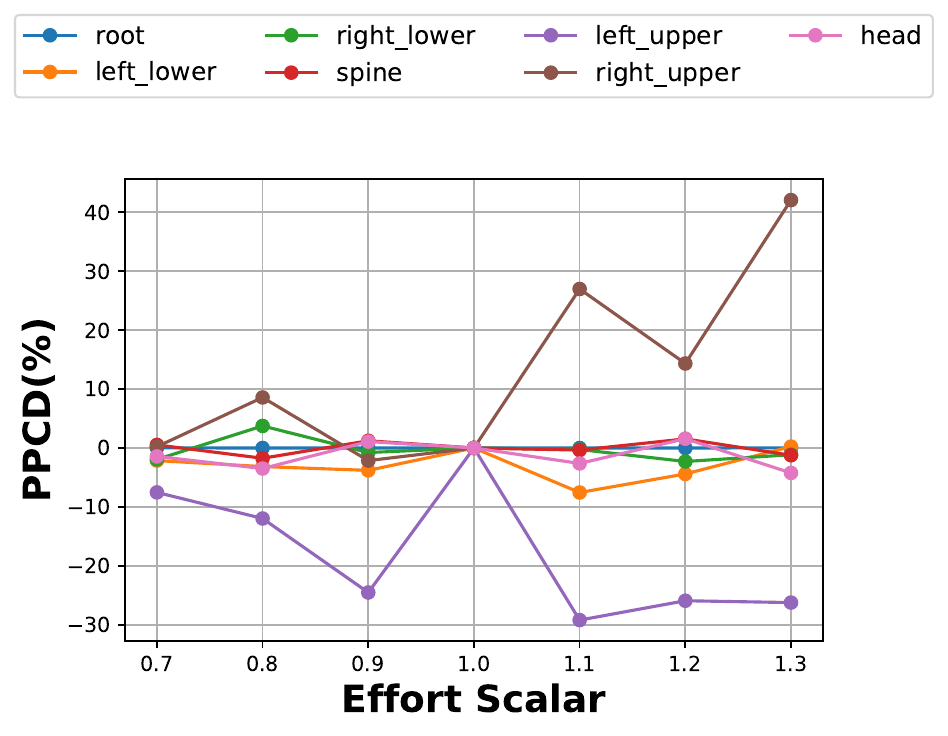}{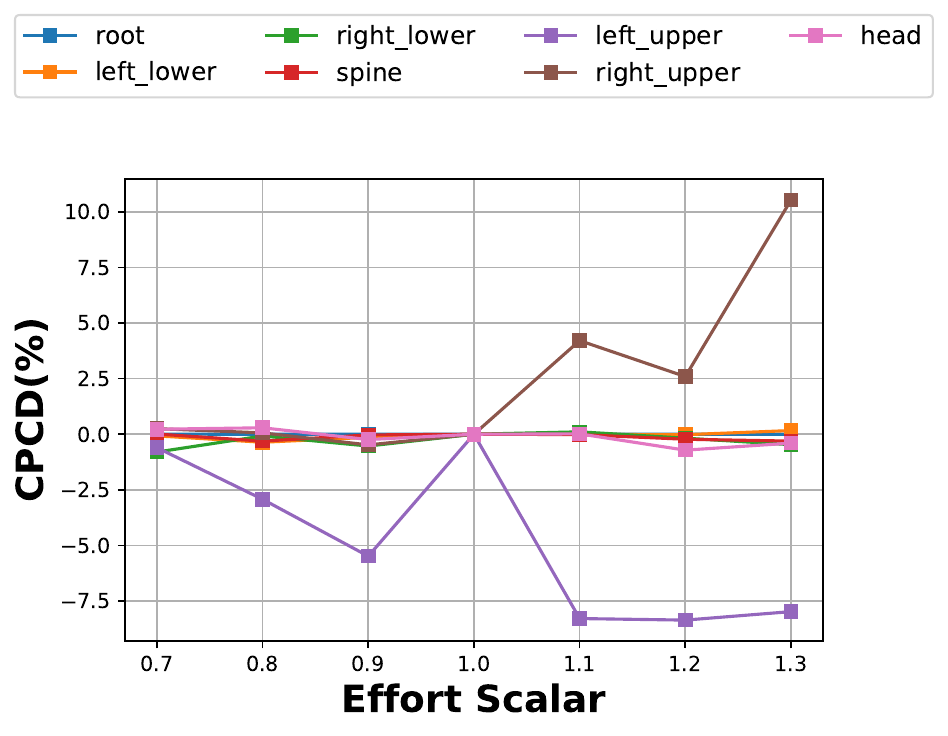}{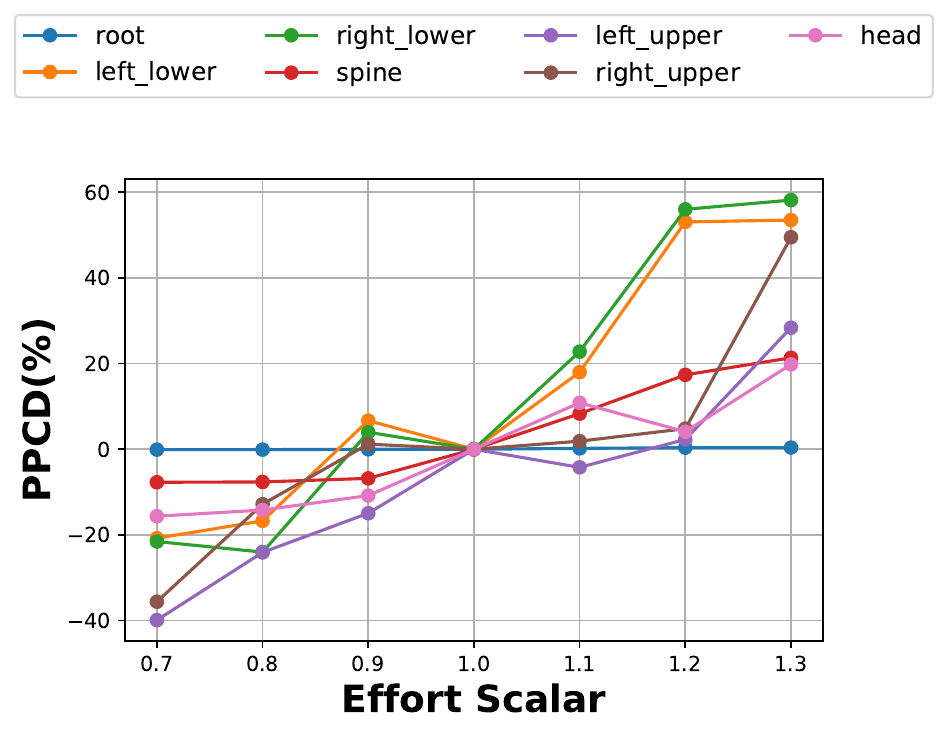}{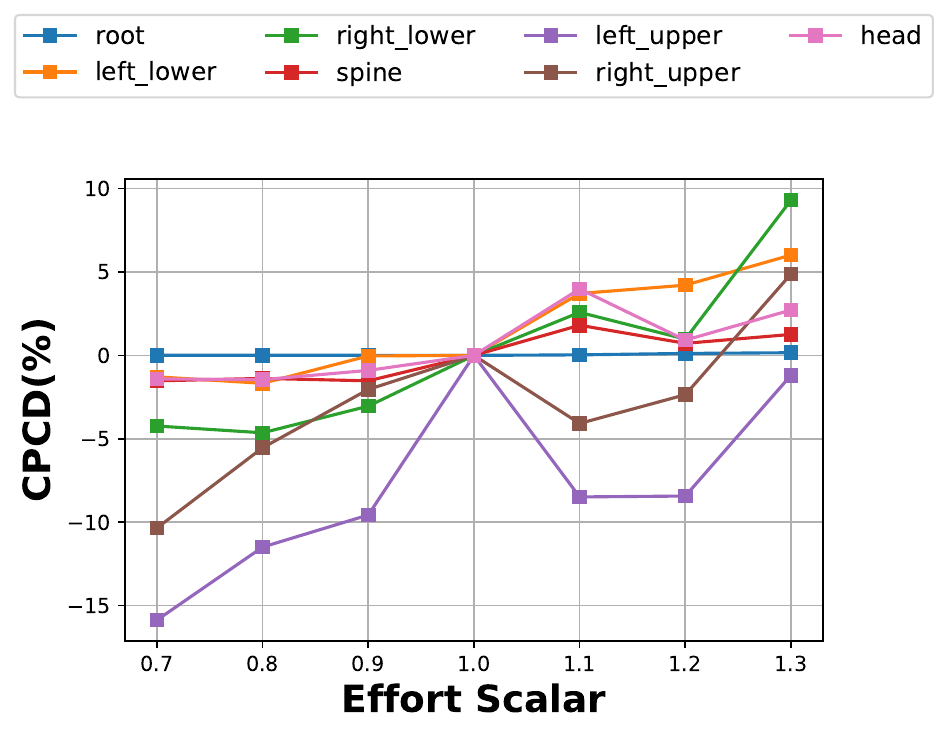}
\caption{Trend comparison across representative actions for SALAD and EMA (Part 5 of 5).}
\label{fig:trend_compact_5}
\end{figure*}

To provide additional evidence for the observed monotonic behavior, we conduct a detailed trend analysis of peak and collective positional change differences (PPCD, CPCD).
Trend plots (Figs.~\ref{fig:trend_compact_0}--\ref{fig:trend_compact_5}) visualize the response of these metrics to increasing effort scalars across 14 action classes, enabling both qualitative and statistical verification.
All referenced trend figures are provided together at the end of the supplementary material for ease of comparison.

\begin{figure*}[t]
\centering
\begin{subfigure}{0.9\textwidth}
  \includegraphics[width=\textwidth]{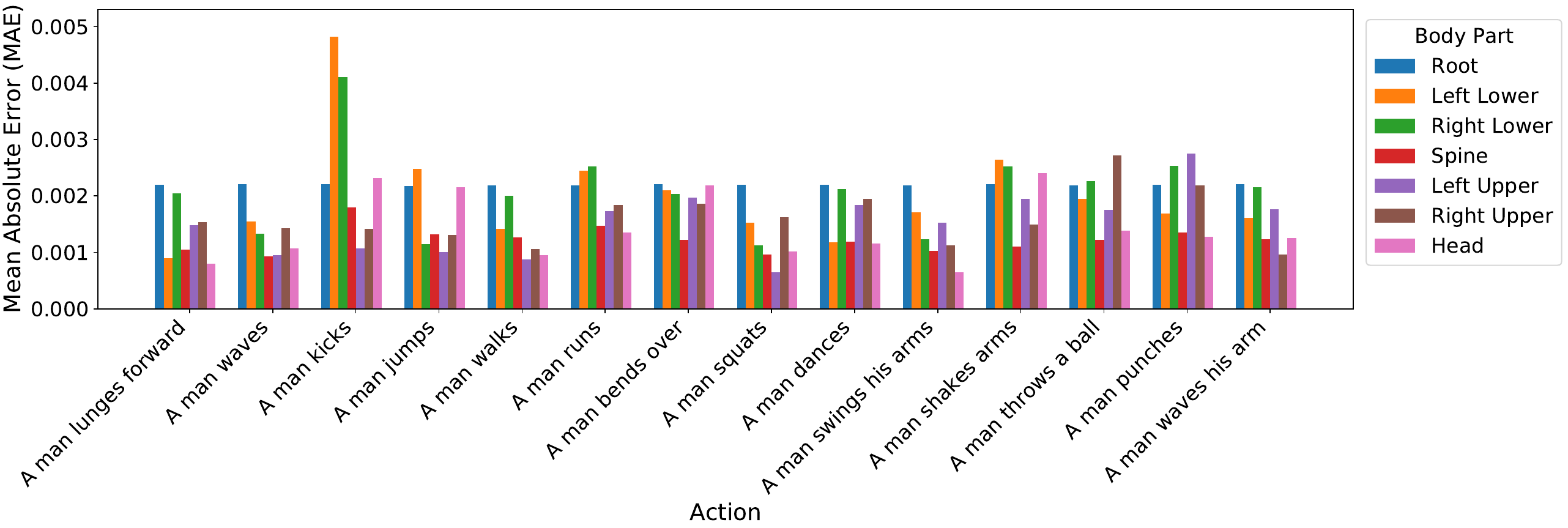}
  \caption{Peak MAE}
  \label{subfig:peak_mae}
\end{subfigure}
\begin{subfigure}{0.9\textwidth}
  \includegraphics[width=\textwidth]{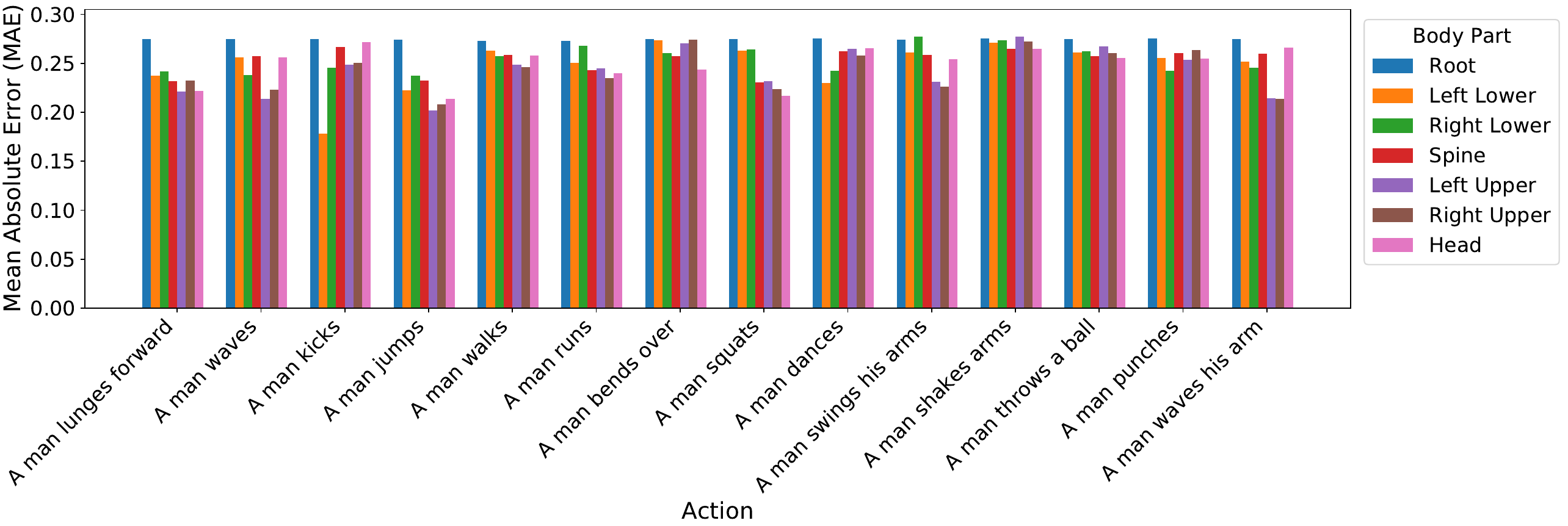}
  \caption{Collective MAE}
  \label{subfig:collective_mae}
\end{subfigure}
\caption{Evaluated actions with a length of 120 frames. 
Mean Absolute Error (MAE) between target and generated effort metrics across different actions and body parts.  
Errors remain below 0.005 for peak metrics but are substantially higher (0.2--0.3) for collective metrics due to accumulated per-frame deviations.}
\label{fig:MAE_metric}
\end{figure*}

Compared to SALAD, EMA exhibits consistently monotonic trends in peak positional changes, whereas SALAD shows irregular and non-monotonic responses under identical scaling.
These results corroborate the quantitative findings reported in the main paper and support the claim that EMA enables stable numerical control over motion intensity.
Additionally, for our generated samples, we observed that: Peak positional MAEs remain below 0.005 across all skeletal groups (Fig.~\ref{fig:MAE_metric}), indicating that monotonic control is achieved with high numerical fidelity.
Collective positional changes yield larger MAEs $<$ 0.3 (Fig.~\ref{fig:MAE_metric}) due to accumulated per-frame deviations, but still preserve monotonic trends that support qualitative controllability.

Supplementary videos are provided to facilitate visual inspection of the generated motions.
These include representative EMA samples, complex motion cases, and isolation tests corresponding to the examples shown in Fig.~\ref{fig:pnc2}.

%---------------------------------------------------------------------------------
\section{Relation to Other Motion Diffusion Control Paradigms}
\label{sec:task_diff}
%---------------------------------------------------------------------------------

To contextualize the contribution of EMA, we provide a structured comparison with existing controllable Motion Diffusion Models (MDMs).
As summarized in Table~\ref{tab:mdm_variant_scope}, prior adaptations primarily address control through spatial constraints (e.g., trajectories) or stylistic transfer via text or reference motions.
In contrast, EMA introduces a distinct \emph{Quantitative-Local} control axis, enabling direct modulation of motion intensity—how an action is executed—rather than where it moves or which style it imitates.

%---------------------------------------------------------------------------------
\subsection{Extended Comparison and Taxonomy}
%---------------------------------------------------------------------------------

As illustrated in Table~\ref{tab:mdm_variant_scope}, representative MDM variants can be categorized according to their conditioning modalities, control mechanisms, and spatial scope.
While this taxonomy is not exhaustive, it captures the dominant control paradigms most relevant to intensity modulation.
We observe three prevalent categories:

\begin{table*}[h]
    \caption{Comparison of MDM variants and their conditioning scopes. We categorize models by input modalities (Text, Numerical, Style, Trajectory), control mechanisms, and the spatial scope of the condition (Global vs. Local).}
    \centering
    \resizebox{\textwidth}{!}{%
    \begin{tabular}{l | c | c | c | c | c | c}
        \hline
        \textbf{Model} & \textbf{Text} & \textbf{Numerical} & \textbf{Style} & \textbf{Trajectory} & \textbf{Control} & \textbf{Condition Scope} \\
        \hline
        \text{MDM}~\cite{tevet2022human} & \checkmark & - & - & - & Inpainting & \text{Global} (Text) / \text{Local} (Inpainting) \\
        \hline
        \text{OmniControl}~\cite{xie2024omnicontrol} & \checkmark & - & - & \checkmark & Keyframes \& Waypoints & \text{Local} (Spatial Path \& Waypoints) \\
        \hline
        \text{PhysDiff}~\cite{yuan2022physdiff} & \checkmark & \checkmark (Physics) & - & - & - & \text{Global} (Text) \\
        \hline
        \text{FlexMotion}~\cite{tashakori2025flexmotion} & \checkmark & \checkmark (Phys. States) & - & \checkmark & Physics-Aware AE & \text{Dense} (Per-Frame Velocities) \\
        \hline
        \text{SMooDi}~\cite{zhong2024smoodistylizedmotiondiffusion} & \checkmark & - & \checkmark (Ref Motion) & - & Addition & \text{Global} (Style Token) \\
        \hline
        \text{LoraMDM}~\cite{sawdayee2025dance} & \checkmark & - & \checkmark (Ref Motion) & \checkmark & Style Token & \text{Global} (Learned Token) \\
        \hline
        \text{MotionDiffuse}~\cite{zhang2022motiondiffuse} & \checkmark & - & - & - & Probabilistic & \text{Global} (Text) \\
        \hline
        \text{InterGen}~\cite{liang2024intergen} & \checkmark & - & - & \checkmark & Interaction & \text{Global} (Text) + \text{Local} (Spatial) \\
        \hline
        \text{GMD}~\cite{karunratanakul2023guidedmotiondiffusioncontrollable} & \checkmark & - & - & \checkmark & Obstacles / Spatial & \text{Global} (Text) + \text{Local} (Trajectory) \\
        \hline
        \text{PRIMAL}~\cite{primal:iccv:2025} & Actions & \checkmark (Impulse) & \checkmark (Few-shot) & \checkmark & Reaction / Adaptor & \text{Instant} (Impulse) + \text{Global} (Goal) \\
        \hline
        \text{MoMo}~\cite{raab2024monkeyseemonkeydo} & \checkmark & - & \checkmark (Ref Motion) & - & Attn Injection & \text{Global} (Attn Swap) \\
        \hline
        \text{Motion Puzzle}~\cite{Jang_motion_puzzle_2022} & - & - & \checkmark (Ref Motion) & - & Body Parts & \text{Local} (Body Segmentation) \\
        \hline
        \text{MotionCLR}~\cite{motionclr} & \checkmark & - & \checkmark (Ref Motion) & - & Zero-Shot & \text{Global} (Contrastive) \\
        \hline
        \text{Style-ERD}~\cite{tao2022style} & - & - & \checkmark (Ref Motion) & - & Online Transfer & \text{Instant} (Real-time Stream) \\
        \hline
        \textbf{EMA}~(ours) & \checkmark & \checkmark (Peak/Coll.) & \checkmark (Metrics) & - & Cross-Attn & \textbf{Local} (Body-Part Effort) \\
        \hline
    \end{tabular}
    }
    \label{tab:mdm_variant_scope}
\end{table*}

\begin{itemize}
    \item \textbf{Semantic-Only Control.}
    Models such as MDM and MotionDiffuse rely exclusively on the expressiveness of text embeddings.
    Motion variation is achieved through categorical changes in the prompt, often resulting in discrete semantic shifts rather than smooth, continuous scaling of motion intensity.
    
    \item \textbf{Spatial-Numerical Control.}
    Approaches, including OmniControl and GMD, incorporate numerical inputs representing physical states or spatial coordinates.
    These methods prioritize trajectory adherence and physical feasibility by explicitly constraining the external path of the root or limbs within a coordinate space.
    
    \item \textbf{Reference-Based Style Control.}
    Methods such as SMooDi and Motion Puzzle leverage reference motion clips to transfer abstract stylistic attributes through global latent embeddings.
    Control is therefore indirect and dependent on the availability and selection of representative reference motions.
\end{itemize}

%---------------------------------------------------------------------------------
\subsection{A Complementary Control Layer}
%---------------------------------------------------------------------------------

EMA introduces a fundamentally different task, referred to as \emph{Quantitative Intensity Modulation}.
Rather than encoding physical states or stylistic exemplars, EMA conditions generation on numerical scalars corresponding to human-perceived effort.
This mechanism modifies internal kinematic intensity without explicitly constraining trajectories.

\paragraph{Differentiation from Trajectory-Based Models.}
EMA is not designed for trajectory following and should not be directly compared with spatially constrained approaches such as GMD.
Because EMA does not require explicit spatial paths, it functions as a complementary control layer, particularly suited to scenarios in which the trajectory is implicitly defined by the text prompt while motion intensity must be adjusted dynamically.

\paragraph{Differentiation from Style Transfer Methods.}
Unlike reference-based style transfer, EMA does not rely on exemplar motion clips.
Instead, it employs explicit numerical parameters to directly modulate motion dynamics, enabling continuous, parametric intensity control unattainable through discrete style tokens or reference-driven embeddings.

\paragraph{Unique Contribution.}
EMA occupies a distinct niche by bridging high-level semantic intent and low-level kinematic intensity while preserving local body-part independence.
By introducing a Quantitative-Local control axis, EMA complements existing semantic, spatial, and physics-based diffusion frameworks rather than competing with them.

\begin{figure*}[t]
\centering
\trendrow{Lower body (Runs)}{fig/trendplotsv6/SALAD/amanruns_peak.pdf}{fig/trendplotsv6/SALAD/amanruns_collective.pdf}{fig/trendplotsv6/EMA/amanruns_peak.pdf}{fig/trendplotsv6/EMA/amanruns_collective.pdf}
\trendrow{Upper body (Swings)}{fig/trendplotsv6/SALAD/amanswingshisarms_peak.pdf}{fig/trendplotsv6/SALAD/amanswingshisarms_collective.pdf}{fig/trendplotsv6/EMA/amanswingshisarms_peak.pdf}{fig/trendplotsv6/EMA/amanswingshisarms_collective.pdf}
\trendrow{Full body (Bends Over)}{fig/trendplotsv6/SALAD/amanbendsover_peak.pdf}{fig/trendplotsv6/SALAD/amanbendsover_collective.pdf}{fig/trendplotsv6/EMA/amanbendsover_peak.pdf}{fig/trendplotsv6/EMA/amanbendsover_collective.pdf}
\caption{Trend comparison across representative actions for SALAD and EMA (Part 1 of 5).
Each row shows SALAD-PPCD, SALAD-CPCD, EMA-PPCD, and EMA-CPCD.
PPCD = Peak Positional Change Difference, CPCD = Collective Positional Change Difference.
EMA yields monotonic, near-proportional scaling (notably in PPCD), while SALAD exhibits irregular trends and asymmetries.
For SALAD, the horizontal axis values (0.7--1.3) correspond to adverbs: \textit{Extremely Slow} (0.7), \textit{Very Slow} (0.8), \textit{Slow} (0.9), \textit{Normal} (1.0), \textit{Fast} (1.1), \textit{Very Fast} (1.2), and \textit{Extremely Fast} (1.3). }
\label{fig:trend_compact_0}
\end{figure*}

% Figure 2: Next 5 actions

% Figure 3: Last 4 actions, including outliers

% Figure 4: Last 4 actions, including outliers

\end{document}